\documentclass{article}
\usepackage{ijcai23}

\usepackage{times}
\usepackage{soul}
\usepackage{url}
\usepackage[hidelinks]{hyperref}
\usepackage[utf8]{inputenc}
\usepackage[small]{caption}
\usepackage{graphicx}
\usepackage{amsmath}
\usepackage{amsthm}
\usepackage{booktabs}
\usepackage{algorithm}
\usepackage[noend]{algpseudocode}
\usepackage[switch]{lineno}
\usepackage{mathtools}
\usepackage{hhline}
\usepackage{makecell}
\usepackage{multirow}
\usepackage{amsfonts}
\usepackage{caption}
\usepackage{subcaption}
\usepackage{changepage}
\usepackage{amsthm}
\usepackage{amsmath}
\usepackage{xspace}
\usepackage{colortbl}
\usepackage{booktabs}
\usepackage{tabularx}
\usepackage{tikz}
\usetikzlibrary{calc,positioning}

\newcommand{\spara}[1]{\smallskip\noindent{\bf #1}}

\newcommand{\para}[1]{\noindent{\bf #1}}

\algnewcommand\Params{\item[\textbf{Params:}]}

\definecolor{cycle1}{RGB}{228, 26, 28}
\definecolor{cycle2}{RGB}{55, 126, 184}
\definecolor{cycle3}{RGB}{77, 175, 74}
\definecolor{cycle4}{RGB}{152, 78, 163}
\definecolor{cycle5}{RGB}{255, 127, 0}
\definecolor{cycle6}{RGB}{153, 153, 153}
\definecolor{cycle7}{RGB}{166, 86, 40}
\definecolor{cycle8}{RGB}{247, 129, 191}

\newcommand*{\bigO}{\mathcal{O}}

\newcommand{\reals}{\mathbb{R}}

\newcommand{\ourname}{ActUp\xspace}
\newcommand{\ourmethod}{GDR\xspace}
\newcommand{\ourmethodU}{$\text{\ourmethod}_{\text{umap}}$\xspace}
\newcommand{\ourmethodN}{$\text{\ourmethod}_{\text{tsne}}$\xspace}

\newcommand{\ourcell}{\cellcolor{cycle3!20}}

\newtheorem{theorem}{Theorem}

\title{\ourname: Analyzing and Consolidating tSNE \& UMAP}

\author{
Andrew Draganov$^1$
\and
Jakob Rødsgaard Jørgensen$^1$\and
Katrine Scheel Nellemann$^1$\and
Davide Mottin$^1$\and
Ira Assent$^1$\and
Tyrus Berry$^2$\and
Cigdem Aslay$^1$
\affiliations
$^1$Aarhus University\\
$^2$George Mason University\\
\emails
\{draganovandrew, jakobrj, scheel, davide, ira, cigdem\}@cs.au.dk,
tberry@gmu.edu,
}

\begin{document}

\maketitle

\begin{abstract}
tSNE and UMAP are popular dimensionality reduction algorithms due to their speed and interpretable low-dimensional embeddings. Despite their popularity, however, little work has been done to study their full span of differences. We theoretically and experimentally evaluate the space of parameters in both tSNE and UMAP and observe that a single one -- the normalization -- is responsible for switching between them. This, in turn, implies that a majority of the algorithmic differences can be toggled without affecting the embeddings. We discuss the implications this has on several theoretic claims behind UMAP, as well as how to reconcile them with existing tSNE interpretations.

Based on our analysis, we provide a method (\ourmethod) that combines previously incompatible techniques from tSNE and UMAP and can replicate the results of either algorithm. This allows our method to incorporate further improvements, such as an acceleration that obtains either method's outputs faster than UMAP. We release improved versions of tSNE, UMAP, and \ourmethod that are fully plug-and-play with the traditional libraries.
\end{abstract}


\section{Introduction}
Dimensionality Reduction (DR) algorithms are invaluable for qualitatively inspecting high-dimensional data and are widely used across scientific disciplines. Broadly speaking, these algorithms transform a high-dimensional input into a faithful lower-dimensional embedding. This embedding aims to preserve similarities among the points, where similarity is often measured by distances in the corresponding spaces.

\begin{figure}[ht]
    \small
    \setlength{\tabcolsep}{6pt}
    \newcolumntype{C}{ >{\centering\arraybackslash} m{2.2cm} }
    \newcolumntype{D}{ >{\centering\arraybackslash} m{0.03cm} }
    \begin{tabularx}{\textwidth}{DCCC}
        & \quad \; MNIST & \hspace{5mm} \makecell{Fashion-\\MNIST} &  \quad \, Coil-100 \\
        \\[-1.2em]
        \rotatebox{90}{\bf tSNE} & 
        \includegraphics[width=2.7cm]{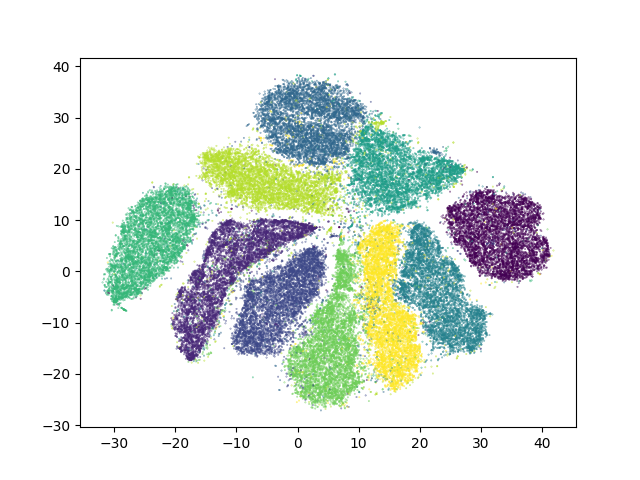} &
        \includegraphics[width=2.7cm]{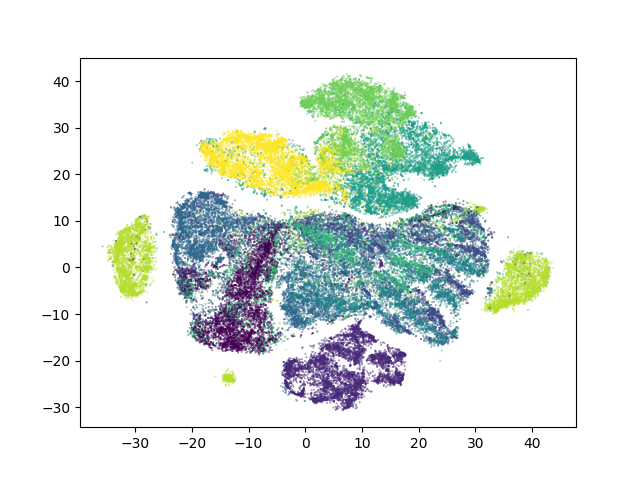} &
        \includegraphics[width=2.7cm]{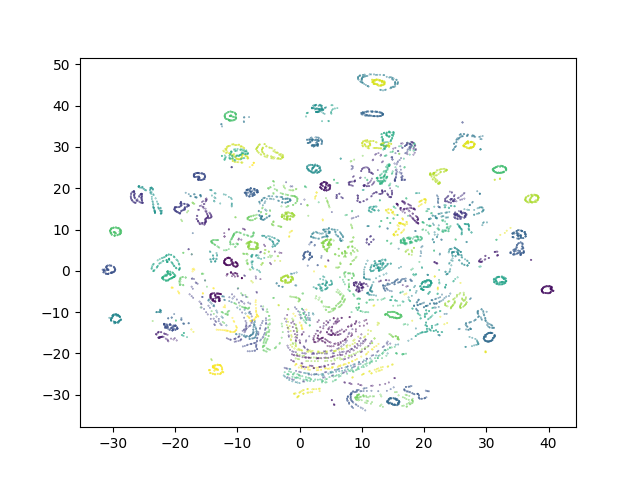} \\
        \\[-2.em]
        
        \rotatebox{90}{\bf \ourmethodN} &
        \includegraphics[width=2.7cm]{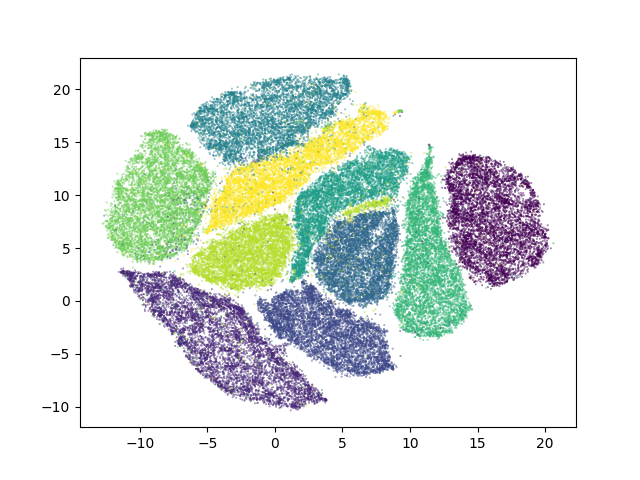} &
        \includegraphics[width=2.7cm]{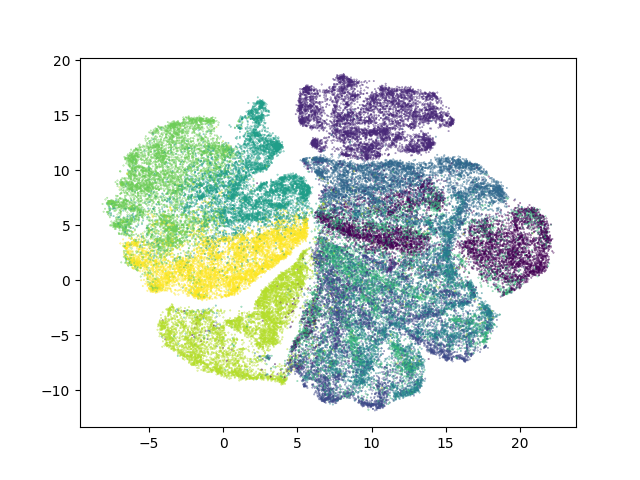} &
        \includegraphics[width=2.7cm]{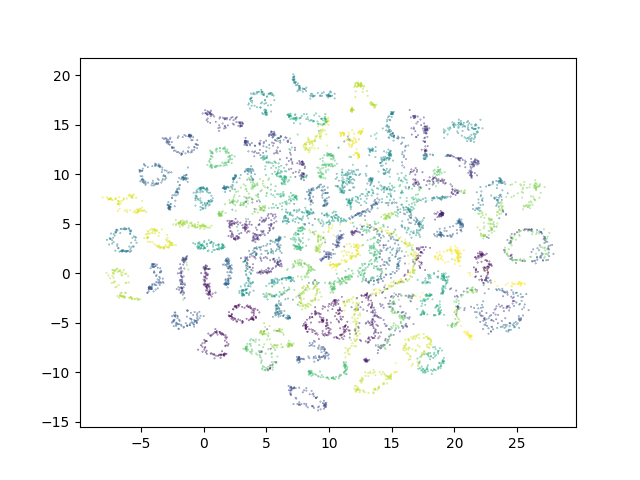} \\
        \\[-2em]
        \rotatebox{90}{\bf UMAP} &
        \includegraphics[width=2.7cm]{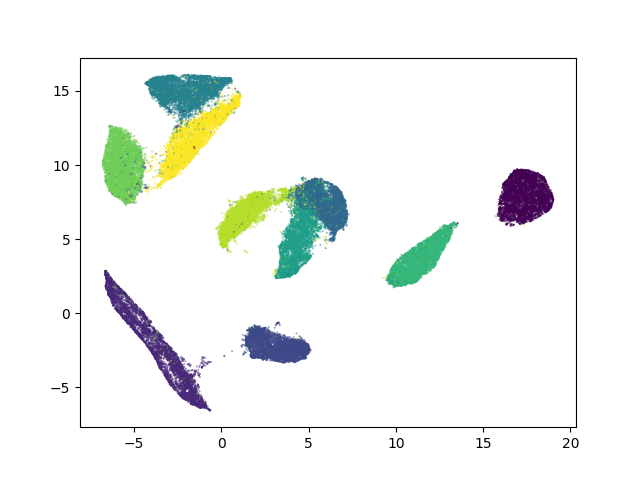} & 
        \includegraphics[width=2.7cm]{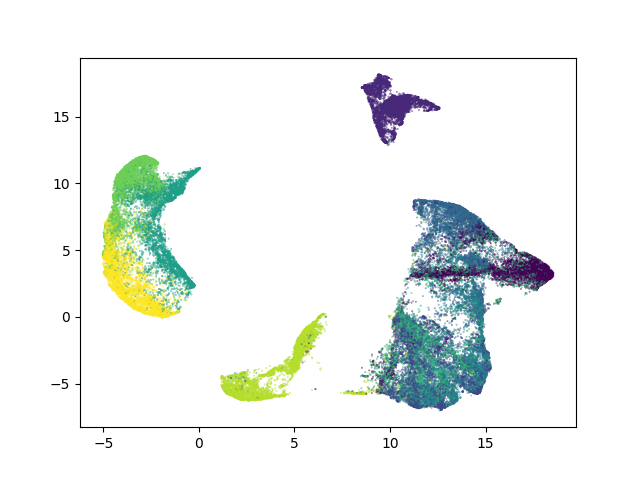} & 
        \includegraphics[width=2.7cm]{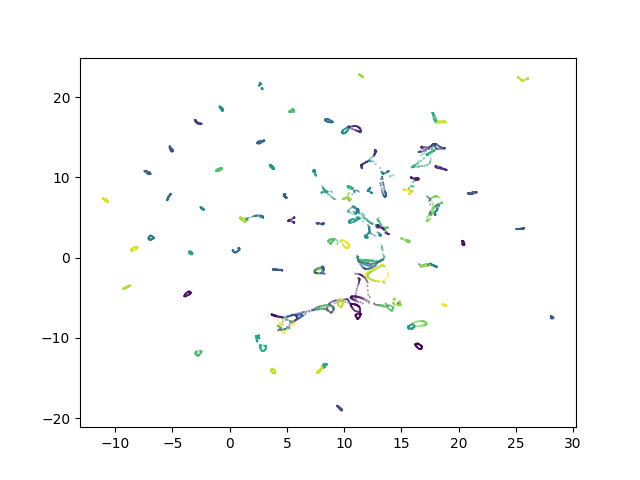} \\
        \\[-2.em]
        
        \rotatebox{90}{\bf \ourmethodU} &
        \includegraphics[width=2.7cm]{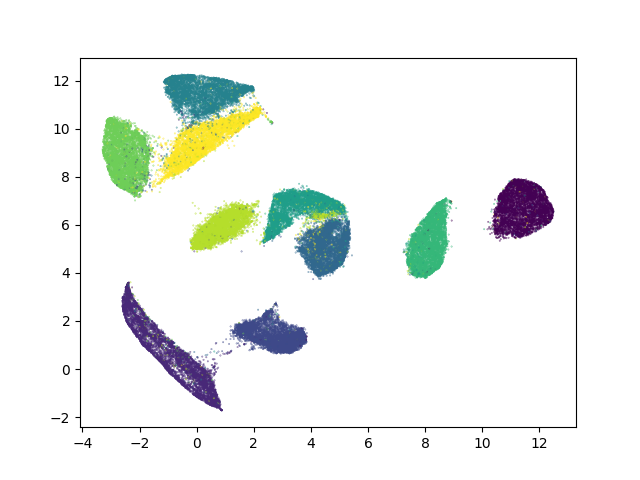} &
        \includegraphics[width=2.7cm]{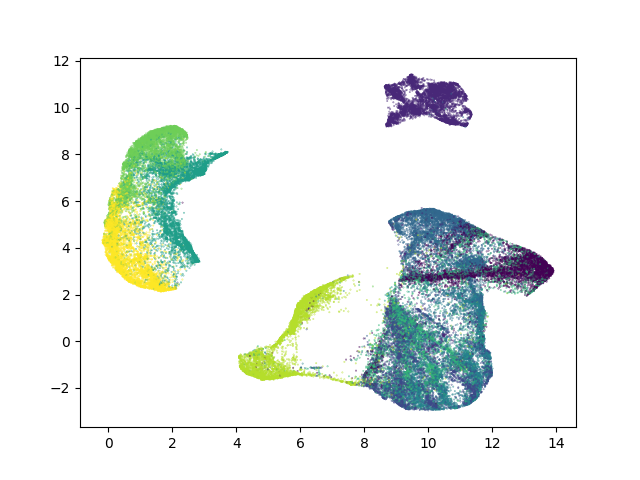} &
        \includegraphics[width=2.7cm]{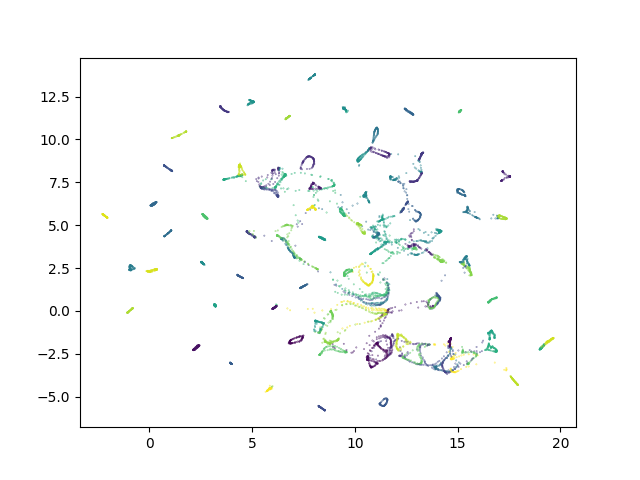} \\
        \\[-2em]
    \end{tabularx}
    \caption{A single method (\ourmethod) can recreate tSNE and UMAP outputs just by changing the normalization.}
    \label{embedding_vis}
\end{figure}

tSNE~\cite{van2008visualizing} \cite{van2014accelerating} and UMAP~\cite{mcinnes2018umap} are two widely popular DR algorithms due to their efficiency and interpretable embeddings. Both algorithms establish analogous similarity measures, share comparable loss functions, and find an embedding through gradient descent. Despite these similarities, tSNE and UMAP have several key differences. First, although both methods obtain similar results, UMAP prefers large inter-cluster distances while tSNE leans towards large intra-cluster distances. Second, UMAP runs significantly faster as it performs efficient sampling during gradient descent. While attempts have been made to study the gaps between the algorithms \cite{kobak2021initialization},~\cite{bohm2020unifying},~\cite{damrich2021umap}, there has not yet been a comprehensive analysis of their methodologies nor a method that can obtain both tSNE and UMAP embeddings at UMAP speeds.

We believe that this is partly due to their radically different presentations. While tSNE takes a computational angle, UMAP originates from category theory and topology. Despite this, many algorithmic choices in UMAP and tSNE are presented without theoretical justification, making it difficult to know which algorithmic components are necessary.

In this paper, we make the surprising discovery that the differences in \textit{both} the embedding structure \textit{and} computational complexity between tSNE and UMAP can be resolved via a single algorithmic choice -- the normalization factor. We come to this conclusion by deriving both algorithms from first principles and theoretically showing the effect that the normalization has on the gradient structure. We supplement this by identifying every implementation and hyperparameter difference between the two methods and implementing tSNE and UMAP in a common library. Thus, we study the effect that each choice has on the embeddings and show both quantitatively and qualitatively that, other than the normalization of the pairwise similarity matrices, none of these parameters significantly affect the outputs.

Based on this analysis, we introduce the necessary changes to the UMAP algorithm such that it can produce tSNE embeddings as well. We refer to this algorithm as Gradient Dimensionality Reduction (\ourmethod) to emphasize that it is consistent with the presentations of \textit{both} tSNE and UMAP. We experimentally validate that \ourmethod can simulate both methods through a thorough quantitative and qualitative evaluation across many datasets and settings. Lastly, our analysis provides insights for further speed improvements and allows \ourmethod to perform gradient descent faster than the standard implementation of UMAP.

In summary, our contributions are as follows:
\begin{enumerate}
\item We perform the first comprehensive analysis of the differences between tSNE and UMAP, showing the effect of each algorithmic choice on the embeddings.
\item We theoretically and experimentally show that changing the normalization is a sufficient condition for switching between the two methods.
\item We release simple, plug-and-play implementations of \ourmethod, tSNE and UMAP that can toggle all of the identified hyperparameters. Furthermore, \ourmethod obtains embeddings for both algorithms faster than UMAP.
\end{enumerate}


\section{Related Work}
\label{sec:related_work}

When discussing tSNE we are referring to~\cite{van2014accelerating} which established the nearest neighbor and sampling improvements and is generally accepted as the standard tSNE method. A popular subsequent development was presented in~\cite{linderman2019fast}, wherein Fast Fourier Transforms were used to accelerate the comparisons between points. Another approach is LargeVis ~\cite{tang2016visualizing}, which modifies the embedding functions to satisfy a graph-based Bernoulli probabilistic model of the low-dimensional dataset. As the more recent algorithm, UMAP has not had as many variations yet. One promising direction, however, has extended UMAP's second step as a parametric optimization on neural network weights~\cite{sainburg2020parametric}.

Many of these approaches utilize the same optimization structure where they iteratively attract and repel points. While most perform their attractions along nearest neighbors in the high-dimensional space, the repulsions are the slowest operation and each method approaches them differently. tSNE samples repulsions by utilizing Barnes-Hut (BH) trees to sum the forces over distant points. The work in~\cite{linderman2019fast} instead calculates repulsive forces with respect to specifically chosen interpolation points, cutting down on the $\bigO(n\log n)$ BH tree computations. UMAP and LargeVis, on the other hand, simplify the repulsion sampling by only calculating the gradient with respect to a constant number of points. These repulsion techniques are, on their face, incompatible with one another, i.e., several modifications have to be made to each algorithm before one can interchange the repulsive force calculations.

There is a growing amount of work that compares tSNE and UMAP through a more theoretical analysis~\cite{damrich2021umap}\cite{bohm2020unifying}\cite{damrich2022contrastive}\cite{kobak2021initialization}. \cite{damrich2021umap} find that UMAP's algorithm does not optimize the presented loss and provide its effective loss function. Similarly \cite{bohm2020unifying} analyze tSNE and UMAP through their attractive and repulsive forces, discovering that UMAP diverges when using $\bigO(n)$ repulsions per epoch. We expand on the aforementioned findings by showing that the forces are solely determined by the choice of normalization, giving a practical treatment to the proposed ideas. Lastly, \cite{damrich2022contrastive} provides the interesting realization that tSNE and UMAP can both be described through contrastive learning approaches. Our work differs from theirs in that we analyze the full space of parameters in the algorithms and distill the difference to a single factor, allowing us to connect the algorithms without the added layers of contrastive learning theory. Lastly, the authors in~\cite{kobak2021initialization} make the argument that tSNE can perform UMAP's manifold learning if given UMAP's initialization. Namely, tSNE randomly initializes the low dimensional embedding whereas UMAP starts from a Laplacian Eigenmap~\cite{belkin2003laplacian} projection. While this may help tSNE preserve the local $k$NN structure of the manifold, it is not true of the macro-level distribution of the embeddings. Lastly, ~\cite{wang2021understanding} discusses the role that the loss function has on the resulting embedding structure. This is in line with our results, as we show that the normalization's effect on the loss function is fundamental in the output differences between tSNE and UMAP.

\section{Comparison of tSNE and UMAP} \label{comparison}
We begin by formally introducing the tSNE and UMAP algorithms. Let $X \in \reals^{n \times D}$ be a high dimensional dataset of $n$ points and let $Y\in \reals^{n\times d}$ be a previously initialized set of $n$ points in lower-dimensional space such that $d < D$. Our aim is to define similarity measures between the points in each space and then find the embedding $Y$ such that the pairwise similarities in $Y$ match those in $X$.

To do this, both algorithms define high- and low-dimensional non-linear functions $p: X \times X \rightarrow [0, 1]$ and $q: Y \times Y \rightarrow [0, 1]$. These form pairwise similarity matrices $P(X),Q(Y) \in \reals^{n \times n}$, where the $i, j$-th matrix entry represents the similarity between points $i$ and $j$. Formally,
\begin{equation}
\begin{aligned}
    p^{tsne}_{j|i}(x_i, x_j) &= \dfrac{\text{exp}(-d(x_i, x_j)^2 / 2 \sigma_i^2)}{\sum_{k \neq l} \text{exp}(-d(x_k, x_l)^2 / 2 \sigma_k^2)} \\[0.5ex]
    q^{tsne}_{ij}(y_i, y_j) &= \dfrac{(1 + ||y_i - y_j||^2_2)^{-1}}{\sum_{k \neq l} (1 + ||y_k - y_l||^2_2)^{-1}}
\end{aligned}
\label{eq:tsne_prob}
\end{equation}
\begin{equation}
\begin{aligned}
    p^{umap}_{j|i}(x_i, x_j) &= \text{exp} \left( (-d(x_i, x_j)^2 + \rho_{i}) /\tau_i \right) \\[0.3ex]
    q^{umap}_{ij}(y_i, y_j) &= \left( 1 + a(||y_i - y_j||^2_2)^b \right) ^{-1},
\end{aligned}
\end{equation}
where $d(x_i, x_j)$ is the high-dimensional distance function, $\sigma$ and $\tau$ are point-specific variance scalars\footnote{In practice, we can assume that $2 \sigma_i^2$ is functionally equivalent to $\tau_i$, as they are both chosen such that the entropy of the resulting distribution is equivalent.}, $\rho_i = \min_{l \neq i} d(x_i, x_l)$, and $a$ and $b$ are constants. Note that the tSNE denominators in Equation~\ref{eq:tsne_prob} are the sums of all the numerators. We thus refer to tSNE's similarity functions as being \emph{normalized} while UMAP's are \emph{unnormalized}.

The high-dimensional $p$ values are defined with respect to the point in question and are subsequently symmetrized. WLOG, let $p_{ij} = S(p_{j|i}, p_{i|j})$ for some symmetrization function $S$. Going forward, we write $p_{ij}$ and $q_{ij}$ without the superscripts when the normalization setting is clear from the context.

Given these pairwise similarities in the high- and low-dimensional spaces, tSNE and UMAP attempt to find the embedding $Y$ such that $Q(Y)$ is closest to $P(X)$. Since both similarity measures carry a probabilistic interpretation, we find an embedding by minimizing the KL divergence $KL(P \| Q)$. This gives us:
\begin{align}
\label{eq:losses}
    \mathcal{L}_{tsne} &= \sum_{i \neq j} p_{ij} \log \dfrac{p_{ij}}{q_{ij}} \\
    \mathcal{L}_{umap} &= \sum_{i \neq j} p_{ij} \log \dfrac{p_{ij}}{q_{ij}} + (1 - p_{ij}) \log \dfrac{1 - p_{ij}}{1 - q_{ij}}
\end{align}
In essence, tSNE minimizes the KL divergence of the entire pairwise similarity matrix since its $P$ and $Q$ matrices sum to $1$. UMAP instead defines Bernoulli probability distributions $\{p_{ij}, 1-p_{ij}\}, \{q_{ij}, 1 - q_{ij}\}$ and sums the KL divergences between the $n^2$ pairwise probability distributions \footnote{Both tSNE and UMAP set the diagonals of $P$ and $Q$ to $0$}.

\begin{figure}[htb]
    \includegraphics[width=.95\linewidth]{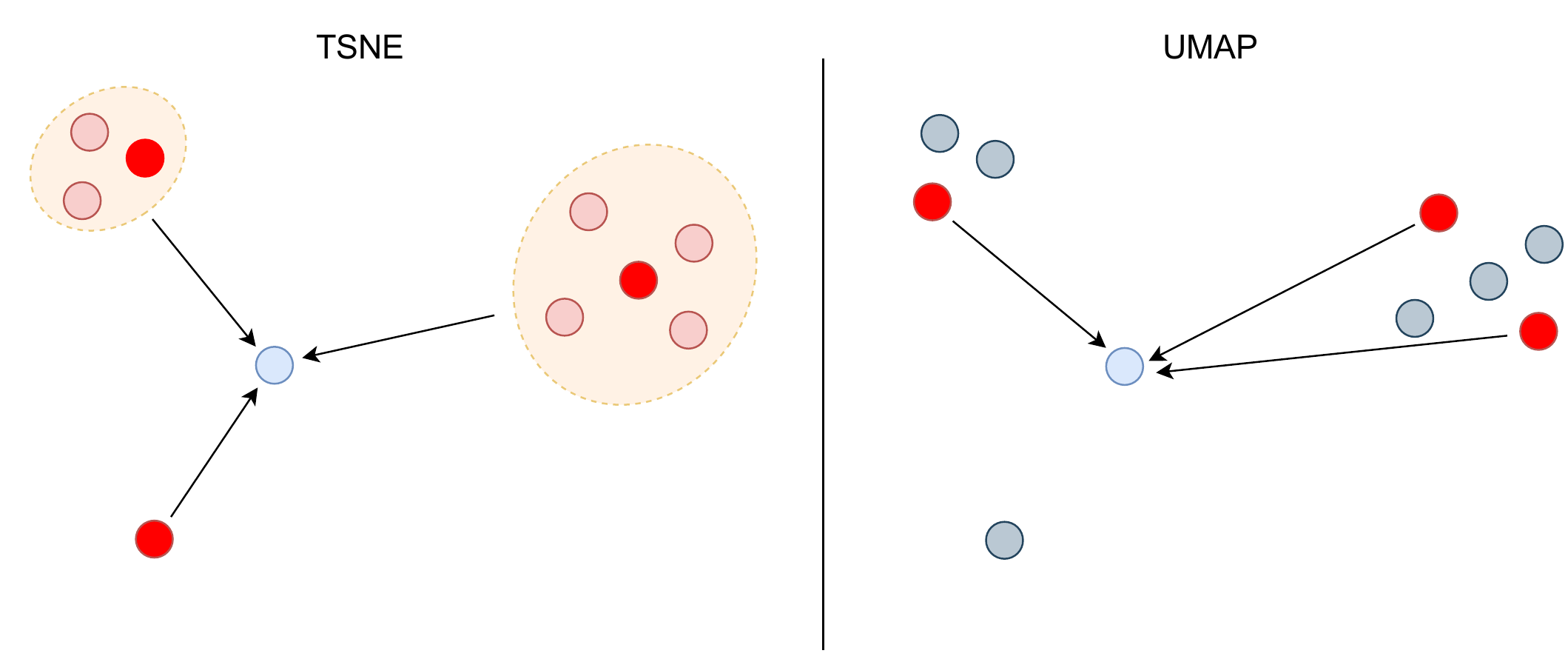}
    \caption{Visualization of the repulsive forces in tSNE (left) and UMAP (right). tSNE calculates the repulsion for representative points and uses this as a proxy for nearby points, giving $\bigO(n)$ total repulsions acting on each point. UMAP calculates the repulsion to a pre-defined number of points and ignores the others, giving $\bigO(1)$ per-point repulsions. Bright red points are those for which the gradient is calculated; arrows are the direction of repulsion.}
    \label{repulsion_vis}
\end{figure}

\begin{figure}[ht]
    \hspace*{-0.9cm}
    \begin{tabular}{p{2.7cm}p{2.7cm}p{2.7cm}}
        \;\;\;\;\;\;\;\;\;\;\;\;tSNE & \;\;\;\;\;\;\;\;\;\;\;\;UMAP & \;\;\;\;\;\;\;\;Frob-UMAP \\
        \includegraphics[width=3.55cm]{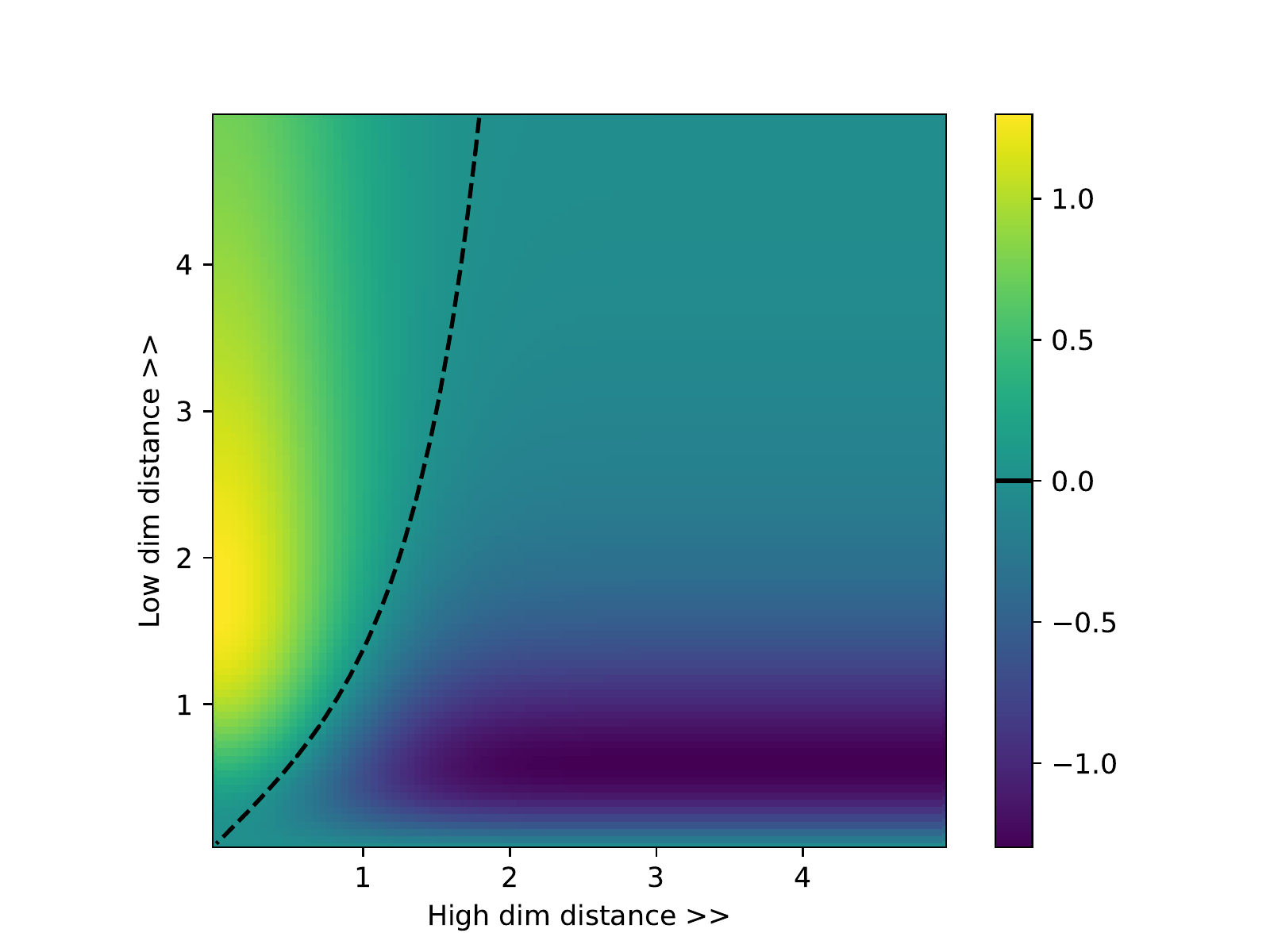} &
        \includegraphics[width=3.55cm]{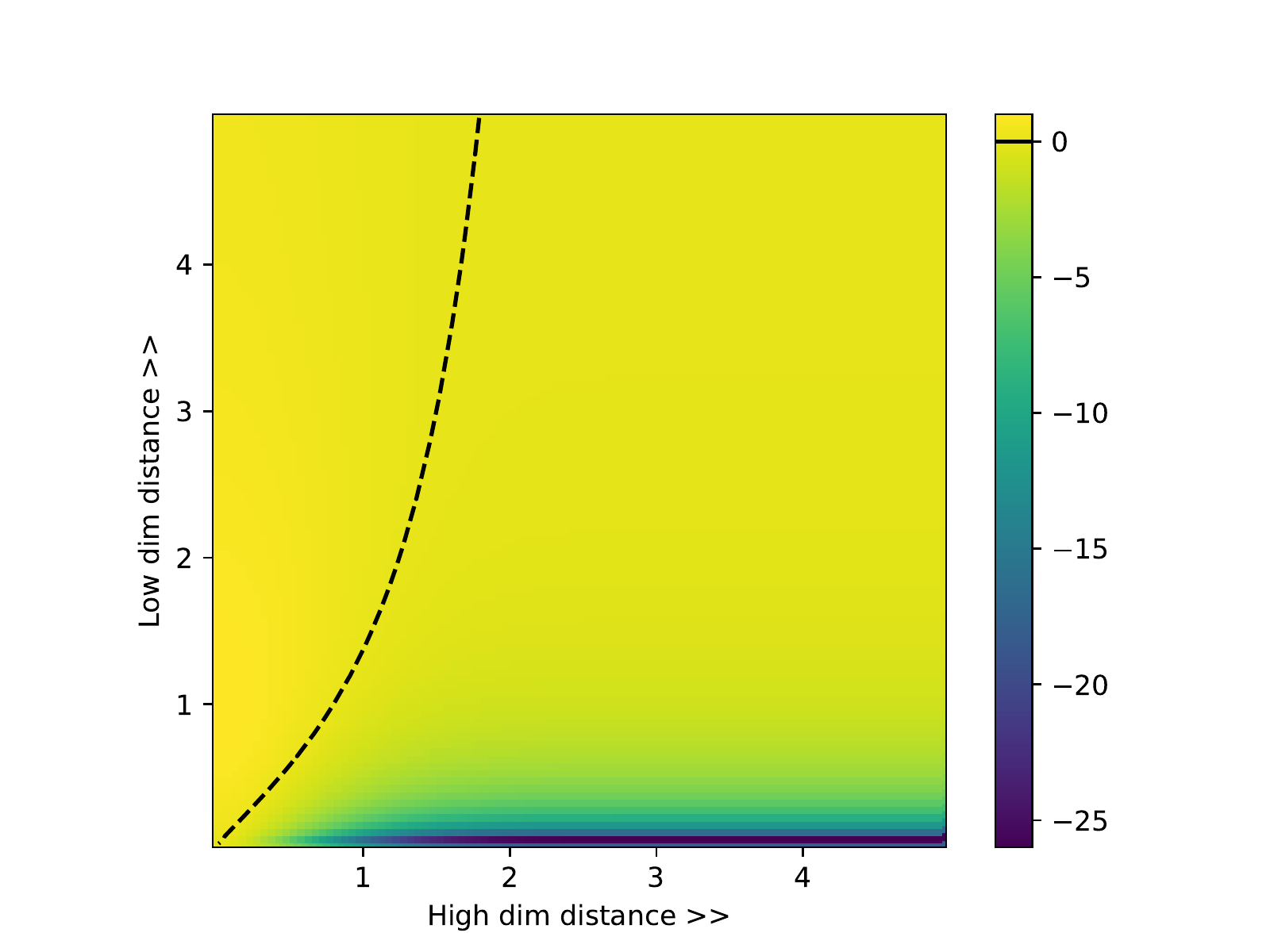} &
        \includegraphics[width=3.55cm]{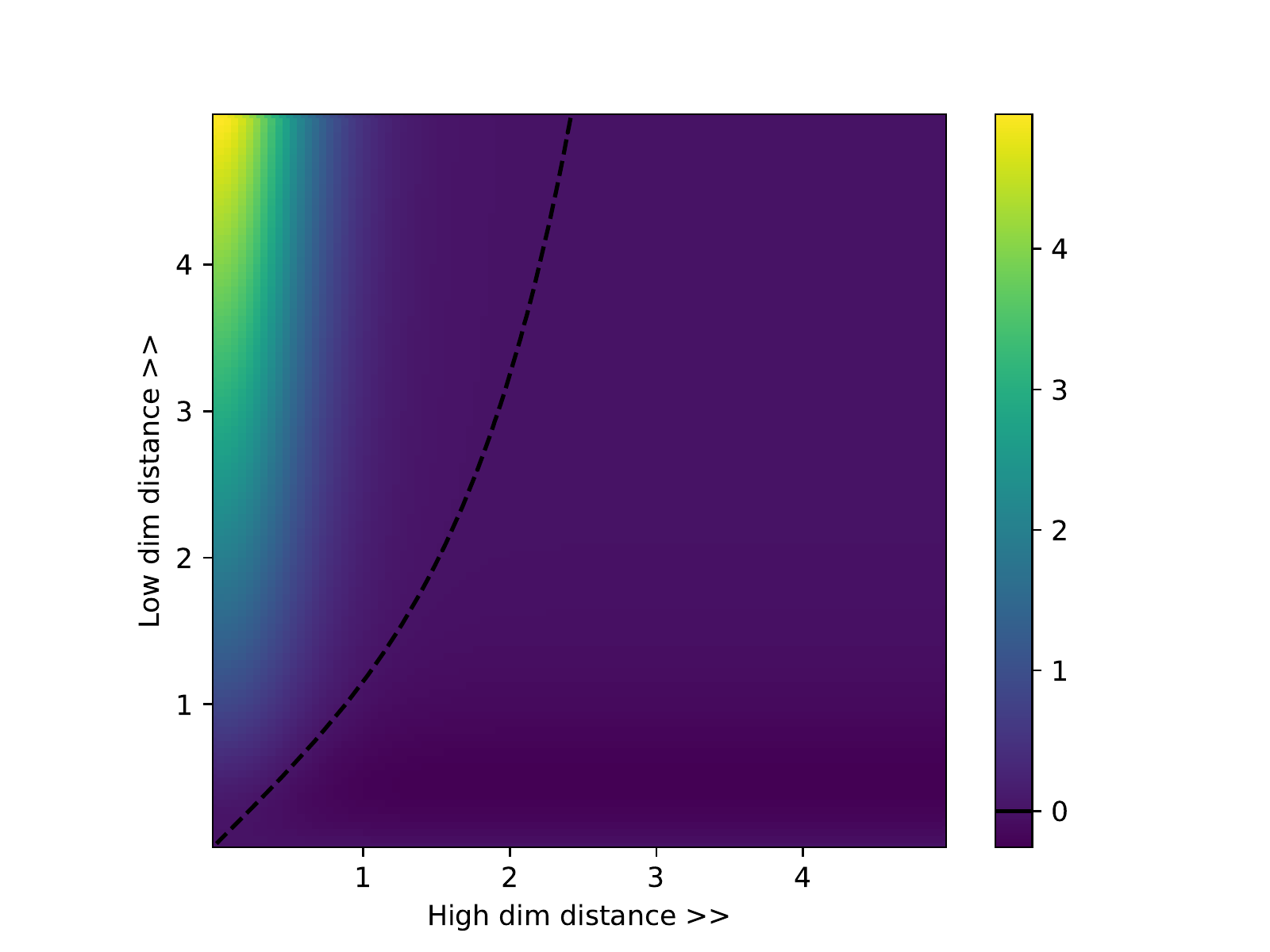}
    \end{tabular}
\caption{Gradient relationships between high- and low-dimensional distances for tSNE, UMAP, and UMAP under the Frobenius norm. The dotted line represents the locations of magnitude-$0$ gradients. Higher values correspond to attractions while lower values correspond to repulsions. The left image is a recreation of the original gradient plot in \protect\cite{van2008visualizing}.}
\label{grad_plots}
\end{figure}

\subsection{Gradient Calculations}
\label{grad_calc_sec}

We now describe and analyze the gradient descent approaches in tSNE and UMAP. First, notice that the gradients of each algorithm change substantially due to the differing normalizations. In tSNE, the gradient can be written as an attractive $\mathcal{A}^{tSNE}_i$ and a repulsive $\mathcal{R}^{tsne}_i$ force acting on point $y_i$ with
{\small
\begin{align}
    \label{tsne_grad_equations}
    \dfrac{\partial \mathcal{L}_{tsne}}{\partial y_i} &
        = -4Z \left[ \sum_{j, j \neq i} p_{ij}q_{ij} (y_i - y_j) - \sum_{k, k \neq i} q_{ik}^2 (y_i - y_k) \right]\\
    &   
        = 4Z (\mathcal{A}_i^{tsne} + \mathcal{R}_i^{tsne}) \nonumber 
\end{align}}
where $Z$ is the normalization term in $q_{ij}^{tSNE}$. On the other hand, UMAP's attractions and repulsions\footnote{The $\varepsilon$ value is only inserted for numerical stability} are presented as \cite{mcinnes2018umap}
\begin{align}
    \mathcal{A}_i^{umap} = & \sum_{j, j \neq i} \dfrac{-2ab\|y_i - y_j\|_2^{2(b-1)}}{1 + \|y_i - y_j\|_2^2} p_{ij} (y_i - y_j) \label{umap_attr} \\
    \mathcal{R}_i^{umap} = & \sum_{k, k \neq i} \dfrac{2b}{\varepsilon + \|y_i - y_k\|_2^2} q_{ik} (1 - p_{ik}) (y_i - y_k). \label{umap_rep}
\end{align}
In the setting where $a=b=1$ and $\varepsilon=0$, Equations~\ref{umap_attr},~\ref{umap_rep} can be written as\footnote{We derive this in section \ref{ssec:umap_derivation} in the supplementary material}
\begin{equation}
\begin{aligned}
    \label{umap_reformulated}
    \mathcal{A}^{umap}_i &= -2 \sum_{j,j \neq i} p_{ij} q_{ij} (y_i - y_j) \\
    \mathcal{R}^{umap}_i &= 2 \sum_{k,k \neq i} q_{ik}^2 \frac{1 - p_{ik}}{1 - q_{ik}}(y_i - y_k)
\end{aligned}
\end{equation}
We remind the reader that we are overloading notation -- $p$ and $q$ are normalized when they are in the tSNE setting and are unnormalized in the UMAP setting.

In practice, tSNE and UMAP optimize their loss functions by iteratively applying these attractive and repulsive forces. It is unnecessary to calculate each such force to effectively estimate the gradient, however, as the $p_{ij}$ term in both the tSNE and UMAP attractive forces decays exponentially. Based on this observation, both methods establish a nearest neighbor graph in the high-dimensional space, where the edges represent nearest neighbor relationships between $x_i$ and $x_j$. It then suffices to only perform attractions between points $y_i$ and $y_j$ if their corresponding $x_i$ and $x_j$ are nearest neighbors.

This logic does not transfer to the repulsions, however, as the Student-t distribution has a heavier tail so repulsions must be calculated evenly across the rest of the points. tSNE does this by fitting a Barnes-Hut tree across $Y$ during \textit{every epoch.} If $y_k$ and $y_l$ are both in the same tree leaf then we assume $q_{ik} = q_{il}$, allowing us to only calculate $\bigO(\log(n))$ similarities. Thus, tSNE estimates all $n-1$ repulsions by performing one such estimate for each cell in $Y$'s Barnes-Hut tree. UMAP, on the other hand, simply obtains repulsions by sampling a constant number of points uniformly and only applying those repulsions. These repulsion schemas are depicted in Figure~\ref{repulsion_vis}. To apply the gradients, tSNE collects all of them and performs momentum gradient descent across the entire dataset whereas UMAP moves each point immediately upon calculating its forces.

There are a few differences between the two algorithms' gradient descent loops. First, the tSNE learning rate stays constant over training while UMAP's linearly decreases. Second, tSNE's gradients are strengthened by adding a ``gains'' term which scales gradients based on whether they point in the same direction from epoch to epoch\footnote{This term has not been mentioned in the literature but is present in common tSNE implementations.}. We refer to these two elements as \textit{gradient amplification}.

Note that UMAP's repulsive force has a $1 - p_{ik}$ term that is unavailable at runtime, as $x_i$ and $x_k$ may not have been nearest neighbors. In practice, UMAP estimates these $1 - p_{ik}$ terms by using the available $p_{ij}$ values\footnote{When possible, we use index $k$ to represent repulsions and $j$ to represent attractions to highlight that $p_{ik}$ is never calculated in UMAP. See Section~\ref{ssec:forces_estimate} in the supplementary material for details.}. We also note that UMAP does not explicitly multiply by $p_{ij}$ and $1 - p_{ik}$. Instead, it samples the forces proportionally to these scalars. For example, if $p_{ij} = 0.1$ then we apply that force \emph{without the $p_{ij}$ multiplier} once every ten epochs. We refer to this as \textit{scalar sampling}.

\subsection{The Choice of Normalization}
\label{ssec:norm_discussion}
We now present a summary of our theoretical results before providing their formal statements. As was shown in ~\cite{bohm2020unifying}, the ratio between attractive and repulsive magnitudes determines the structure of the resulting embedding. Given this context, Theorem~\ref{thm:norm-changes-ratio} shows that the normalization directly changes the ratio of attraction/repulsion magnitudes, inducing the difference between tSNE and UMAP embeddings. Thus, we can toggle the normalization to alternate between their outputs. Furthermore, Theorem~\ref{thm:sampling-unnecessary} shows that the attraction/repulsion ratio  in the normalized setting \emph{is independent of} the number of repulsive samples collected. This second point allows us to accelerate tSNE to UMAP speeds without impacting embedding quality by simply removing the dependency on Barnes-Hut trees and calculating $1$ per-point repulsion as in UMAP. We now provide the necessary definitions for the theorems.

Assume that the $p_{ij}$ terms are given. We now consider the dataset $Y$ probabilistically by defining a set of random variables $v_{ij} = y_i - y_j$ and assume that all $\bigO(n^2)$ $v_{ij}$ vectors are i.i.d. around a non-zero mean. Let $r_{ij} = (1 + |v_{ij}|^2)^{-1}$ and define $Z = \sum_{i, j}^{n^2} r_{ij}$ as the sum over $n^2$ pairs of points and $\tilde{Z} = \sum_{i, j}^{n} r_{ij}$ as the sum over $n$ pairs of points. Then applying $n$ per-point repulsions gives us the force acting on point $y_i$ of $\mathbb{E}[|\mathcal{R}^{tsne}|] = \mathbb{E}[\sum_{j}^n ||(r_{ij}^2 / Z^2) \cdot v_{ij}||]$. We now define an equivalent force term in the setting where we have $1$ per-point repulsion: $\mathbb{E}[|\tilde{\mathcal{R}}^{tsne}|] = \mathbb{E}[||(r_{ij}^2 / \tilde{Z}^2) \cdot v_{ij}||]$. Note that we have a constant number $c$ of attractive forces acting on each point, giving $\mathbb{E}[|\mathcal{A}^{tsne}|] = c \cdot p_{ij}^{tsne} \mathbb{E}[||(r_{ij} / Z) \cdot v_{ij}||]$ and $\mathbb{E}[|\tilde{\mathcal{A}}^{tsne}|] = c \cdot p_{ij}^{tsne} \mathbb{E}[||(r_{ij} / \tilde{Z}) \cdot v_{ij}||]$.

Thus, $|\mathcal{A}^{tsne}|$ and $|\mathcal{R}^{tsne}|$ represent the magnitudes of the forces when we calculate tSNE's $\bigO(n)$ per-point repulsions while $|\tilde{\mathcal{A}}^{tsne}|$ and $|\tilde{\mathcal{R}}^{tsne}|$ represent the forces when we have UMAP's $\bigO(1)$ per-point repulsions. Given this, we have the following theorems:

\begin{theorem}
\label{thm:norm-changes-ratio}
Let $p_{ij}^{tsne} \sim 1/(cn)$ and $d(x_i, x_j) > \sqrt{ \log(n^2 + 1) \tau }$. Then
$ \dfrac{\mathbb{E}[|\mathcal{A}_i^{umap}|]}{\mathbb{E}[|\mathcal{R}_i^{umap}|]} < \dfrac{\mathbb{E}\left[|\tilde{\mathcal{A}}_i^{tsne}|\right]}{\mathbb{E}\left[|\tilde{\mathcal{R}}_i^{tsne}|\right]} $.
\end{theorem}

\begin{theorem}
\label{thm:sampling-unnecessary}
$\dfrac{\mathbb{E}\left[|\mathcal{A}_i^{tsne}|\right]}{\mathbb{E}\left[|\mathcal{R}_i^{tsne}|\right]} = \dfrac{\mathbb{E}\left[|\tilde{\mathcal{A}}_i^{tsne}|\right]}{\mathbb{E}\left[|\tilde{\mathcal{R}}_i^{tsne}|\right]}$
\end{theorem}

The proofs are given in Sections \ref{prf:norm-changes-ratio} and \ref{prf:sampling-unnecessary} of the supplementary material. We point out that $p_{ij}^{tsne}$ is normalized over the sum of all $cn$ attractions that are sampled, giving us the estimate $p_{ij}^{tsne} \sim 1/(cn)$.

Theorem~\ref{thm:norm-changes-ratio}'s result is visualized in the gradient plots in Figure~\ref{grad_plots}. There we see that the UMAP repulsions can be orders of magnitude larger than the corresponding tSNE ones, even when accounting for the magnitude of the attractions. Furthermore, Section~\ref{results} evidences that toggling the normalization is sufficient to switch between the algorithms' embeddings and that no other hyperparameter choice accounts for the difference in inter-cluster distances between tSNE and UMAP.

\newcommand{\dcell}[1]{\small{\it\color{gray}\makecell{#1}}}

\begin{table*}[thb]
    \centering
    \begin{tabularx}{\textwidth}{lccccc}
    & \textbf{Initialization} & \textbf{Distance function} & \textbf{Symmetrization} & \textbf{Sym Attraction} & \textbf{Scalars} \\
     & \dcell{$Y$ initialization} & \dcell{High-dim \\ distances calculation} & \dcell{Setting {$p_{ij} = p_{ji}$}} & \dcell{Attraction$(y_i, y_j)$\\ applied to both} & \dcell{Values for \\ $a$ and $b$} \\
    \midrule
    
    tSNE & Random & $d(x_i, x_j)$ & $(p_{i|j} + p_{j|i})/2$ & No & $a = 1$, $b = 1$\\
    
    UMAP & Lapl. Eigenmap & $d(x_i, x_j) -\min_k d(x_i, x_k)$ & $p_{i|j}{+}p_{j|i}{-}p_{i|j}p_{j|i}$ & Yes & Grid search \\
    \bottomrule
    \end{tabularx}
    \caption{List of differences between hyperparameters of tSNE and UMAP. These are analyzed in Figures~\ref{irrelevant-metrics-row-means}, ~\ref{irrelevant-mnist}, ~\ref{irrelevant-fashion-mnist}, ~\ref{irrelevant-metrics_col_means}. }
    \label{differences_table}
\end{table*}


\section{Unifying tSNE and UMAP}
\label{ssec:ourmethod}
This leads us to \ourmethod -- a modification to UMAP that can recreate both tSNE and UMAP embeddings at UMAP speeds. We choose the general name Gradient Dimensionality Reduction to imply that it is \textit{both} UMAP and tSNE.

\begin{table*}
    \newcolumntype{C}{ >{\centering\arraybackslash} m{2.4cm} }
    \newcolumntype{D}{ >{\centering\arraybackslash} m{.01cm} }
    \hspace*{-0.25cm}
    \begin{tabular}{DCCCCCC}
    & \quad MNIST & \quad MNIST & Fashion-MNIST & Fashion-MNIST & \quad Swiss Roll & \quad Swiss Roll \\
    \\[-1.2em]

    \rotatebox{90}{tSNE} &
    \includegraphics[width=2.7cm]{outputs/mnist/tsne/default_embedding.png}&
    \includegraphics[width=2.7cm]{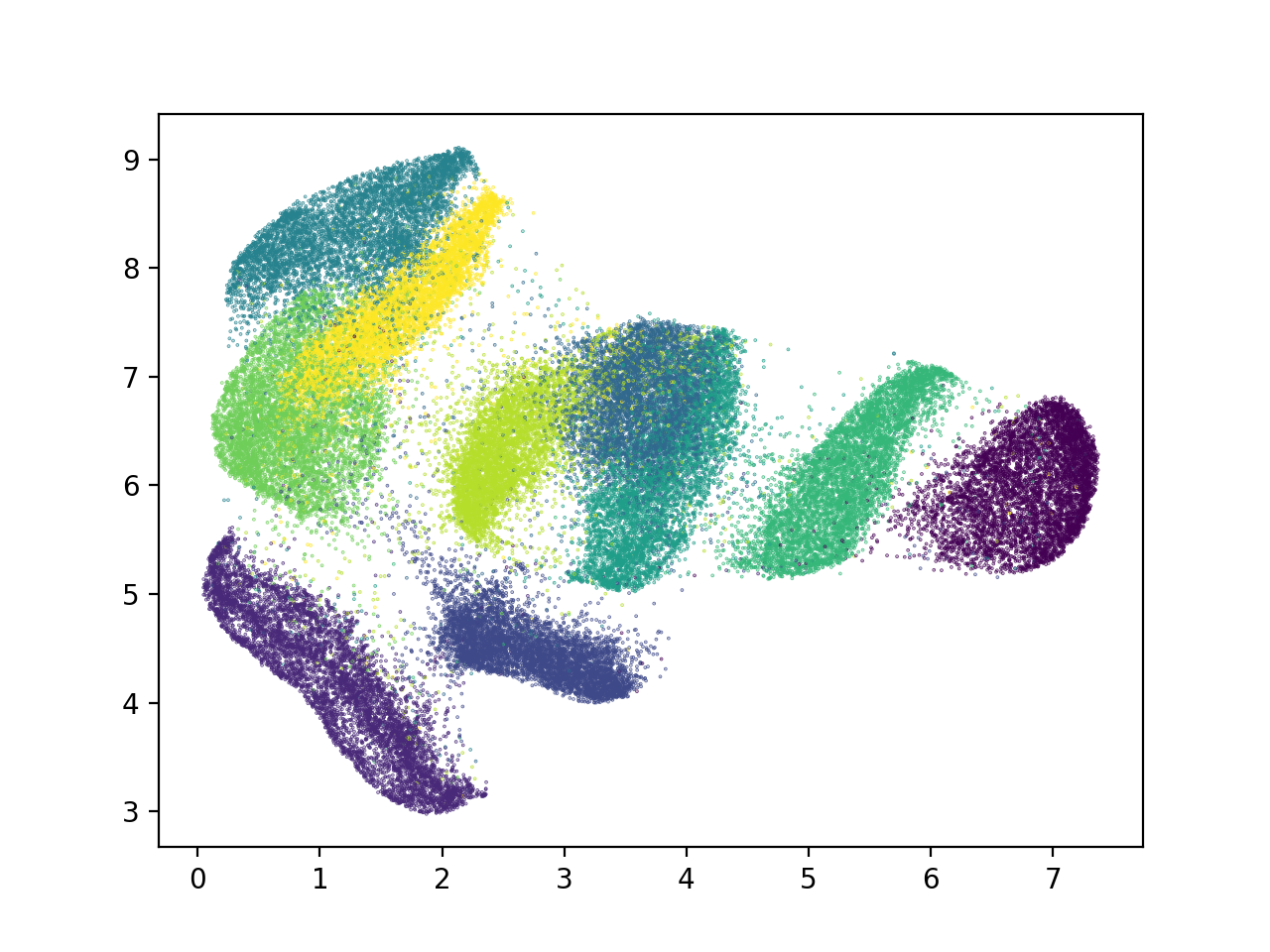}&
    \includegraphics[width=2.7cm]{outputs/fashion_mnist/tsne/default_embedding.png}&
    \includegraphics[width=2.7cm]{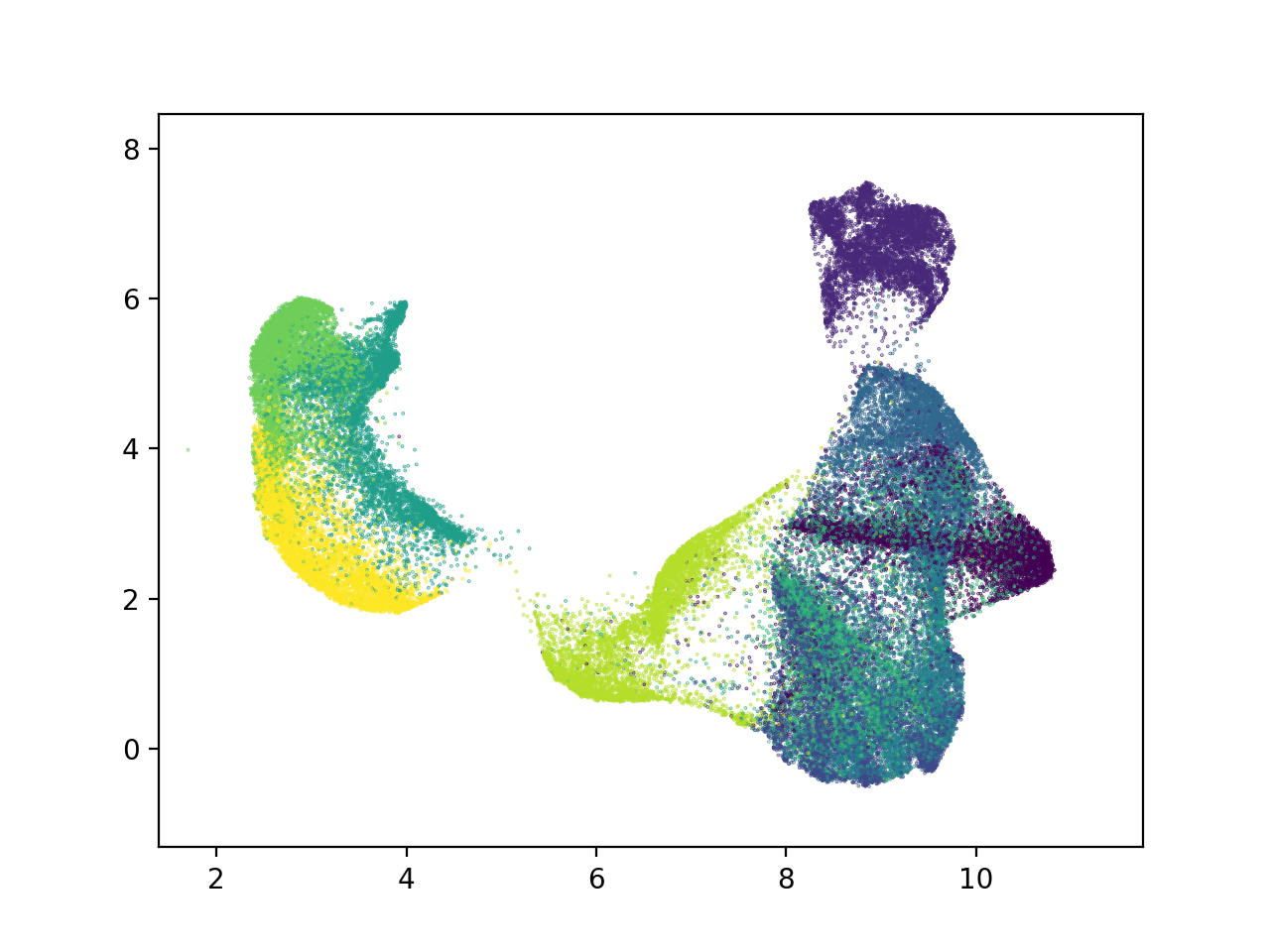}&
    \includegraphics[width=2.7cm]{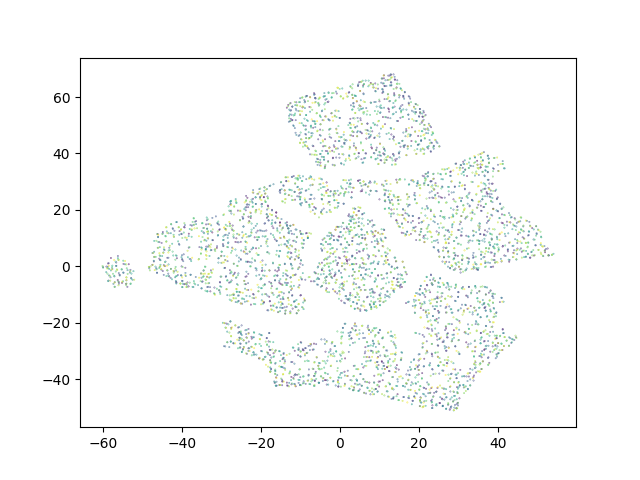}&
    \includegraphics[width=2.7cm]{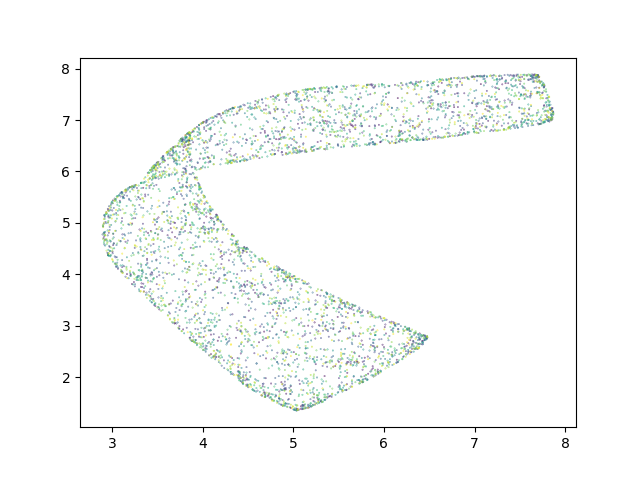}\\
    \\[-1.2em]

    \rotatebox{90}{UMAP} &
    \includegraphics[width=2.7cm]{outputs/mnist/uniform_umap/normalized_embedding.png}&
    \includegraphics[width=2.7cm]{outputs/mnist/umap/default_embedding.png}&
    \includegraphics[width=2.7cm]{outputs/fashion_mnist/uniform_umap/normalized_embedding.png}&
    \includegraphics[width=2.7cm]{outputs/fashion_mnist/umap/default_embedding.png}&
    \includegraphics[width=2.7cm]{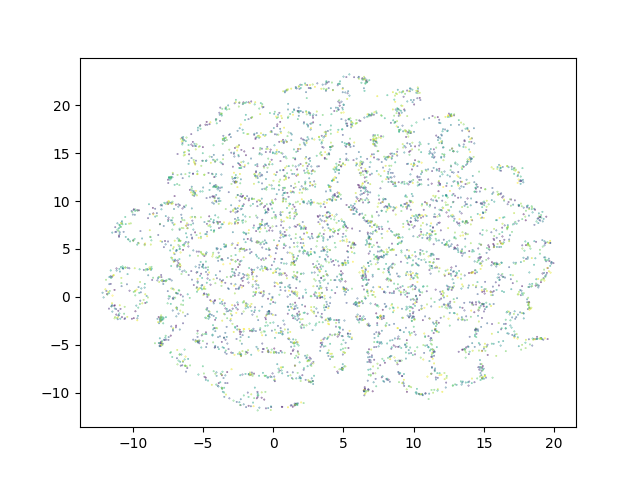}&
    \includegraphics[width=2.7cm]{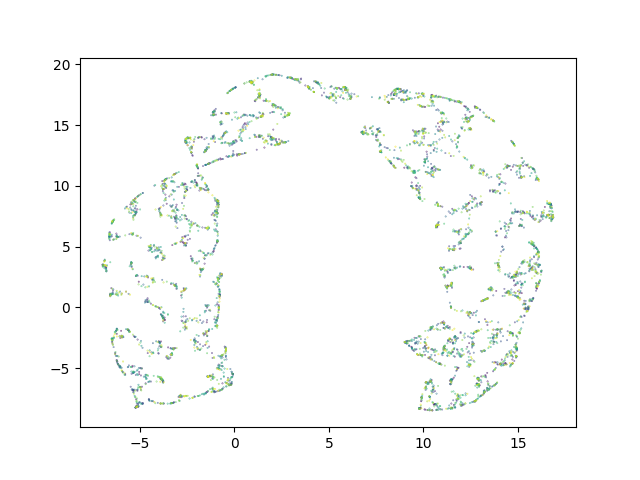}\\
    \\[-1.2em]
    \end{tabular}
    \caption{Effect of changing the normalization for the original tSNE and UMAP algorithms on the MNIST, Fashion-MNIST, and Swiss Roll datasets. Each dataset is shown with normalization followed by no normalization. We use Laplacian Eigenmap initializations for consistent orientation. The normalized UMAP plots were made with the changes described in section \ref{ssec:norm_results}.}
    \label{relevant-mnist}
\end{table*}

Our algorithm follows the UMAP optimization procedure except that we \textbf{(1)} replace the \textit{scalar sampling} by iteratively processing attractions/repulsions and \textbf{(2)} apply the gradients after having collected all of them, rather than immediately upon processing each one. The first change accommodates the gradients under normalization since the normalized repulsive forces do not have the $1 - p_{ik}$ term to which UMAP samples proportionally. The second change allows for performing momentum gradient descent for faster convergence in the normalized setting. 

Since we follow the UMAP optimization procedure, \ourmethod defaults to producing UMAP embeddings. In the case of replicating tSNE, we simply normalize the $P$ and $Q$ matrices and scale the learning rate. Although we only collect $\bigO(1)$ attractions and repulsions for each point, their magnitudes are balanced due to Theorems~\ref{thm:norm-changes-ratio} and~\ref{thm:sampling-unnecessary}. We refer to \ourmethod as \ourmethodU if it is in the unnormalized setting and as \ourmethodN if it is in the normalized setting. We note that changing the normalization necessitates gradient amplification.

By allowing \ourmethod to toggle the normalization, we are free to choose the simplest options across the other parameters. \ourmethod therefore defaults to tSNE's asymmetric attraction and $a$ and $b$  scalars along with UMAP's distance-metric, initialization, nearest neighbors, and $p_{ij}$ symmetrization.

The supplementary material provides some further information on the flexibility of \ourmethod (\ref{sssec:accelerated}), such as an accelerated version of the algorithm where we modify the gradient formulation such that it is quicker to optimize. This change induces a consistent $2\times$ speedup of \ourmethod over UMAP. Despite differing from the true KL divergence gradient, we find that the resulting embeddings are comparable. Our repository also provides a CUDA kernel that calculates \ourmethodU and \ourmethodN embeddings in a distributed manner on a GPU.

\subsection{Theoretical Considerations}
\label{ssec:theoretical_implications}
UMAP's theoretical framework identifies the existence of a locally-connected manifold in the high-dimensional space under the UMAP pseudo-distance metric $\tilde{d}$. This pseudo-distance metric is defined such that the distance from point $x_j$ to $x_i$ is equal to $\tilde{d}(x_i, x_j) {=} d(x_i, x_j) {-} \min_{l \neq i} d(x_i, x_l)$. Despite this being a key element of the UMAP foundation, we find that substituting the Euclidean distance for the pseudo-distance metric seems to have no effect on the embeddings, as seen in Tables~\ref{irrelevant-metrics-row-means} and~\ref{irrelevant-mnist}. It is possible that the algorithm's reliance on highly non-convex gradient descent deviates enough from the theoretical discussion that the pseudo-distance metric loses its applicability. It may also be the case that this pseudo-distance metric, while insightful from a theoretical perspective, is not a necessary calculation in order to achieve the final embeddings.

Furthermore, many of the other differences between tSNE and UMAP are not motivated by the theoretical foundation of either algorithm. The gradient descent methodology is entirely heuristic, so any differences therein do not impact the theory. This applies to the repulsion and attraction sampling and gradient descent methods. Moreover, the high-dimensional symmetrization function, embedding initialization, symmetric attraction, and $a, b$ scalars can all be switched to their alternative options without impacting either method's consistency within its theoretical presentation. Thus, each of these heuristics can be toggled without impacting the embedding's interpretation, as most of them do not interfere with the theory and none affect the output.

We also question whether the choice of normalization is necessitated by either algorithm's presentation. tSNE, for example, treats the normalization of $P$ and $Q$ as an assumption and provides no further justification. In the case of UMAP, it appears that the normalization does not break the assumptions of the original paper~\cite[Sec.~2,3]{mcinnes2018umap}. We therefore posit that the interpretation of UMAP as finding the best fit to the high-dimensional data manifold extends to tSNE as well, as long as tSNE's gradients are calculated under the pseudo-distance metric in the high-dimensional space. We additionally theorize that each method can be paired with either normalization without contradicting the foundations laid out in its paper.

We evidence the fact that tSNE can preserve manifold structure at least as well as UMAP in Table~\ref{relevant-mnist}, where Barnes-Hut tSNE without normalization cleanly maintains the structure of the Swiss Roll dataset. We further discuss these manifold learning claims in the supplementary material (\ref{ssec:manifold_learning}). 

For all of these reasons, we make the claim that tSNE and UMAP are computationally consistent with one another. That is, we conjecture that, up to minor changes, one could have presented UMAP's theoretical foundation and implemented it with the tSNE algorithm or vice-versa.

\subsection{Frobenius Norm for UMAP}
Finally, even some of the standard algorithmic choices can be modified without significantly impacting the embeddings. For example, UMAP and tSNE both optimize the KL divergence, but we see no reason that the Frobenius norm cannot be substituted in its place. Interestingly, the embeddings in Figure~\ref{fig:frob_embeddings}  in the supplementary material show that optimizing the Frobenius norm in the unnormalized setting produces outputs that are indistinguishable from the ones obtained by minimizing the KL-divergence. To provide a possible indication as to why this occurs, Figure~\ref{grad_plots} shows that the zero-gradient areas between the KL divergence and the Frobenius norm strongly overlap, implying that a local minimum under one objective satisfies the other one as well.

We bring this up for two reasons. First, the Frobenius norm is a significantly simpler loss function to optimize than the KL divergence due to its convexity. We hypothesize that there must be simple algorithmic improvements that can exploit this property. Further detail is given in Section~\ref{ssec:frob_norm} in the supplementary material. Second, it is interesting to consider that even fundamental assumptions such as the objective function can be changed without significantly affecting the embeddings across datasets.
\section{Results} \label{results}

\begin{table}
    \newcolumntype{C}{>{\centering\arraybackslash} m{1.245cm}}
    \newcolumntype{M}{>{\centering\arraybackslash} m{0.1cm}}
    \newcolumntype{N}{>{\centering\arraybackslash} m{1.08cm}}
    \hspace*{-0.25cm}
    \begin{tabular}{MNCCCC}
    & & \makecell{Fashion \\ MNIST} & \makecell{Coil\\100} & \makecell{Single \\ Cell} & Cifar-10\\

    \midrule
    \multirow{4}{*}{\rotatebox[origin=c]{90}{\textbf{kNN Acc.}}} 
    & UMAP & 78.0; 0.5 & 80.8; 3.3 & 43.4; 1.9 & 24.2; 1.1 \\ 
    & \ourcell \ourmethodU & \ourcell 77.3; 0.7 & \ourcell 77.4; 3.4 & \ourcell 42.8; 2.2 & \ourcell 23.8; 1.1 \\
    \cmidrule(lr){2-6}
    & tSNE & 80.1; 0.7 & 63.2; 4.2 & 43.3; 1.9 & 28.7; 2.5 \\ 
    & \ourcell \ourmethodN & \ourcell 78.6; 0.6 & \ourcell 77.2; 4.4 & \ourcell 44.8; 1.4 & \ourcell 25.6; 1.1 \\
    \midrule
    
    \multirow{4}{*}{\bf \rotatebox[origin=c]{90}{V-score\;\;}} 
    & UMAP & 60.3; 1.4 & 89.2; 0.9 & 60.6; 1.3 & 7.6; 0.4\\ 
    & \ourcell \ourmethodU & \ourcell 61.7; 0.8 & \ourcell 91.0; 0.6 & \ourcell 60.1; 1.6 & \ourcell 8.1; 0.6 \\
    \cmidrule(lr){2-6}
    & tSNE & 54.2; 4.1 & 82.9; 1.8 & 59.7; 1.1 & 8.5; 0.3 \\ 
    & \ourcell \ourmethodN & \ourcell 51.7; 4.7 & \ourcell 85.7; 2.6 & \ourcell 60.5; 0.8 & \ourcell 8.0; 3.7
    \end{tabular}
    \caption{Row means and std. deviations for kNN-accuracy and V-score on Fashion MNIST, Coil-100, Single-Cell, and Cifar-10 datasets. For example, the cell [Fashion-MNIST, kNN accuracy, tSNE] implies that the mean kNN accuracy across the hyperparameters in Table~\ref{differences_table} was 80.1 for tSNE on the Fashion-MNIST dataset.}
    \label{irrelevant-metrics-row-means}
\end{table}

\label{ssec:irrelevant_alg_params}
\textbf{Metrics.} There is no optimal way to compare embeddings as performing an analysis at the point-level loses global information while studying macro-structures loses local information. To account for this, we employ separate metrics to study the embeddings at the micro- and macro-scales. Specifically, we use the $k$NN accuracy to analyze preservation of local neighborhoods as established in~\cite{van2009dimensionality} and the V-measure~\cite{rosenberg2007v}, a standard tool for evaluating cluster preservation, to study the embedding's global cluster structures\footnote{We provide formalization of these metrics in the supplementary material (\ref{ssec:metrics})}.

\spara{Hyperparameter effects}. We first show that a majority of the differences between tSNE and UMAP do not significantly affect the embeddings. Specifically, Table~\ref{irrelevant-mnist} shows that we can vary the hyperparameters in Table~\ref{differences_table} with negligible change to the embeddings of any discussed algorithm. Equivalent results on other datasets can be found in Tables ~\ref{irrelevant-mnist}~and~\ref{irrelevant-metrics_col_means} in the supplementary material. Furthermore, Table~\ref{irrelevant-metrics-row-means} provides quantitative evidence that the hyperparameters do not affect the embeddings across datasets; similarly, Table~\ref{irrelevant-metrics_col_means} in the supplementary material confirms this finding across algorithms.

\begin{table*}[!thb]
    \setlength{\tabcolsep}{6pt}
    \centering
    \begin{tabular}{p{0.25cm}*{6}{>{\centering\arraybackslash}p{2.4cm}}}
    & \textbf{Default setting} & Random init & Pseudo distance & Symmetrization & Sym attraction & a, b scalars \\

    \\[-1.25em]
    \rotatebox[origin=l]{90}{\bf \;\; tSNE} & 
    \includegraphics[width=2.5cm]{outputs/mnist/tsne/default_embedding.png}&
    \includegraphics[width=2.5cm]{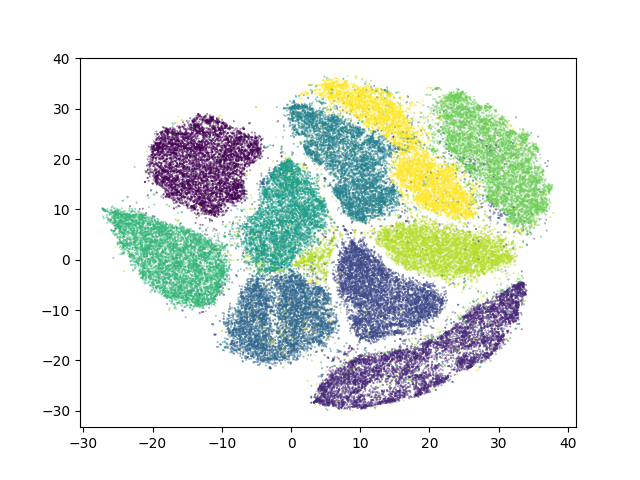}&
    \includegraphics[width=2.5cm]{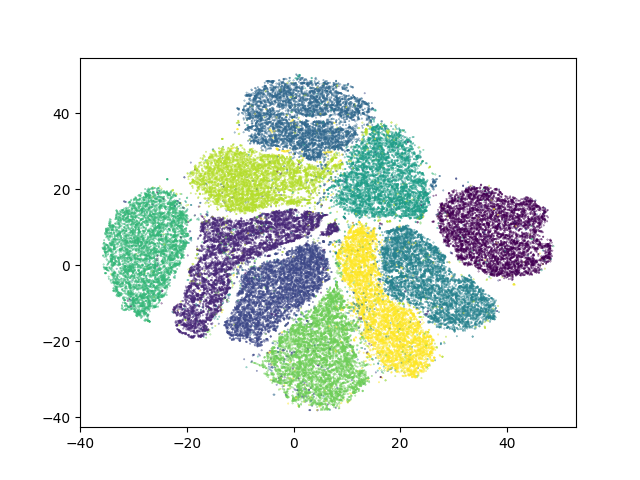} & 
    \includegraphics[width=2.5cm]{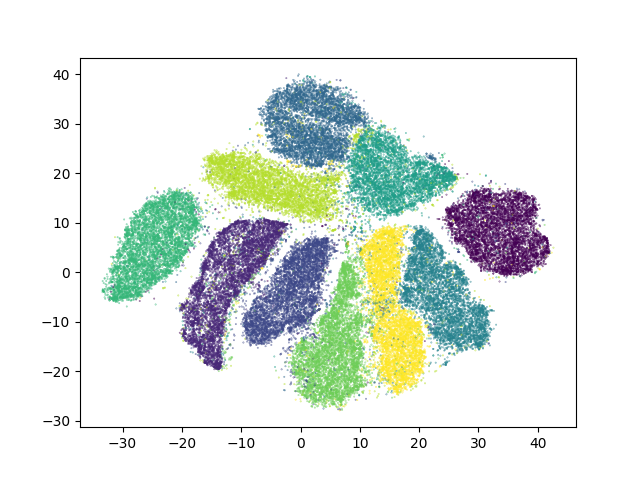}&
    \includegraphics[width=2.5cm]{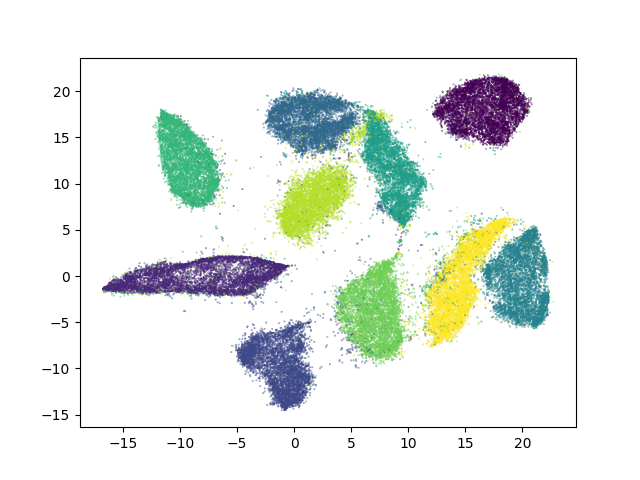}&
    \includegraphics[width=2.5cm]{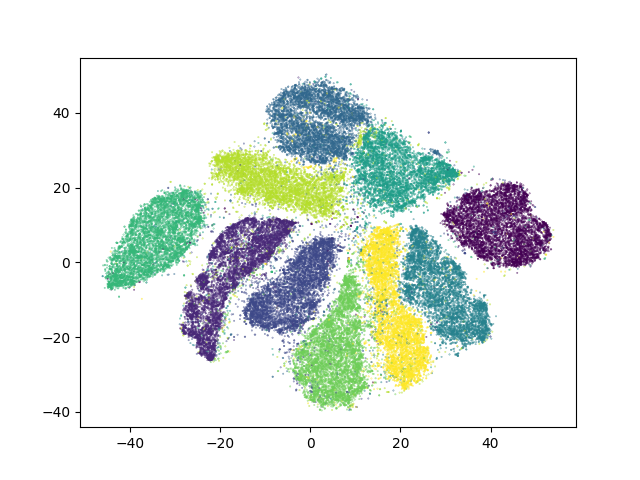} \\
    \\[-1.5em]
    
    & 
    \;95.1; 70.9 & \;95.2; 70.7 & \;96.0; 73.9 & \;94.9; 70.8 & \;94.8; 80.7 & \;95.1; 73.2 \\
    \\[-1em]

    {\ourcell}\rotatebox[origin=l]{90}{\bf \;\;\ourmethodN} & 
    \includegraphics[width=2.5cm]{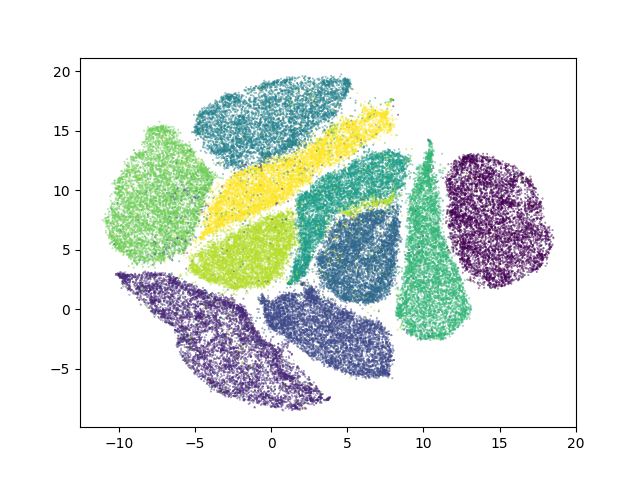}&
    \includegraphics[width=2.5cm]{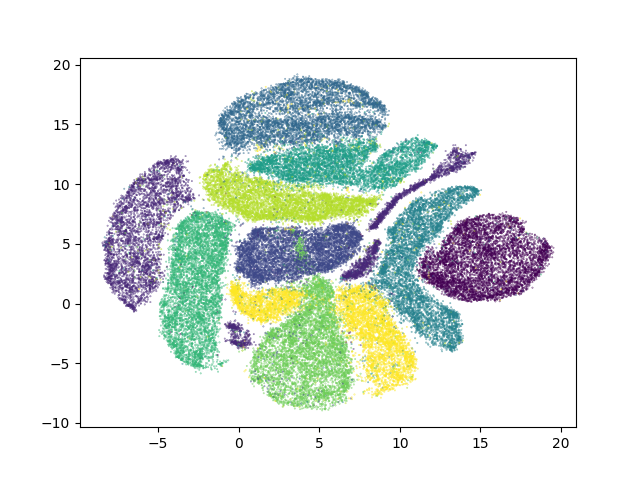}&
    \includegraphics[width=2.5cm]{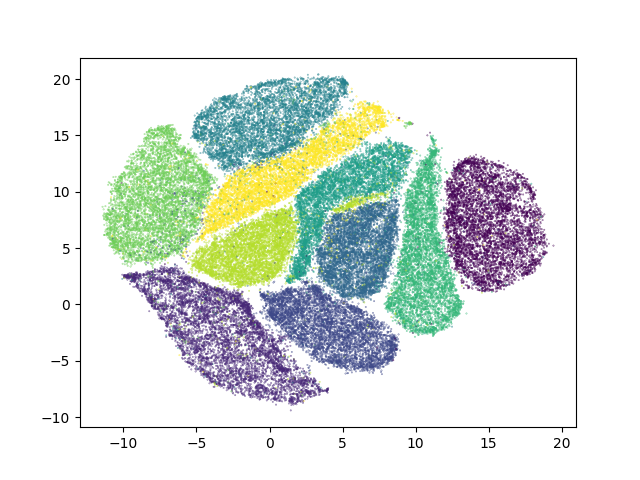}& 
    \includegraphics[width=2.5cm]{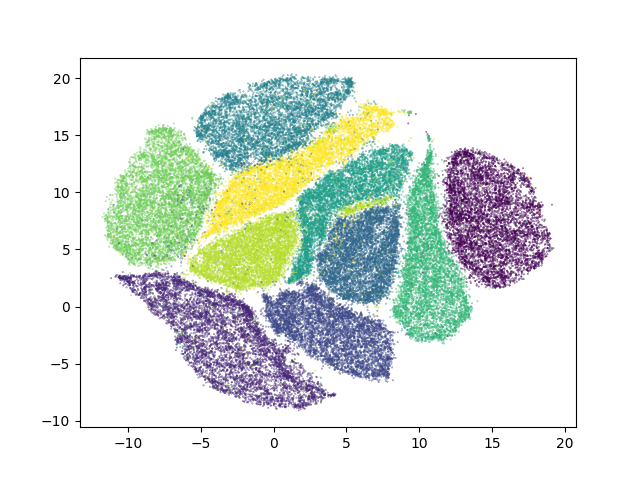}&
    \includegraphics[width=2.5cm]{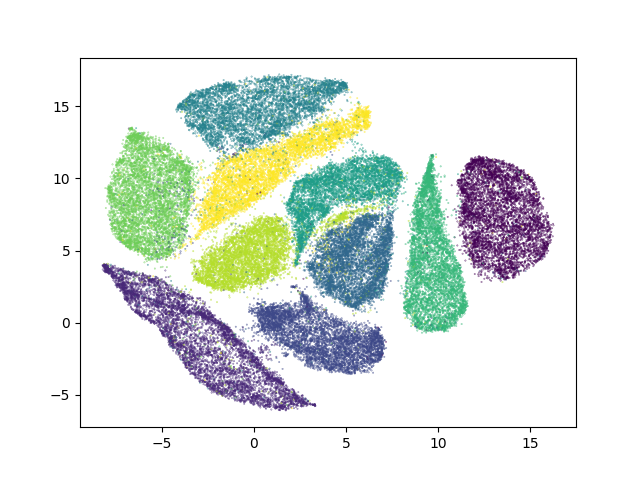}&
    \includegraphics[width=2.5cm]{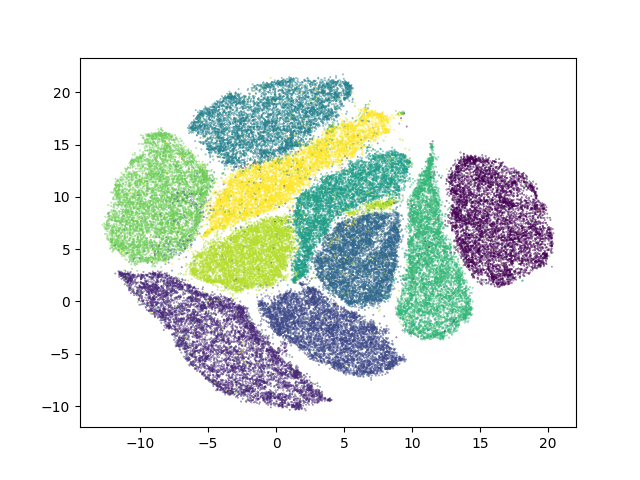} \\
    \\[-1.5em]
    
    &
    \;96.1; 67.8 & \;95.6; 61.3 & \;96.1; 63.0 & \;96.1; 68.4 & \;96.3; 72.7 & \;96.1; 68.8 \\
    \\[-1em]

    \rotatebox[origin=l]{90}{\bf \;\; UMAP} & 
    \includegraphics[width=2.5cm]{outputs/mnist/umap/default_embedding.png}&
    \includegraphics[width=2.5cm]{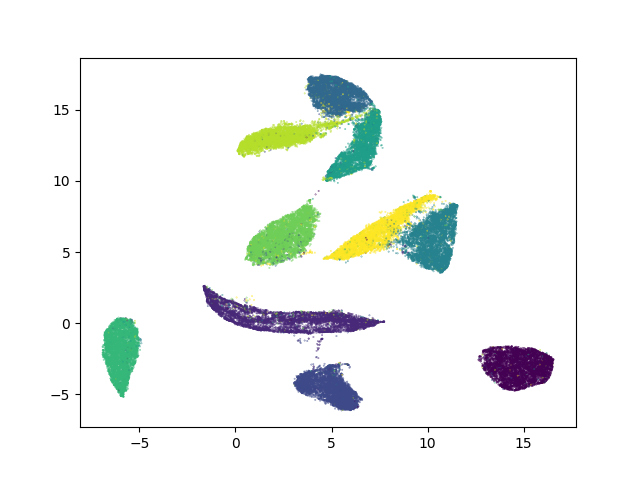}&
    \includegraphics[width=2.5cm]{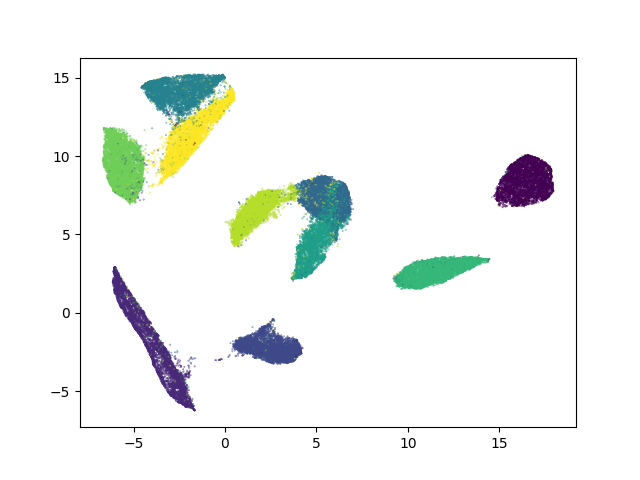}& 
    \includegraphics[width=2.5cm]{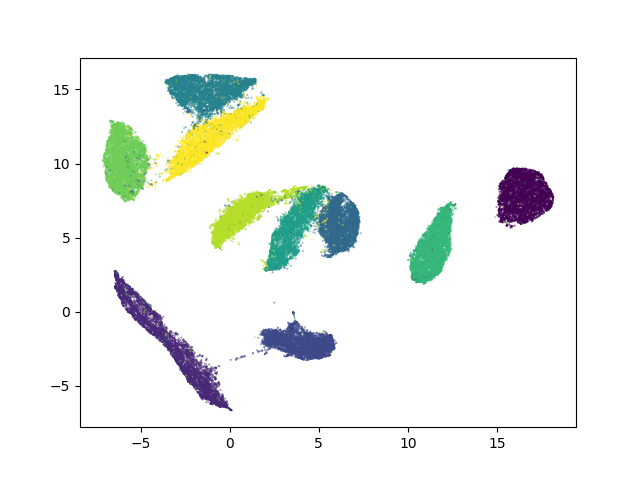}&
    \includegraphics[width=2.5cm]{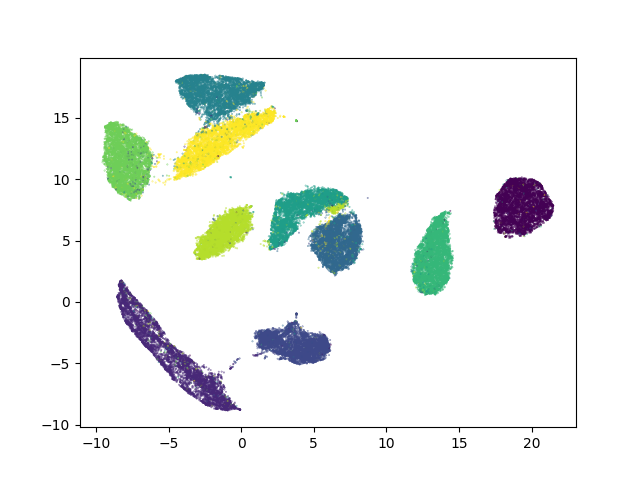}&
    \includegraphics[width=2.5cm]{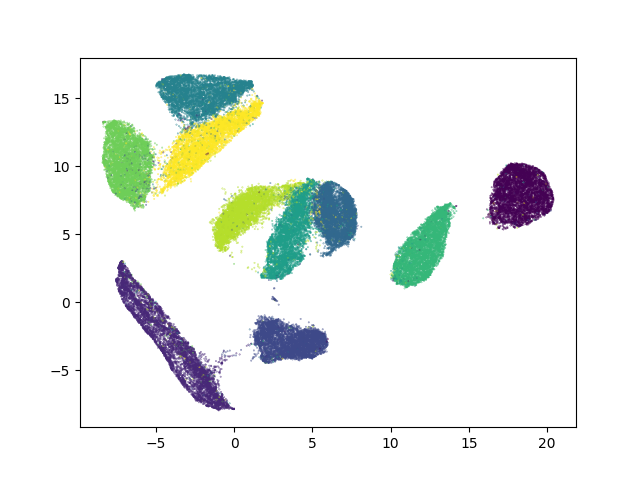} \\
    \\[-1.5em]
    & 
    \;95.4; 82.5 & \;96.6; 84.6 & \;94.4; 82.2 & \;96.7; 82.5 & \;96.6; 83.5 & \;96.5; 82.2 \\
    \\[-1em]

    {\ourcell}\rotatebox[origin=l]{90}{\bf \, \ourmethodU} &  
    \includegraphics[width=2.5cm]{outputs/mnist/uniform_umap/default_embedding.png}&
    \includegraphics[width=2.5cm]{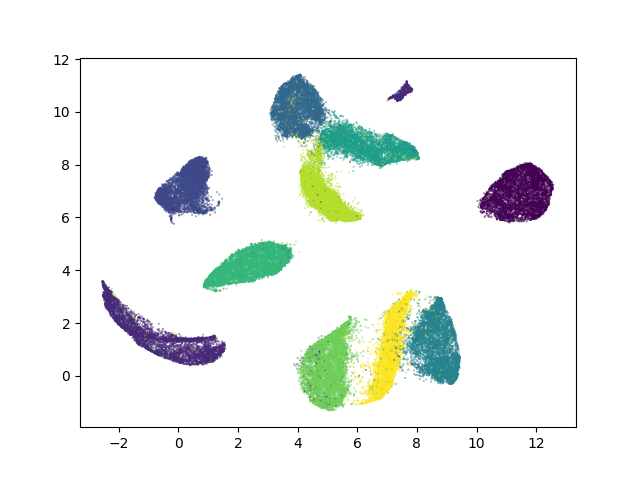}&
    \includegraphics[width=2.5cm]{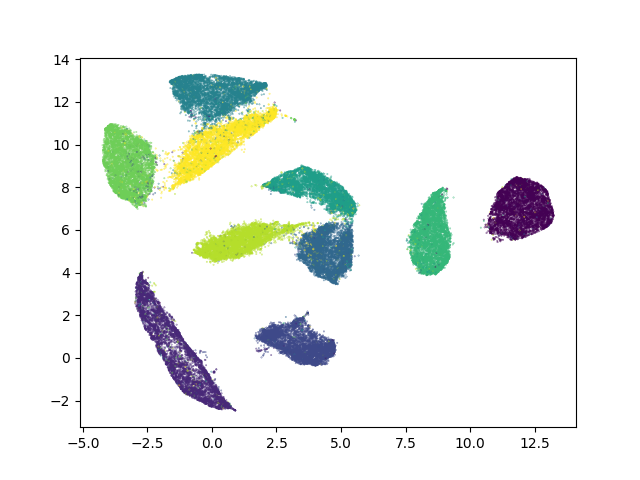}& 
    \includegraphics[width=2.5cm]{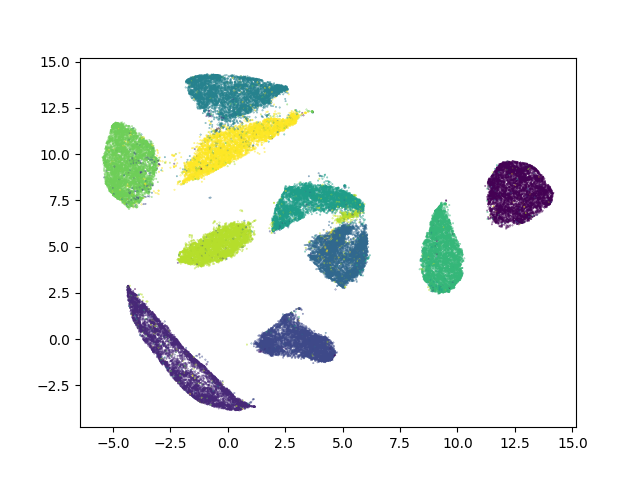}&
    \includegraphics[width=2.5cm]{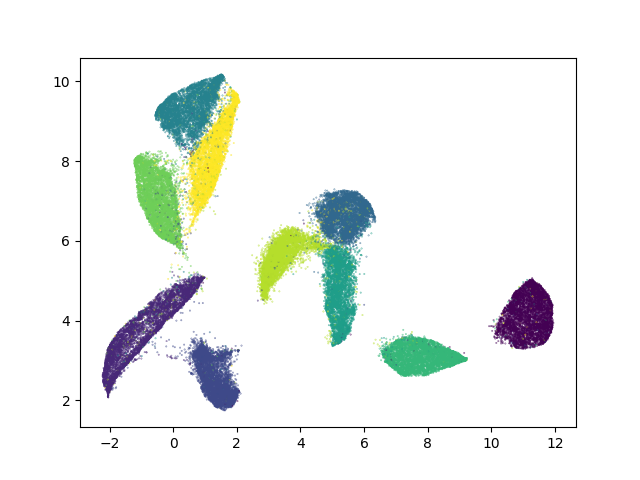}&
    \includegraphics[width=2.5cm]{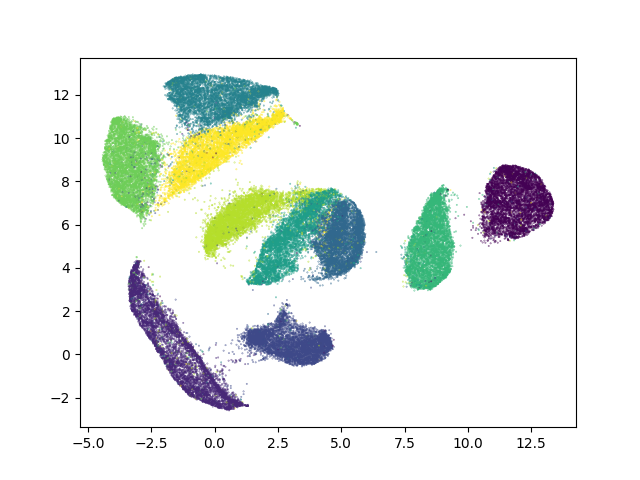} \\
    \\[-1.5em]
    & 
    \;96.2; 84.0 & \;96.4; 82.1 & \;96.7; 85.2 & \;96.6; 85.1 & \;96.5; 83.3 & \;95.8; 81.2\\
    \\[-1em]

    \end{tabular}
    \caption{Effect of the algorithm settings from Table~\ref{differences_table} on MNIST
    dataset. Each parameter is changed from its default to its alternative setting; e.g., the random init column implies that tSNE was initialized with Laplacian Eigenmaps while UMAP and \ourmethod randomly.
    Below each image the KNN-accuracy and K-Means V-score show unchanged performance.}
    \label{irrelevant-mnist}
\end{table*}

Looking at Table~\ref{irrelevant-mnist}, the initialization and the symmetric attraction induce the largest variation in the embeddings. For the initialization, the relative positions of clusters change but the relevant inter-cluster relationships remain consistent\footnote{As such, we employ the Laplacian Eigenmap initialization on small datasets (${<}100$K) due to its predictable output and the random initialization on large datasets (${>}100$K) to avoid slowdowns.}. Enabling symmetric attraction attracts $y_j$ to $y_i$ when we attract $y_i$ to $y_j$. Thus, switching from asymmetric to symmetric attraction functionally scales the attractive force by 2. This leads to tighter tSNE clusters that would otherwise be evenly spread out across the embedding, but does not affect UMAP significantly. We thus choose asymmetric attraction for \ourmethod as it better recreates tSNE embeddings.

We show the effect of \emph{single} hyperparameter changes for combinatorial reasons. However, we see no significant difference between changing one hyperparameter or any number of them. We also eschew including hyperparameters that have no effect on the embeddings and are the least interesting. These include the exact vs. approximate nearest neighbors, gradient clipping, and the number of epochs.

\spara{Effect of Normalization}. \label{ssec:norm_results}
Although Theorem~\ref{thm:sampling-unnecessary} shows that we can take fewer repulsive samples without affecting the repulsion's magnitude, we must also verify that the angle of the repulsive force is preserved as well. Towards this end, we plot the average angle between the tSNE Barnes-Hut repulsions and the UMAP sampled repulsions in Figure~\ref{grad_agreement}. We see that, across datasets, the direction of the repulsion remains consistent throughout the optimization process. Thus, since both the magnitude and the direction are robust to the number of samples taken, we conclude that one can obtain tSNE embeddings with UMAP’s $\bigO(1)$ per-point repulsions.

\input{Figures/gradient_agreement}

We now show that toggling the normalization allows tSNE to simulate UMAP embeddings and vice versa. Table~\ref{relevant-mnist} shows exactly this. First note that tSNE in the unnormalized setting has significantly more separation between clusters in a manner similar to UMAP. The representations are fuzzier than the UMAP ones as we are still estimating $\bigO(n)$ repulsions, causing the embedding to fall closer to the mean of the multi-modal datasets. To account for the $n\times$ more repulsions, we scale each repulsion by $1/n$ for the sake of convergence. This is a different effect than normalizing by $\sum p_{ij}$ as we are not affecting the attraction/repulsion ratio in Theorem~\ref{thm:norm-changes-ratio}.

The analysis is slightly more involved in the case of UMAP. Recall that the UMAP algorithm approximates the $p_{ij}$ and $1 - p_{ik}$ gradient scalars by sampling the attractions and repulsions proportionally to $p_{ij}$ and $1 - p_{ik}$, which we referred to as \textit{scalar sampling}. However, the gradients in the normalized setting (Equation~\ref{tsne_grad_equations}) lose the $1 - p_{ik}$ scalar on repulsions. The UMAP optimization schema, then, imposes an unnecessary weight on the repulsions in the normalized setting as the repulsions are still sampled according to the no-longer-necessary $1 - p_{ik}$ scalar. Accounting for this requires dividing the repulsive forces by $1 - p_{ik}$, but this (with the momentum gradient descent and stronger learning rate) leads to a highly unstable training regime. We refer the reader to Table~\ref{normed_umap} in the supplementary material for details.

This implies that stabilizing UMAP in the normalized setting requires removing the sampling and instead directly multiplying by $p_{ij}$ and $1 - p_{ik}$. Indeed, this is exactly what we do in \ourmethod. Under this change, \ourmethodU and \ourmethodN obtain effectively identical embeddings to the default UMAP and tSNE ones. This is confirmed in the kNN accuracy and K-means V-score metrics in Table~\ref{irrelevant-metrics-row-means}.

\spara{Time efficiency}. We lastly discuss the speeds of UMAP, tSNE, \ourmethod, and our accelerated version of \ourmethod in section \ref{ssec:runtime_analysis} of the supplementary material due to space concerns. Our implementations of UMAP and \ourmethod perform gradient descent an order of magnitude faster than the standard UMAP library, implying a corresponding speedup over tSNE. We also provide an acceleration by doing \ourmethod \emph{with} scalar sampling that provides a further $2\times$ speedup. Despite the fact that this imposes a slight modification onto the effective gradients, we show that this is qualitatively insignificant in the resulting embeddings.
\section{Conclusion \& Future Work}
\label{sec:conclusion}
We discussed the set of differences between tSNE and UMAP and identified that only the normalization significantly impacts the outputs. This provides a clear unification of tSNE and UMAP that is both theoretically simple and easy to implement. Beyond this, our analysis has uncovered multiple misunderstandings regarding UMAP and tSNE while hopefully also clarifying how these methods work.

We raised several questions regarding the theory of gradient-based DR algorithms. Namely, we believe that many assumptions can be revisited. Is there a setting in which the UMAP pseudo-distance changes the embeddings? Does the KL divergence induce a better optimization criterium than the Frobenius norm? Is it true that UMAP's framework can accommodate tSNE's normalization? We hope that we have facilitated future research into the essence of these algorithms through identifying all of their algorithmic components and consolidating them in a simple-to-use codebase.

\bibliographystyle{named}
\bibliography{references}

\begin{thebibliography}{}

\bibitem[\protect\citeauthoryear{Belkin and Niyogi}{2003}]{belkin2003laplacian}
Mikhail Belkin and Partha Niyogi.
\newblock Laplacian eigenmaps for dimensionality reduction and data
  representation.
\newblock {\em Neural computation}, 15(6):1373--1396, 2003.

\bibitem[\protect\citeauthoryear{Bohm \bgroup \em et al.\egroup
  }{2020}]{bohm2020unifying}
Jan~Niklas Bohm, Philipp Berens, and Dmitry Kobak.
\newblock A unifying perspective on neighbor embeddings along the
  attraction-repulsion spectrum.
\newblock {\em arXiv preprint arXiv:2007.08902}, 2020.

\bibitem[\protect\citeauthoryear{Damrich and Hamprecht}{2021}]{damrich2021umap}
Sebastian Damrich and Fred~A Hamprecht.
\newblock On umap's true loss function.
\newblock {\em Advances in Neural Information Processing Systems}, 34, 2021.

\bibitem[\protect\citeauthoryear{Damrich \bgroup \em et al.\egroup
  }{2022}]{damrich2022contrastive}
Sebastian Damrich, Jan~Niklas B{\"o}hm, Fred~A Hamprecht, and Dmitry Kobak.
\newblock Contrastive learning unifies $ t $-sne and umap.
\newblock {\em arXiv preprint arXiv:2206.01816}, 2022.

\bibitem[\protect\citeauthoryear{Deng}{2012}]{lecun-mnisthandwrittendigit-2010}
Li~Deng.
\newblock The mnist database of handwritten digit images for machine learning
  research.
\newblock {\em IEEE Signal Processing Magazine}, 29(6):141--142, 2012.

\bibitem[\protect\citeauthoryear{Dong \bgroup \em et al.\egroup
  }{2011}]{dong2011efficient}
Wei Dong, Charikar Moses, and Kai Li.
\newblock Efficient k-nearest neighbor graph construction for generic
  similarity measures.
\newblock In {\em Proceedings of the 20th international conference on World
  wide web}, pages 577--586, 2011.

\bibitem[\protect\citeauthoryear{{Hull}}{1994}]{uspsdataset}
J.~J. {Hull}.
\newblock A database for handwritten text recognition research.
\newblock {\em IEEE Transactions on Pattern Analysis and Machine Intelligence},
  16(5):550--554, 1994.

\bibitem[\protect\citeauthoryear{Kobak and
  Linderman}{2021}]{kobak2021initialization}
Dmitry Kobak and George~C Linderman.
\newblock Initialization is critical for preserving global data structure in
  both t-sne and umap.
\newblock {\em Nature biotechnology}, 39(2):156--157, 2021.

\bibitem[\protect\citeauthoryear{Krizhevsky}{2009}]{krizhevsky2009learning}
Alex Krizhevsky.
\newblock Learning multiple layers of features from tiny images.
\newblock {\em Technical Report, University of Toronto}, 2009.

\bibitem[\protect\citeauthoryear{Linderman \bgroup \em et al.\egroup
  }{2019}]{linderman2019fast}
George~C Linderman, Manas Rachh, Jeremy~G Hoskins, Stefan Steinerberger, and
  Yuval Kluger.
\newblock Fast interpolation-based t-sne for improved visualization of
  single-cell rna-seq data.
\newblock {\em Nature methods}, 16(3):243--245, 2019.

\bibitem[\protect\citeauthoryear{McInnes \bgroup \em et al.\egroup
  }{2018}]{mcinnes2018umap}
Leland McInnes, John Healy, and James Melville.
\newblock Umap: Uniform manifold approximation and projection for dimension
  reduction.
\newblock {\em arXiv preprint arXiv:1802.03426}, 2018.

\bibitem[\protect\citeauthoryear{Mikolov \bgroup \em et al.\egroup
  }{2013}]{mikolov2013efficient}
Tomas Mikolov, Kai Chen, Greg Corrado, and Jeffrey Dean.
\newblock Efficient estimation of word representations in vector space.
\newblock {\em arXiv preprint arXiv:1301.3781}, 2013.

\bibitem[\protect\citeauthoryear{NENE}{1996}]{nene1996columbia}
SA~NENE.
\newblock Columbia object image library (coil-100).
\newblock {\em Technical Report CUCS-006-96}, 1996.

\bibitem[\protect\citeauthoryear{Rosenberg and
  Hirschberg}{2007}]{rosenberg2007v}
Andrew Rosenberg and Julia Hirschberg.
\newblock V-measure: A conditional entropy-based external cluster evaluation
  measure.
\newblock In {\em EMNLP-CoNLL}, pages 410--420, 2007.

\bibitem[\protect\citeauthoryear{Sainburg \bgroup \em et al.\egroup
  }{2020}]{sainburg2020parametric}
Tim Sainburg, Leland McInnes, and Timothy~Q Gentner.
\newblock Parametric umap embeddings for representation and semi-supervised
  learning.
\newblock {\em arXiv preprint arXiv:2009.12981}, 2020.

\bibitem[\protect\citeauthoryear{Tang \bgroup \em et al.\egroup
  }{2016}]{tang2016visualizing}
Jian Tang, Jingzhou Liu, Ming Zhang, and Qiaozhu Mei.
\newblock Visualizing large-scale and high-dimensional data.
\newblock In {\em TheWebConf}, pages 287--297, 2016.

\bibitem[\protect\citeauthoryear{Tasic \bgroup \em et al.\egroup
  }{2018}]{tasic2018shared}
Bosiljka Tasic, Zizhen Yao, Lucas~T Graybuck, Kimberly~A Smith, Thuc~Nghi
  Nguyen, Darren Bertagnolli, Jeff Goldy, Emma Garren, Michael~N Economo,
  Sarada Viswanathan, et~al.
\newblock Shared and distinct transcriptomic cell types across neocortical
  areas.
\newblock {\em Nature}, 563(7729):72--78, 2018.

\bibitem[\protect\citeauthoryear{Van~der Maaten and
  Hinton}{2008}]{van2008visualizing}
Laurens Van~der Maaten and Geoffrey Hinton.
\newblock Visualizing data using t-sne.
\newblock {\em Journal of machine learning research}, 9(11), 2008.

\bibitem[\protect\citeauthoryear{Van Der~Maaten \bgroup \em et al.\egroup
  }{2009}]{van2009dimensionality}
Laurens Van Der~Maaten, Eric Postma, Jaap Van~den Herik, et~al.
\newblock Dimensionality reduction: a comparative.
\newblock {\em J Mach Learn Res}, 10(66-71):13, 2009.

\bibitem[\protect\citeauthoryear{Van Der~Maaten}{2014}]{van2014accelerating}
Laurens Van Der~Maaten.
\newblock Accelerating t-sne using tree-based algorithms.
\newblock {\em JMLR}, 15(1):3221--3245, 2014.

\bibitem[\protect\citeauthoryear{Wang \bgroup \em et al.\egroup
  }{2021}]{wang2021understanding}
Yingfan Wang, Haiyang Huang, Cynthia Rudin, and Yaron Shaposhnik.
\newblock Understanding how dimension reduction tools work: an empirical
  approach to deciphering t-sne, umap, trimap, and pacmap for data
  visualization.
\newblock {\em The Journal of Machine Learning Research}, 22(1):9129--9201,
  2021.

\bibitem[\protect\citeauthoryear{Xiao \bgroup \em et al.\egroup
  }{2017}]{xiao2017fashion}
Han Xiao, Kashif Rasul, and Roland Vollgraf.
\newblock Fashion-mnist: a novel image dataset for benchmarking machine
  learning algorithms.
\newblock {\em arXiv preprint arXiv:1708.07747}, 2017.

\end{thebibliography}

\clearpage
\appendix

\section{Supplementary Material}

\subsection{Runtime Analysis}
\label{ssec:runtime_analysis}
The difference in speed between UMAP and Barnes-Hut tSNE is almost entirely due to two factors -- the nearest neighbor search and the repulsion sampling. Specifically, UMAP uses nearest neighbor descent \cite{dong2011efficient} to quickly approximate the high-dimensional similarities and then avoids many gradient calculations by sampling according to the $p_{ij}$ and $1 - p_{ik}$ terms. tSNE, on the other hand, finds exact nearest neighbors for the high-dimensional similarities and fits Barnes-Hut trees during every epoch thereafter.

In joining the two methods, we have taken out both of these differences. Thus, when we run \ourmethodN, we are still performing nearest neighbor descent and avoiding the Barnes-Hut trees. Figure \ref{data_size_runtimes} shows that \ourmethodU and \ourmethodN both outperform the available UMAP library by an order of magnitude. The difference in speed between the normalized and unnormalized algorithms comes down to two factors. First, the normalized variant requires about two times more epochs to converge. Second, it must still calculate the $Q$ normalization term, which must be summed over all parallel threads and requires more accesses to shared memory.

For an honest comparison, we also attempted to speed up the original UMAP gradient descent algorithm. A significant portion of this improvement came from changing the parallelization by explicitly handling memory allocations. As seen in Figure \ref{data_size_runtimes}, our UMAP implementation runs each epoch at about the same speed as \ourmethod. However, the accelerated variants of \ourmethod run about two times faster, thus outperforming our fastest UMAP implementation by the same factor of two.

Despite the fact that UMAP only calculates attractions and repulsions with proportional frequency to the scalars, it requires multiple repulsions for each attraction. By instead applying all of the gradients outside of the optimization loop, we can utilize gradient descent methodologies such that we only require one repulsion for each attraction. The end result becomes a similar number of gradient operations to UMAP.

\begin{table}[ht]
    \newcolumntype{C}{ >{\centering\arraybackslash} m{2.2cm} }
    \newcolumntype{D}{ >{\centering\arraybackslash} m{0.01cm} }
    \begin{tabularx}{\textwidth}{DCCC}
        & \quad MNIST & \quad \makecell{Fashion-\\MNIST} & \quad Coil-100 \\
        \rotatebox{90}{\ourmethodN} & 
        \includegraphics[width=2.7cm]{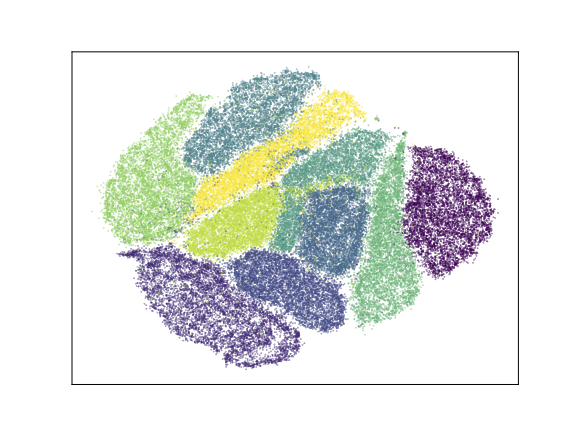} &
        \includegraphics[width=2.7cm]{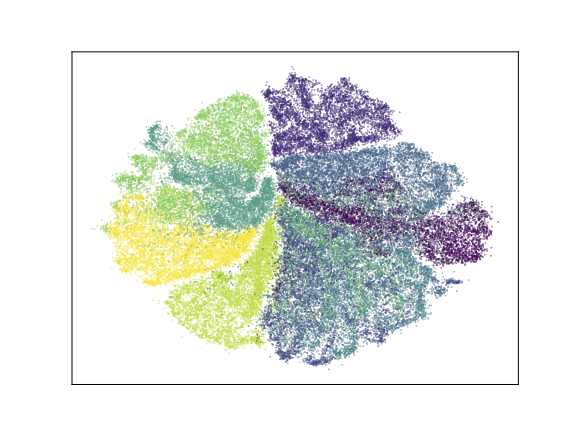} &
        \includegraphics[width=2.7cm]{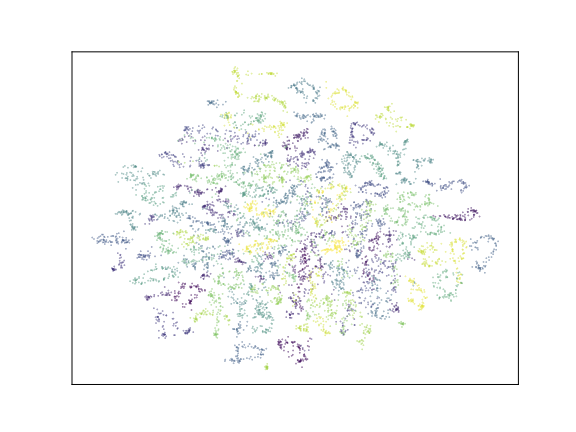} \\
        
        \rotatebox{90}{\ourmethodU} &
        \includegraphics[width=2.7cm]{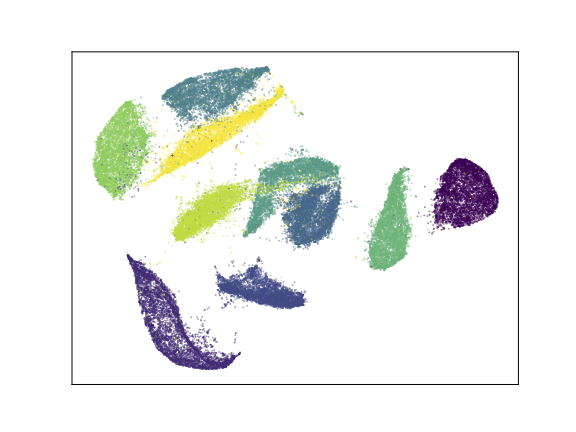} &
        \includegraphics[width=2.7cm]{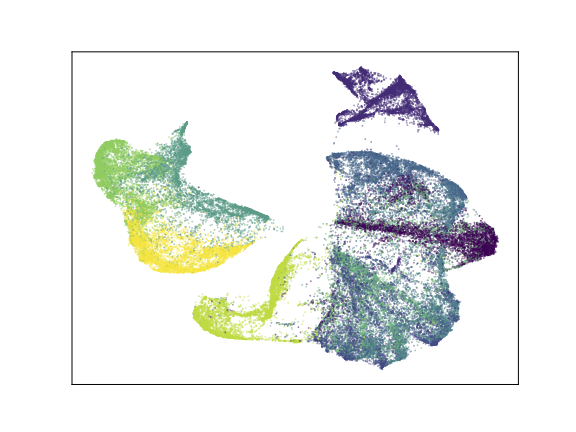} &
        \includegraphics[width=2.7cm]{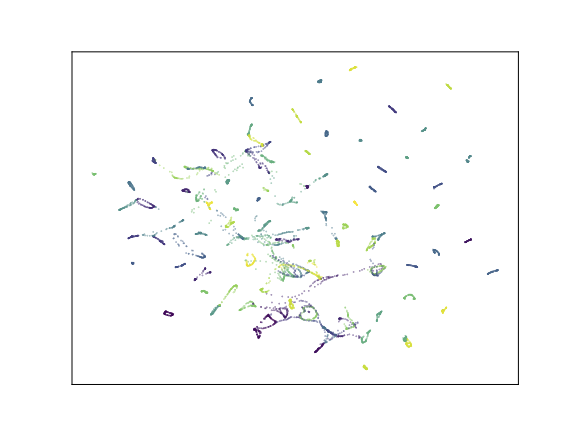} \\
    \end{tabularx}
    \caption{We show embeddings for the accelerated variants of \ourmethodU and \ourmethodN. Notice that the embeddings are very similar to those in Figure \ref{embedding_vis}.}
    \label{acc_embeddings}
\end{table}

\subsubsection{Accelerated \ourmethod}
\label{sssec:accelerated}
\begin{figure}[!htb]
    \includegraphics[width=.95\linewidth]{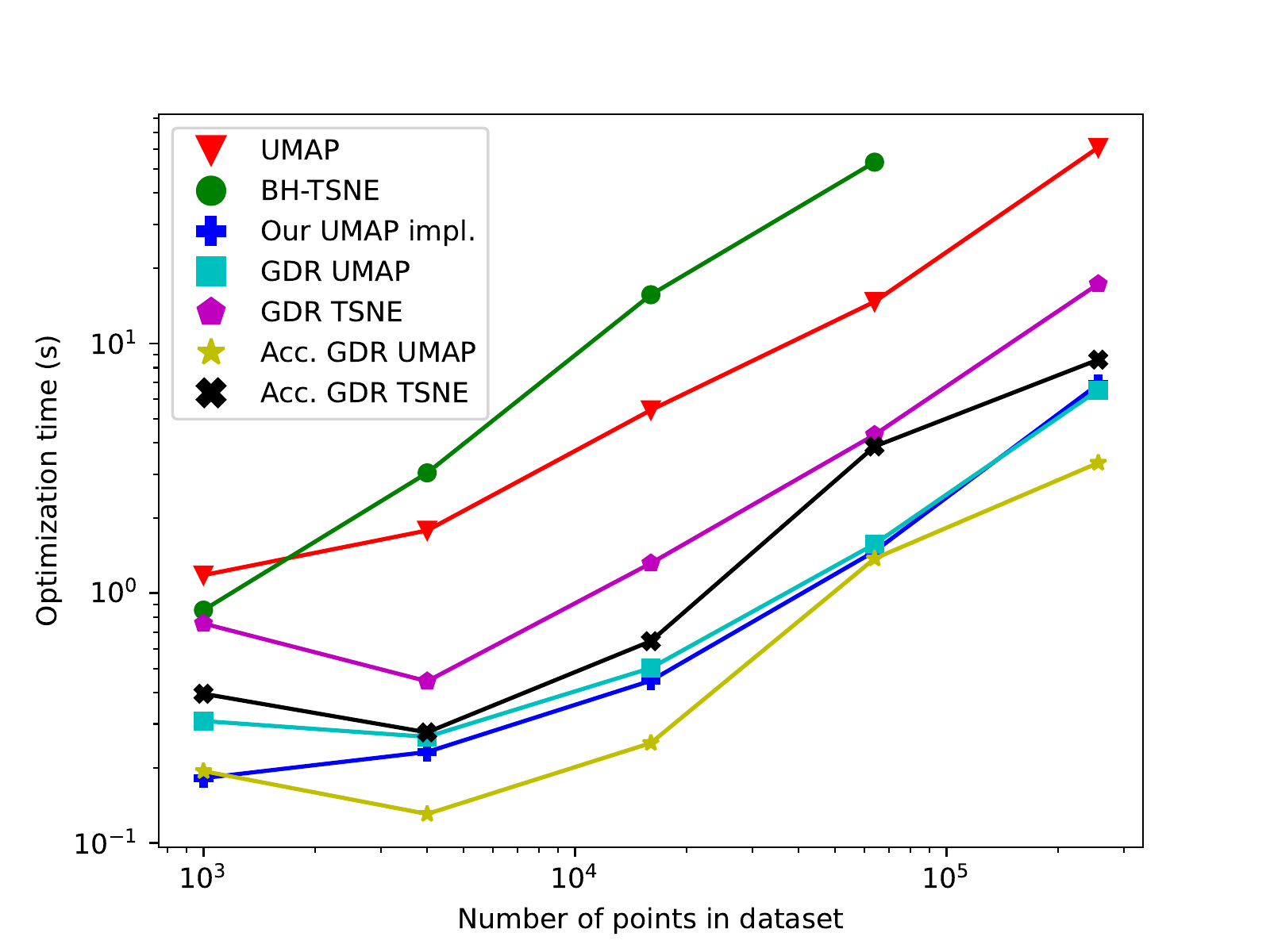}
    \caption{Runtimes for the methods discussed in this paper. We show both the speed of the original UMAP algorithm and our implementation of it for a fair comparison. Both axes are log-scale. Original dataset was the MNIST dataset and points were upsampled by randomly copying and adding noise.}
    \label{data_size_runtimes}
\end{figure}

We now turn to the accelerated version of \ourmethod. This can be seen in Figure~\ref{data_size_runtimes} as the models with `Acc.' in their name. The acceleration comes from a heuristic approximation of the gradient. Recall that UMAP uses scalar sampling, where they perform the $p_{ij}$ and $1 - p_{ik}$ multiplications by sampling the attractions and repulsions proportionally to those $0 \leq p \leq 1$ values. On the other hand, \ourmethod explicitly performs the multiplications.

The accelerated version of \ourmethod, then, has both the explicit $p_{ij}$ and $1 - p_{ik}$ multiplication \textit{and} only performs the attractions/repulsions by sampling proportionally to the $p$ value. This effectively means that we are choosing to focus our optimization on those points that have the highest contribution to the loss. We justify this by observing that the $p$ values directly scale the loss of the embedding under the KL divergence. Thus, we can incorporate UMAP's scalar sampling into \ourmethod as a means of heuristically preferring those connections that influence our loss the most. By doing both the explicit multiplication and the scalar sampling, our gradients for the accelerated variant effectively square the $p_{ij}$ and $1 - p_{ik}$ terms.

Surprisingly, we find that this obtains similar embeddings in both the normalized and unnormalized settings, which we show in Figure \ref{acc_embeddings}. This is yet another example of the flexibility of these methods. Not only could we replace the KL divergence with the Frobenius norm, but we can even arbitrarily modify the gradient formulas without significantly impacting the embeddings.

\subsection{UMAP gradient derivation}
\label{ssec:umap_derivation}

We start from the equations as they are presented in the UMAP paper and show that many terms cancel when we assume that $a = b = 1$ and that $\varepsilon = 0$. Recall that we had
\begin{align*}
    \mathcal{A}_i^{umap} = & \sum_{j, j \neq i} \dfrac{-2ab\|y_i - y_j\|_2^{2(b-1)}}{1 + \|y_i - y_j\|_2^2} p_{ij} (y_i - y_j) \\
    \mathcal{R}_i^{umap} = & \sum_{k, k \neq i} \dfrac{2b}{\varepsilon + \|y_i - y_k\|_2^2} q_{ik} (1 - p_{ik}) (y_i - y_k)
\end{align*}

We start with $\mathcal{A}_i^{umap}$:
\begin{align*}
    \mathcal{A}_i^{umap} &= \sum_{j, j \neq i} \dfrac{-2}{1 + \|y_i - y_j\|_2^2} p_{ij} (y_i - y_j) \\
    &= -2 \sum_{j, j \neq i} p_{ij} q_{ij} (y_i - y_j)
\end{align*}

For $\mathcal{R}_i^{umap}$, notice that
\begin{align*}
    q_{ik} &= \dfrac{1}{1 + \|y_i - y_k\|_2^2} \\
    \implies \|y_i - y_k\|_2^2 &= \dfrac{1}{q_{ik}} - 1 \\
    &= \dfrac{1 - q_{ik}}{q_{ik}}
\end{align*}
Setting $a=b=1$ and $\varepsilon = 0$ in $\mathcal{R}_i^{umap}$ and plugging this in then gives
\begin{align*}
    \mathcal{A}_i^{umap} &= 2 \sum_{k, k \neq i} \dfrac{1}{\left( \dfrac{1 - q_{ik}}{q_{ik}} \right)} q_{ik} (1 - p_{ik}) (y_i - y_k) \\
    &= 2 \sum_{k, k \neq i} q_{ik}^2 \dfrac{1 - p_{ik}}{1 - q_{ik}} (y_i - y_k)
\end{align*}

\subsection{Proof of Theorem \ref{thm:norm-changes-ratio}}
\label{prf:norm-changes-ratio}
We first restate the theorem before showing the proof.

Let $p_{ij}^{tsne} \sim 1/(cn)$ and $d(x_i, x_j) > \sqrt{ \log(n^2 + 1) \tau }$. Then
\[ \frac{\mathbb{E}[|\mathcal{A}_i^{umap}|]}{\mathbb{E}[|\mathcal{R}_i^{umap}|]} < \frac{\mathbb{E}\left[|\tilde{\mathcal{A}}_i^{tsne}|\right]}{\mathbb{E}\left[|\tilde{\mathcal{R}}_i^{tsne}|\right]} \]

As mentioned in section~\ref{ssec:norm_discussion}, $p_{ij}^{tsne}$ is normalized over the sum of all $cn$ attractions that are sampled, so $p_{ij}^{tsne} \sim 1/(cn)$ is an appropriate estimate.

\begin{proof}
For simplicity we assume that we are not using the pseudo-distance function for UMAP's high-dimensional kernel, meaning that $\rho_i = 0$. This only slightly affects the theorem's bound on $d(x_i, x_j)$ and we showed that the pseudo-distance does not affect the outputs per the experiments in Section~\ref{results}. The exclusion of $\rho_i$ only comes into play on the last line of the proof.

Note that the unnormalized attractive and repulsive forces can be written in terms of our random variables as $\mathcal{A}_i^{umap} = p_{ij}^{umap} r_{ij} v_{ij}$ and $\mathcal{R}_i^{umap} = r_{ij}^2 (1 - p_{ij}^{umap})/(1 - r_{ij}) v_{ij}$.

By Theorem \ref{thm:sampling-unnecessary}, $\mathbb{E}[\tilde{\mathcal{A}}_i^{tsne}] / \mathbb{E}[\tilde{\mathcal{R}}_i^{tsne}] = c p_{ij}^{tsne} / n$. Thus, plugging in $p_{ij}^{tsne} \sim 1/(cn)$ gives $\mathbb{E}[\tilde{\mathcal{A}}_i^{tsne}] / \mathbb{E}[\tilde{\mathcal{R}}_i^{tsne}] \sim 1 / n^2$.

Since $\mathcal{A}^{umap}_i$, $\mathcal{R}^{umap}_i$, $\tilde{\mathcal{A}}^{tsne}_i$, and $\tilde{\mathcal{R}}^{tsne}_i$ represent the force exerted by a single point, there are no summations to account for. Additionally, since the scalars are all non-negative, the scalars can be placed outside the norms.

Looking at the unnormalized force ratio, we have
\[    \frac{\mathbb{E}[|\mathcal{A}_i^{umap}|]}{\mathbb{E}[|\mathcal{R}_i^{umap}|]} = \frac{p_{ij}^{umap} \mathbb{E}\left[ || r_{ij} v_{ij} || \right]}{ (1-p_{ij}^{umap})\mathbb{E}\left[ || \frac{r_{ij}^2}{1 - r_{ij}} v_{ij} || \right]} \]

Consider that $|| r_{ij}^2/(1 - r_{ij}) \cdot v_{ij} || = || (r_{ij} v_{ij}) \cdot r_{ij}/(1 - r_{ij}) || > || r_{ij} v_{ij} ||$, since $r_{ij} < 1$ is a scalar. Thus, we can write
\[ \frac{p_{ij}^{umap} \mathbb{E}\left[ || r_{ij} v_{ij} || \right]}{ (1-p_{ij}^{umap})\mathbb{E}\left[ || \frac{r_{ij}^2}{1 - r_{ij}} v_{ij} || \right]} < \frac{p_{ij}^{umap}}{1 - p_{ij}^{umap}} \]

Therefore, it suffices to show that
\[ \frac{p_{ij}^{umap}}{1 - p_{ij}^{umap}} < \frac{1}{n^2} \]

Solving for $p_{ij}^{umap}$ gives
\[ p_{ij}^{umap} < \frac{1}{n^2 + 1} \]

Further solving this for the high-dimensional distance $d(x_i, x_j)$ gives
\[ d(x_i, x_j) > \sqrt{ \log(n^2 + 1) \tau_i }\]

If we had incorporated the pseudo-distance metric, we would have
\[ d(x_i, x_j) > \sqrt{ \log(n^2 + 1) \tau_i + \rho_i}\]

\end{proof}

\subsection{Proof of theorem \ref{thm:sampling-unnecessary}}
\label{prf:sampling-unnecessary}
Recall the theorem statement:
\[ \dfrac{\mathbb{E}\left[|\mathcal{A}_i^{tsne}|\right]}{\mathbb{E}\left[|\mathcal{R}_i^{tsne}|\right]} = \dfrac{\mathbb{E}\left[|\tilde{\mathcal{A}}_i^{tsne}|\right]}{\mathbb{E}\left[|\tilde{\mathcal{R}}_i^{tsne}|\right]} \]
\begin{proof}
First, notice that $Z$ and $\tilde{Z}$ are sums of i.i.d. random variables, so we can write $Z = n^2 r_{ij}$ and $\tilde{Z} = n r_{ij}$. Then the proof is a matter of simple algebraic steps.
\begin{align*}
    \mathbb{E}\left[|\mathcal{A}_i^{tsne}|\right] &= c \cdot \mathbb{E}\left[ p_{ij}^{tsne} ||\frac{r_{ij}}{Z} v_{ij} || \right] \\
    &= c \cdot p_{ij}^{tsne} \mathbb{E}\left[ || \frac{r_{ij}}{n^2 r_{ij}} v_{ij} ||\right] \\
    &= c \cdot \frac{p_{ij}^{tsne}}{n^2}  \mathbb{E}[||v_{ij}||] \\
    \mathbb{E}\left[|\mathcal{R}_i^{tsne}|\right] &= \mathbb{E}\left[ \sum_{j} ||\frac{r_{ij}^2}{Z^2} v_{ij}|| \right] \\
    &= \sum_{j} \mathbb{E}\left[ || \frac{r_{ij}^2 v_{ij}}{n^4 r_{ij}^2} || \right] \\
    &= \frac{1}{n^4}\sum_{j} \mathbb{E}[ || v_{ij} || ] \\
    &= \frac{n \mathbb{E}[||v_{ij}||]}{n^4} = \frac{\mathbb{E}[||v_{ij}||]}{n^3} \\
    \mathbb{E}\left[|\tilde{\mathcal{A}}_i^{tsne}|\right] &= c \cdot \mathbb{E}\left[ p_{ij}^{tsne} ||\frac{r_{ij}}{\tilde{Z}} v_{ij}|| \right] \\
    &= c \cdot \frac{p_{ij}^{tsne}}{n} \mathbb{E}[||v_{ij}||] \\
    \mathbb{E}\left[|\tilde{\mathcal{R}}_i^{tsne}|\right] &= \mathbb{E}\left[ ||\frac{r_{ij}^2}{\tilde{Z}^2} v_{ij}|| \right] \\
    &= \mathbb{E}\left[ ||\frac{r_{ij}^2}{n^2 r_{ij}^2} v_{ij}|| \right] = \frac{\mathbb{E}[||v_{ij}||]}{n^2}
\end{align*}
Plugging these values in, we get our theorem statement:
\[ \dfrac{\mathbb{E}\left[|\mathcal{A}_i^{tsne}|\right]}{\mathbb{E}\left[|\mathcal{R}_i^{tsne}|\right]} = \dfrac{\mathbb{E}\left[|\tilde{\mathcal{A}}_i^{tsne}|\right]}{\mathbb{E}\left[|\tilde{\mathcal{R}}_i^{tsne}|\right]}, \]
where both attraction/repulsion ratios are equal to $ c p_{ij}^{tsne}/n $.
\end{proof}

\subsection{Manifold Learning with tSNE and UMAP}
\label{ssec:manifold_learning}
\begin{figure}[ht]
    \centering
    \begin{tabular}{m{0.3em}cc}
        & TSNE & UMAP \\
        \makecell{\rotatebox[origin=c]{90}{Normalized}} &
        \raisebox{-0.5\height}{\includegraphics[width=3.75cm]{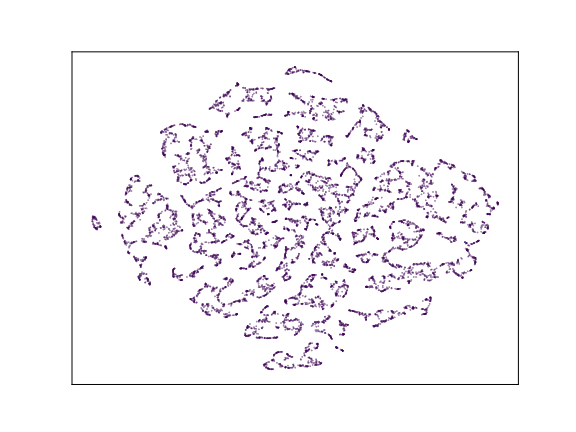}} &
        \raisebox{-0.5\height}{\includegraphics[width=3.75cm]{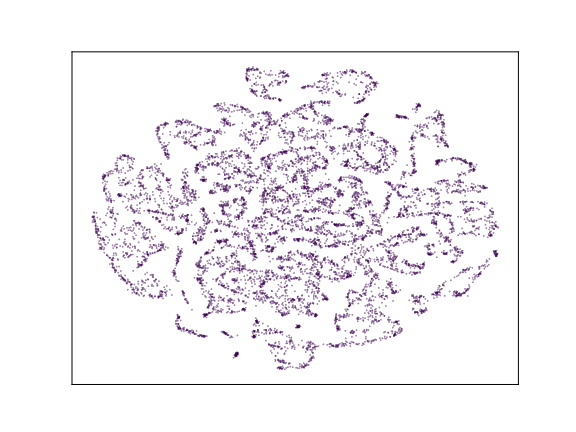}}\\
        \makecell{\rotatebox[origin=c]{90}{Unnormalized}} & 
        \raisebox{-0.5\height}{\includegraphics[width=3.75cm]{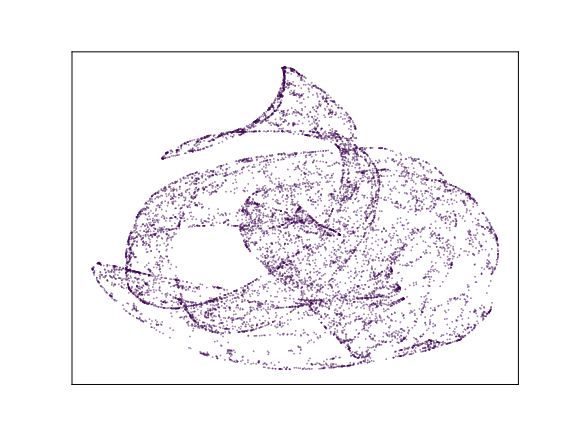}} &
        \raisebox{-0.5\height}{\includegraphics[width=3.75cm]{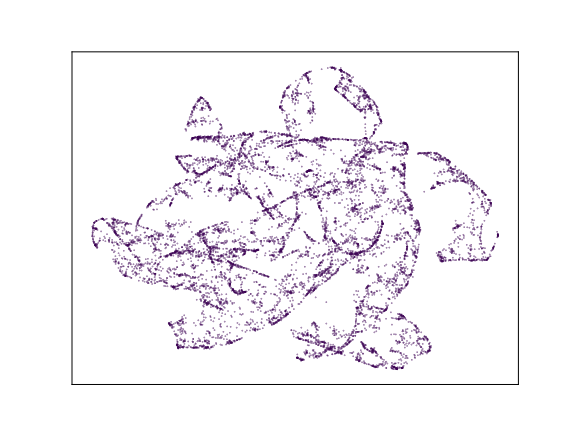}}\\
    \end{tabular}
\caption{A look at manifold preservation of tSNE and UMAP under random initialization on the swiss roll dataset. Notice that the manifold is not preserved by \textit{any} of the methods, regardless of normalization. This implies that the manifold learning effects of UMAP are largely due to the Laplacian Eigenmap initialization, at least on this dataset.}
\label{swiss_roll_plots}
\end{figure}

Despite UMAP's reputation for faithfully representing the high-dimensional manifold, we question the extent to which tSNE and UMAP can perform manifold learning. As seen in Table~\ref{swiss_roll_plots}, simply removing the Laplacian Eigenmap initialization prevents tSNE and UMAP from unrolling a swiss-roll dataset in $\mathbb{R}^{10000 \times 3}$. Since Laplacian Eigenmaps are based on graph structures that closely match the manifold, using them for the initialization will pre-condition the embedding to respect the manifold structure. We do not intend to say that tSNE and UMAP cannot perform manifold learning -- we instead use this counterexample to show that these claims could be better substantiated.

There has been work suggesting that the Laplacian Eigenmap initialization accounts for the global structure of the data. This was in reference to the manifold structure, which is defined by the nearest-neighbor graph. Our results are instead on a more macro-level, where we show that the normalization impacts the overall structure of the clusters and the spaces between them. However, we hypothesize that if the Laplacian eigenmap initialization starts with large inter-cluster distances, then it may subsequently find itself in the large areas of 0 gradients in figure \ref{grad_plots}.

\begin{figure}[ht]
    \newcolumntype{C}{ >{\centering\arraybackslash} m{2.3cm} }
    \begin{tabular}{CCC}
        \includegraphics[width=2.7cm]{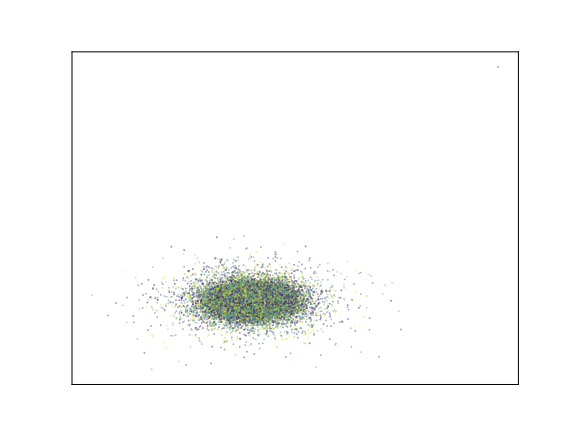} & 
        \includegraphics[width=2.7cm]{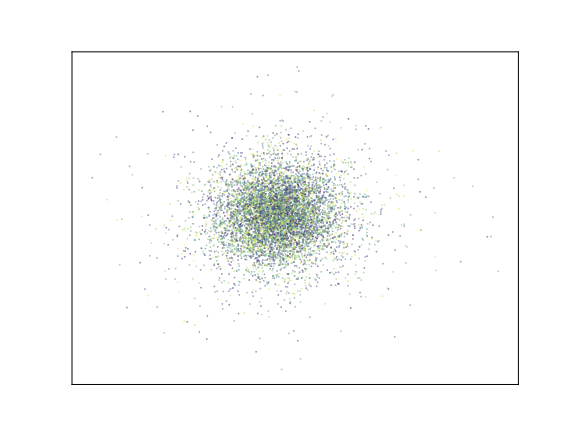} &
        \includegraphics[width=2.7cm]{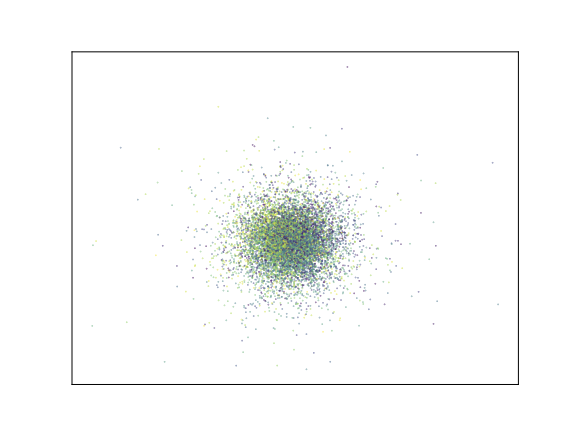} \\
        \quad MNIST & \quad COIL100 & \quad Swiss Roll \\
    \end{tabular}
\caption{The effect of normalizing UMAP without the changes discussed in \ref{ssec:norm_results}.}
\label{normed_umap}
\end{figure}

\subsection{Looking closer at \texorpdfstring{$\mathcal{A}$}{A} and \texorpdfstring{$\mathcal{R}$}{R}}
\label{ssec:forces_estimate}

First, note that the UMAP repulsive force in Equation~\ref{umap_rep} is inversely quadratic with respect to the low-dimensional distance, leading to extreme repulsions between points that are too close in the low-dimensional space. This comes as a direct result of the $\log(1 - q_{ij})$ term in equation \ref{eq:losses}, since
\begin{align*}
\dfrac{\partial \log(1 - q_{ij})  }{\partial y_i} &= \dfrac{1}{1 - q_{ij}} \cdot \dfrac{\partial q_{ij}}{\partial y_i} \\
&= \dfrac{1 + a \|y_i - y_j\|^{2b}}{a \|y_i - y_j\|^{2b}} \cdot \dfrac{\partial q_{ij}}{\partial y_i},
\end{align*}
where the numerator cancels out after expanding $\partial q_{ij} / \partial y_i$.

This inverse relationship to the distance is inherently unstable when $||y_i - y_j||_2^2$ is small and is handled computationally by adding an $\epsilon$ additive term in equation \ref{umap_rep}. Nonetheless, UMAP repulsions are still unwieldy and must be managed by clipping the gradients and disallowing momentum-based SGD.

We also point out that most values of $p_{ik}$ are unavailable to us during optimization since we only calculated the $P$ matrix for nearest neighbors in the high-dimensional space. UMAP approximates these $p_{ik}$ values by plugging in the available $p_{ij}$ term instead. We standardize the UMAP approximation of $p_{ik} \approx p_{ij}$ by setting $p_{ik} = \bar{p}_{ij} \; \forall \; p_{ik}$, where $\bar{p}_{ij}$ is the mean value of $P$ (for known values in $P$).

We lastly mention that $Zq^{tsne} \sim q^{umap}$, so the $Zq$ term in the tSNE forces is equivalent to the $q$ in the UMAP forces. However the other $p_{ij}$ and $q_{ik}$ terms in $\mathcal{A}^{tsne}$ and $\mathcal{R}^{tsne}$ remain normalized. This creates the effect that UMAP's gradients are about a factor of $n$ stronger than the corresponding tSNE ones. We account for this in \ourmethod by scaling the tSNE learning rate by a factor of $n/k$ when operating in a normalized setting.

\subsection{\ourmethod with the Frobenius Norm}
\label{ssec:frob_norm}
\begin{figure}[ht]
    \centering
    \begin{tabular}{m{0.3em}cc}
        & UMAP & \ourmethodU \\
        \makecell{\rotatebox[origin=c]{90}{MNIST}} &
        \raisebox{-0.5\height}{\includegraphics[width=3.75cm]{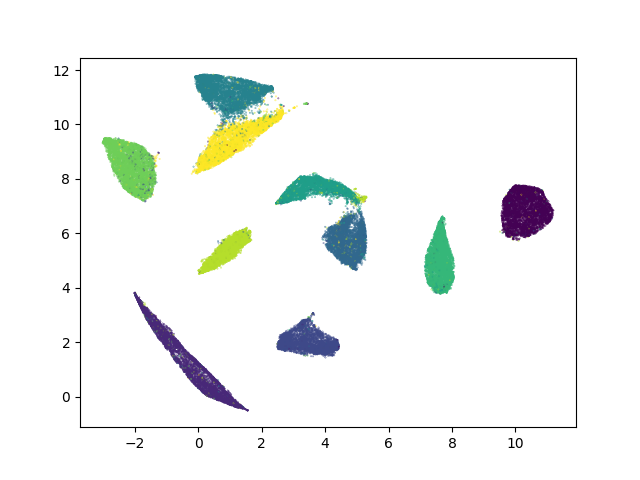}} &
        \raisebox{-0.5\height}{\includegraphics[width=3.75cm]{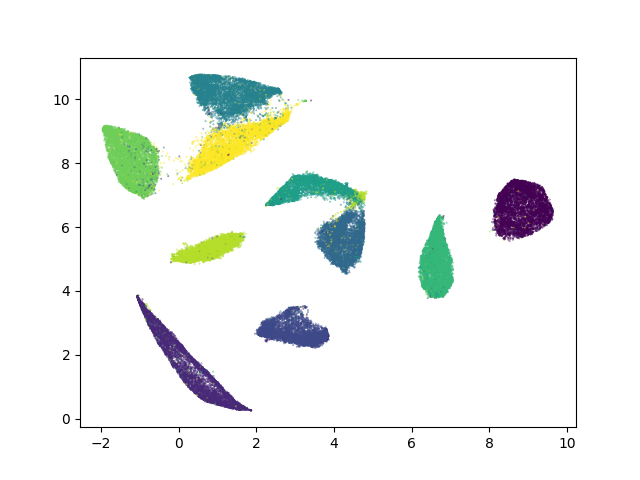}}\\
        \makecell{\rotatebox[origin=c]{90}{Fashion-MNIST}} & 
        \raisebox{-0.5\height}{\includegraphics[width=3.75cm]{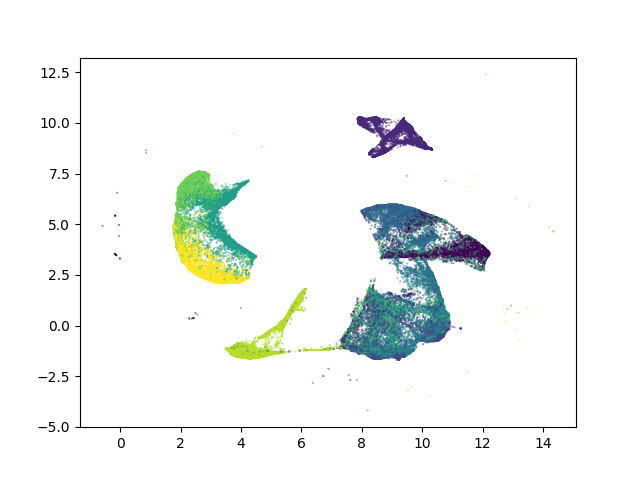}} &
        \raisebox{-0.5\height}{\includegraphics[width=3.75cm]{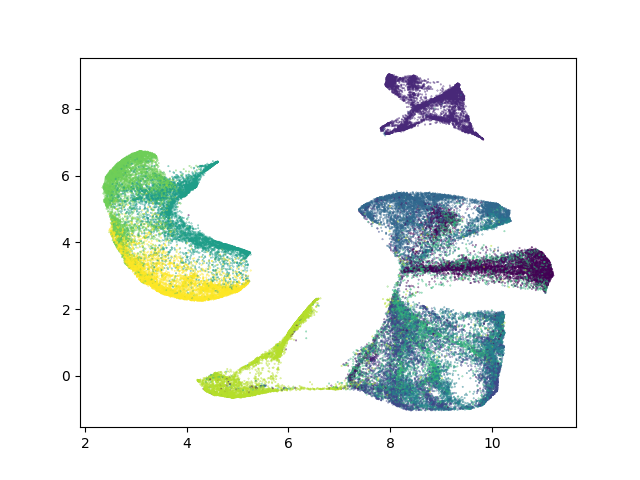}}\\
    \end{tabular}
\caption{Embeddings obtained in the unnormalized setting by optimizing the Frobenius norm of $P(X) - Q(Y)$ on the MNIST and Fashion-MNIST datasets.}
\label{fig:frob_embeddings}
\end{figure}

As discussed in Section \ref{ssec:forces_estimate}, the unknown-at-runtime $1 - p_{ik}$ scalar in equation \ref{umap_rep} is a direct consequence of the KL-divergence in the unnormalized setting. Surprisingly, replacing the KL divergence by the Frobenius norm gives almost identical embeddings while reconciling the $1 - p_{ik}$ concern. We denote the loss under the Frobenius norm by 
\[ \mathcal{L}^{frob}(X, Y) = \sum_{i, j} (p(x_i, x_j) - q(y_i, y_j))^2 \]

This presents us with the following attractive and repulsive forces acting on point $y_i$
\begin{align*}
    \mathcal{A}_i^{frob-umap} &= -4 \sum_{j} p_{ij} q_{ij}^2 (y_i - y_j) \\
    \mathcal{R}_i^{frob-umap} &= 4 \sum_{j} q_{ij}^3 (y_i - y_j)  
\end{align*}
We note that the Frobenius norm gradients are only stable in the unnormalized setting.

Although the embeddings under the squared Frobenius norm and KL divergence in Tables~\ref{irrelevant-mnist}~and~\ref{irrelevant-metrics_col_means} have a completely different theoretical structure, they appear to be qualitatively and quantitatively comparable. Interestingly, the gradient plots in \ref{grad_plots} show that a majority of the gradient space under the Frobenius norm still has magnitude zero (deep blue area in the gradient plots).

The Frobenius norm has multiple advantages over the KL-divergence. First, its convexity opens the door to many new approaches towards optimization. Secondly, it avoids the $1 - p_{ik}$ scaling factor. Lastly, the Frobenius norm gradients\footnote{We wrote the gradients under the assumption that $a = b = 1$, although they are simple to calculate in the general setting as well.} provide a convenient function for the attractions and repulsions in the unnormalized setting that do not require $\epsilon$ values for stability. We do not default to minimizing the Frobenius norm in \ourmethod as it is not the traditionally accepted loss function.

\subsection{Metrics and Datasets}
\label{ssec:metrics}

\spara{Metrics.} We report the \emph{kNN-accuracy}, i.e., the accuracy of a k-NN classifier, to assert that objects of a similar class remain close in the embedding. Assuming that intra-class distances are smaller than inter-class distances in the high-dimensional space, a high kNN-accuracy implies that the method effectively preserves similarity during dimensionality reduction. Unless stated otherwise, we choose $k = 100$ in-line with prior work~\cite{mcinnes2018umap}.

We study embedding consistency by evaluating KMeans clustering on datasets that have class labels. We report cluster quality in terms of \emph{homogeneity} and \emph{completeness}~\cite{rosenberg2007v}. Homogeneity is maximized by assigning \textit{only} data points of a single class to a single cluster while completeness is maximized by assigning \textit{all} data points of a class to a cluster. For brevity, we report the \emph{V-measure}, the average between the homogeneity and completeness. This metric simply relies on the labeling and does not take the point locations into account. As such, it is invariant to the biases inherent in KMeans and serves as a more objective measure than KMeans loss.

\para{Experiment setup}
We performed each experiment once as we show the averages over all hyperparameters, algorithms, and datasets in our results. Timing experiments were done on an Intel Core i9 10940X 3.3GHz 14-Core processor, fully parallelized over the cores.

\spara{Datasets.} We use standard datasets that are common in dimensionality reduction papers \cite{van2014accelerating}, \cite{mcinnes2018umap}, \cite{tang2016visualizing}. In particular, we employ the popular MNIST~\cite{lecun-mnisthandwrittendigit-2010}, Fashion-MNIST~\cite{xiao2017fashion}, CIFAR~\cite{krizhevsky2009learning}, Coil~\cite{nene1996columbia} image datasets, and the Swiss Roll synthetic dataset. As tSNE and UMAP are often used on biological data, we also include the single cell dataset described in~\cite{tasic2018shared}. We chose the `cell\_cluster' metadata as the label.

\begin{table}[H]
\centering
\setlength{\tabcolsep}{2pt}
\newcolumntype{C}{>{\centering\arraybackslash}X}
\begin{tabularx}{\linewidth}{XrrrX}
\toprule
\textbf{Dataset} & $n$ & $c$ & $D$ & Type\\ 
\midrule
MNIST &  60\,000  & 10 & 784 & Images \\
Fashion-MNIST &  60\,000  & 10 & 784 & Images\\
CIFAR-10 & 60\,000 & 10 & 3\,072 & Images\\
Coil-100 & 7\,200 & 100 & 49\,152 & Images \\
\midrule
Single Cell & 23\,100 & 152 & 45\,769 & Single Cell \\
\midrule
Google News & 350\,000 & -- & 200 & Word Vectors\\
Swiss Roll & 5\,000 & -- & 3 & Synthetic\\
\bottomrule
\end{tabularx}
\caption{Dataset characteristics -- $n$ samples, $c$ classes, $D$ dimensions \protect\cite{lecun-mnisthandwrittendigit-2010}\protect\cite{xiao2017fashion}\protect\cite{krizhevsky2009learning}\protect\cite{nene1996columbia}\protect\cite{mikolov2013efficient}\protect\cite{uspsdataset}. Google News and Swiss Roll are unlabeled. The Google News dataset is the 350K most frequent word vectors and can be found at https://data.world/jaredfern/googlenews-reduced-200-d}
\label{tbl:datasets}
\vspace{-4mm}
\end{table}

\begin{figure*}[ht]
    \centering
    \begin{tabular}{ccc}
        UMAP & \ourmethodU & \ourmethodN \\
        \includegraphics[width=5cm]{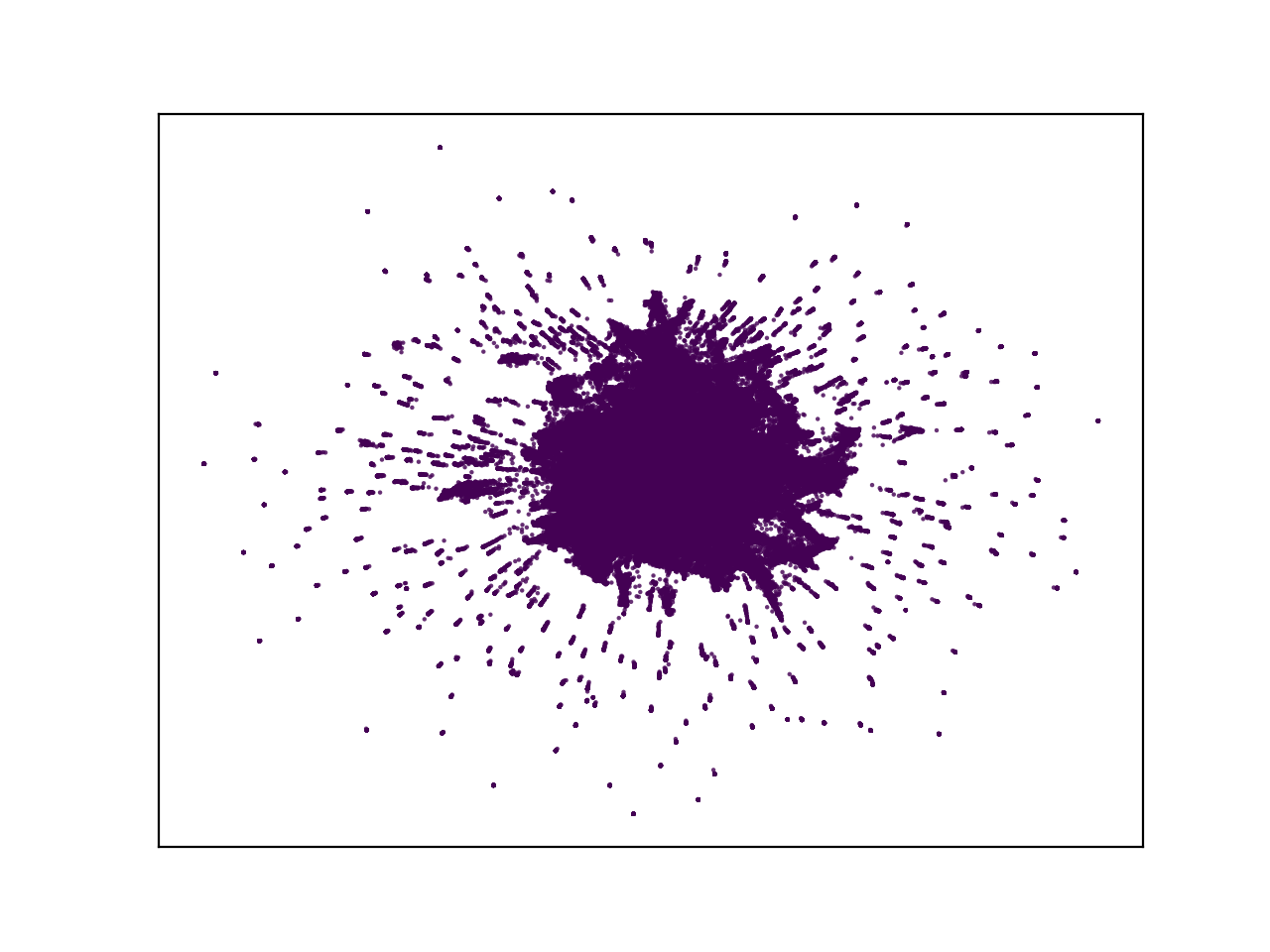} &
        \includegraphics[width=5cm]{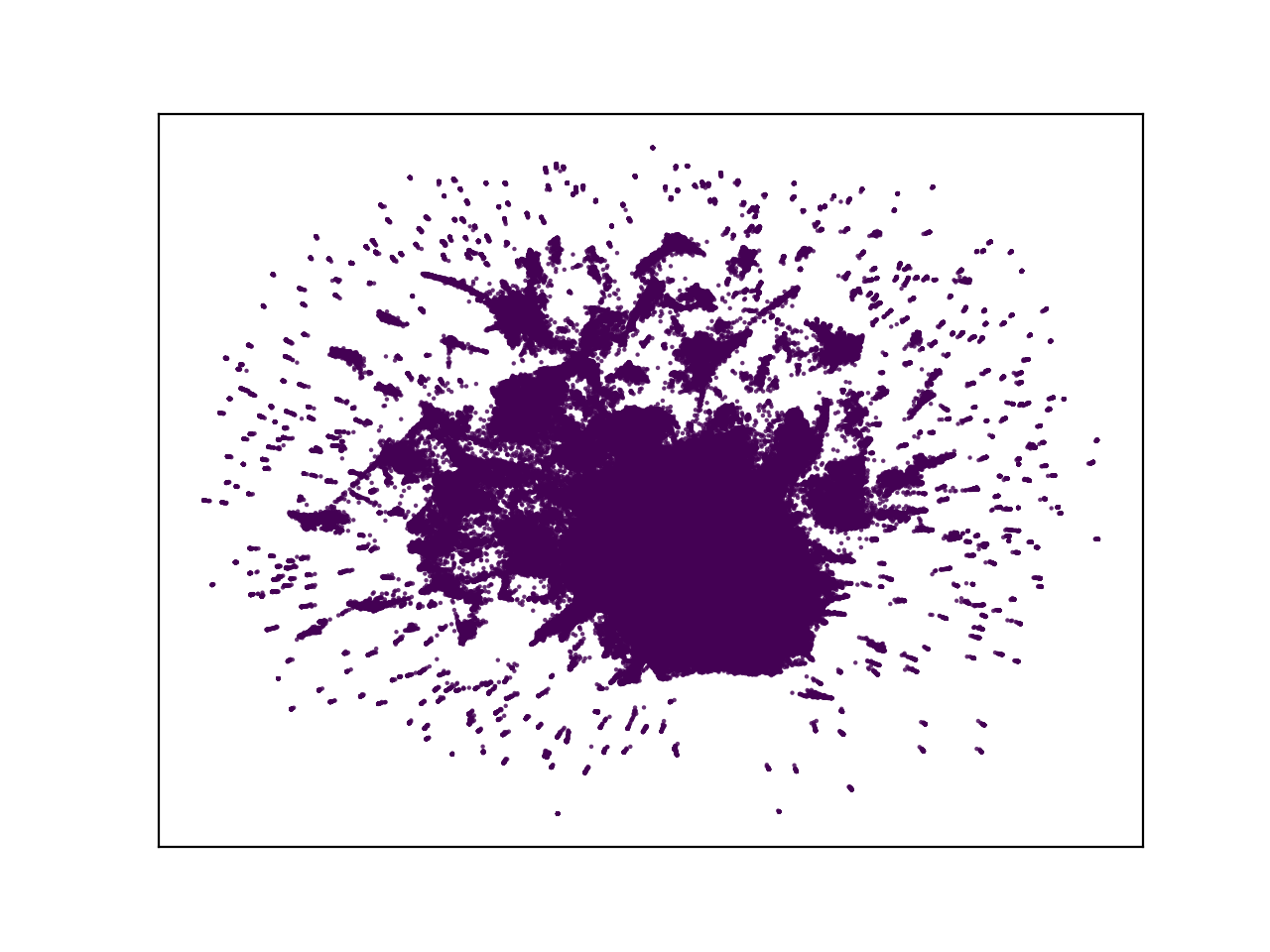} &
        \includegraphics[width=5cm]{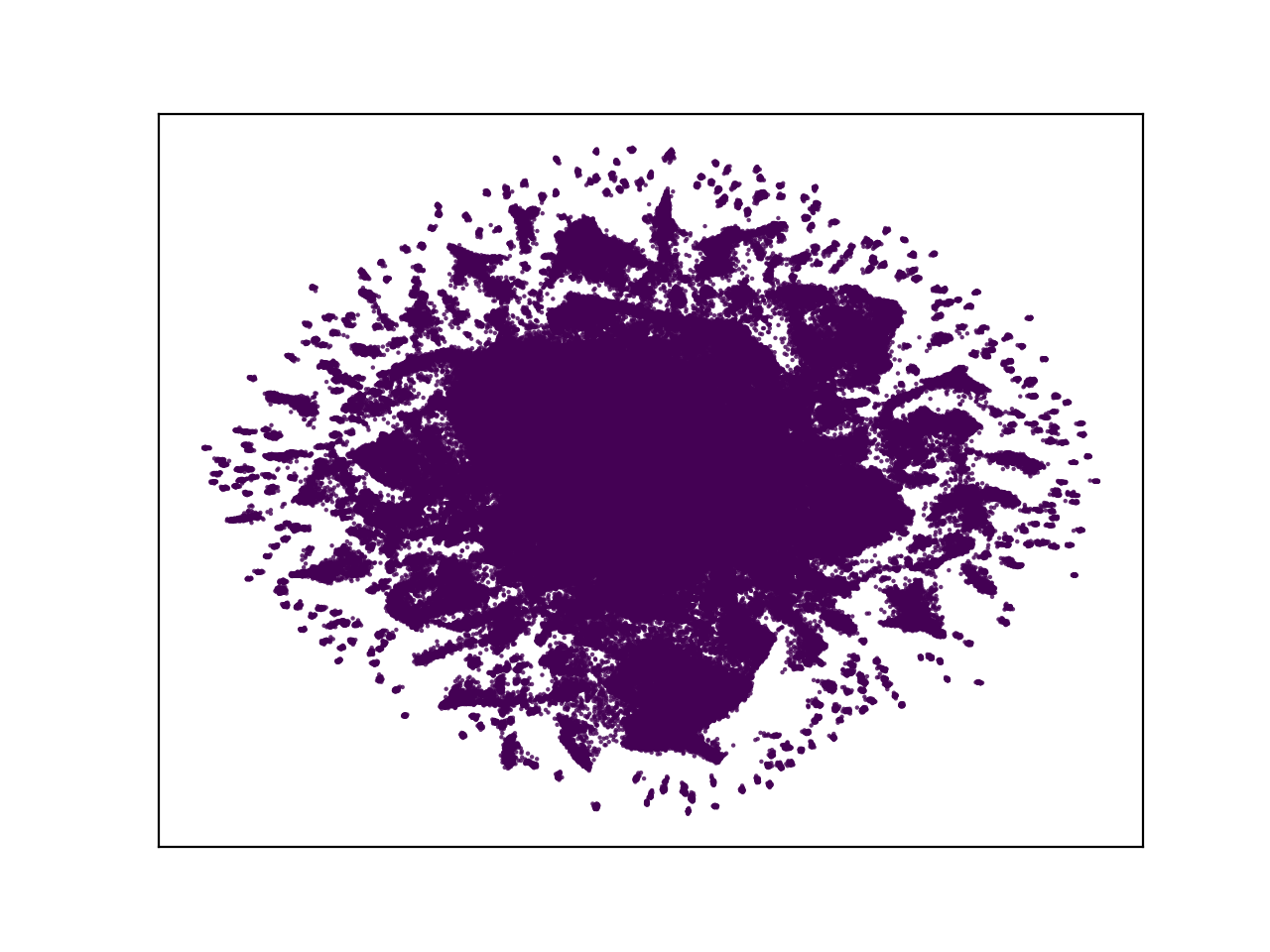}
    \end{tabular}
\caption{Embeddings on the Google news dataset. We did not run tSNE out of time considerations.}
\label{gnews_plots}
\end{figure*}

\begin{table*}[!thb]
    \setlength{\tabcolsep}{6pt}
    \centering
    \begin{tabular}{p{0.25cm}*{6}{>{\centering\arraybackslash}p{2.4cm}}}
    & \textbf{Default setting} & Random init & Pseudo distance & Symmetrization & Sym attraction & a, b scalars \\

    \rotatebox[origin=l]{90}{\bf \;\; tSNE} & 
    \includegraphics[width=2.5cm]{outputs/fashion_mnist/tsne/default_embedding.png}&
    \includegraphics[width=2.5cm]{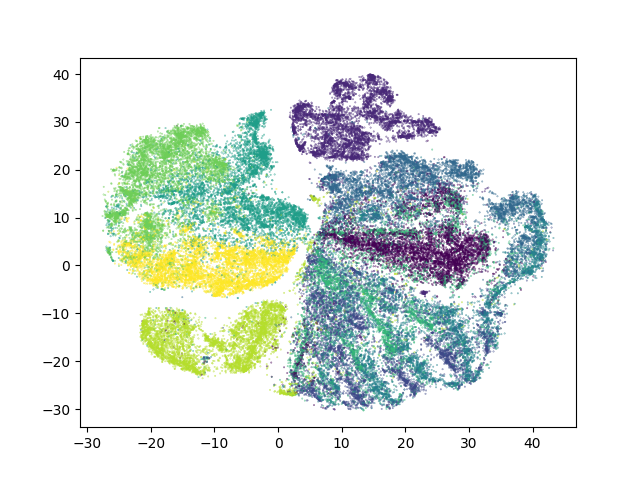}&
    \includegraphics[width=2.5cm]{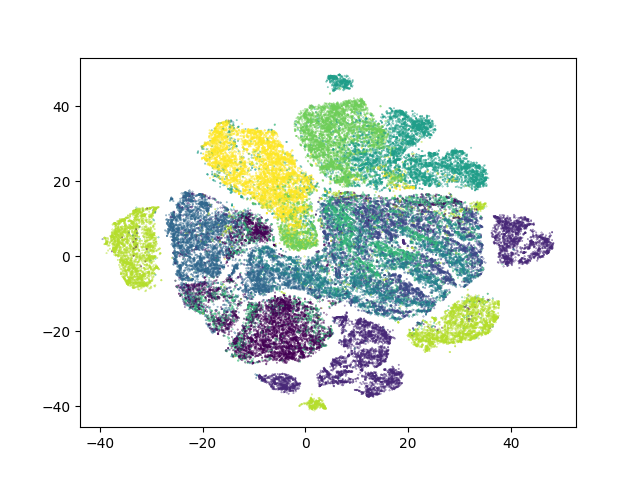} & 
    \includegraphics[width=2.5cm]{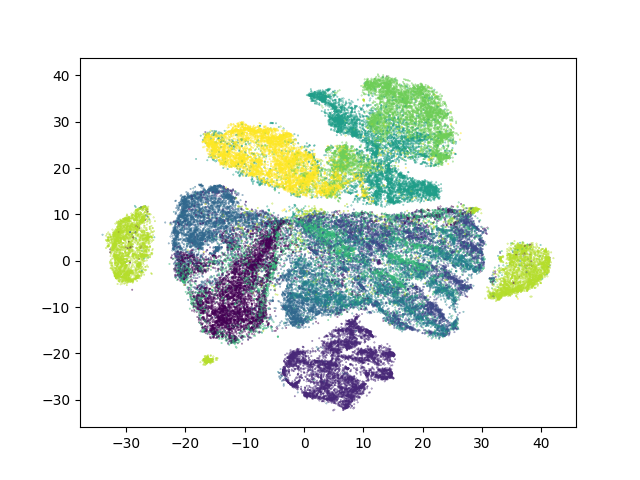}&
    \includegraphics[width=2.5cm]{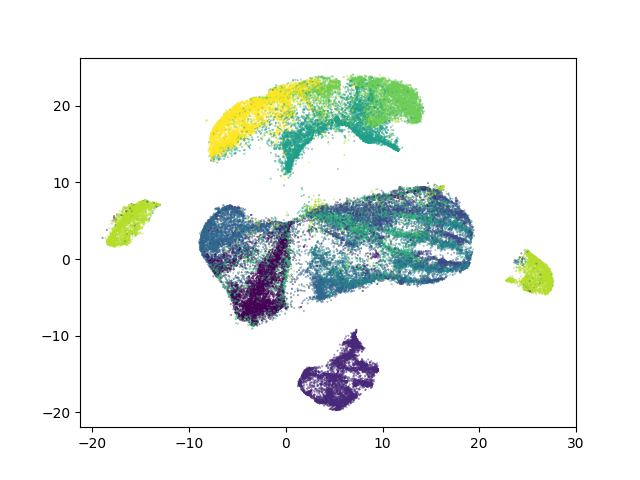}&
    \includegraphics[width=2.5cm]{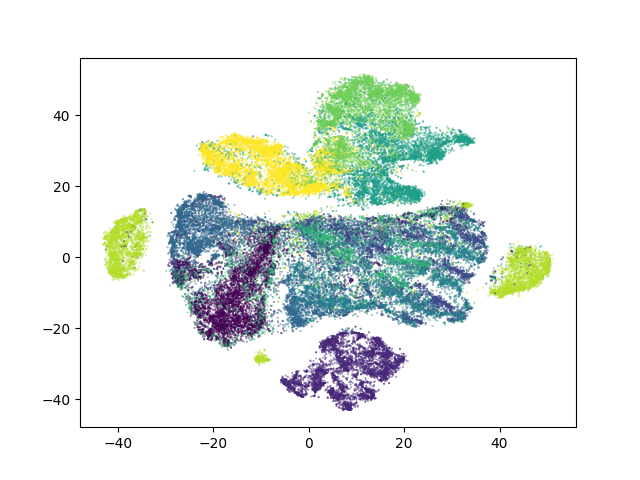} \\

    \\[-1.25em]
    {\ourcell}\rotatebox[origin=l]{90}{\bf \; \ourmethodN} & 
    \includegraphics[width=2.5cm]{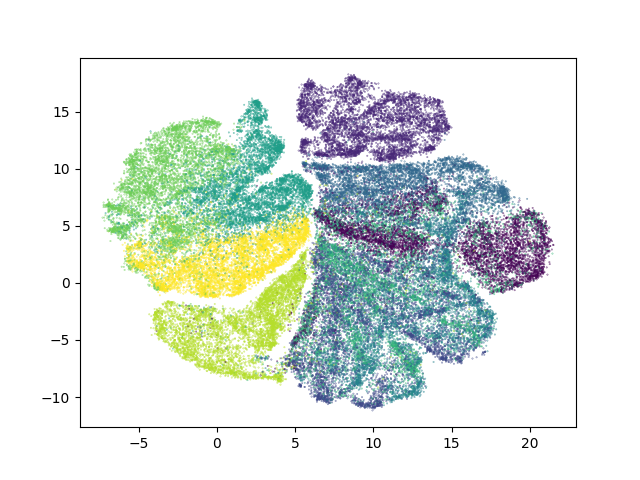}&
    \includegraphics[width=2.5cm]{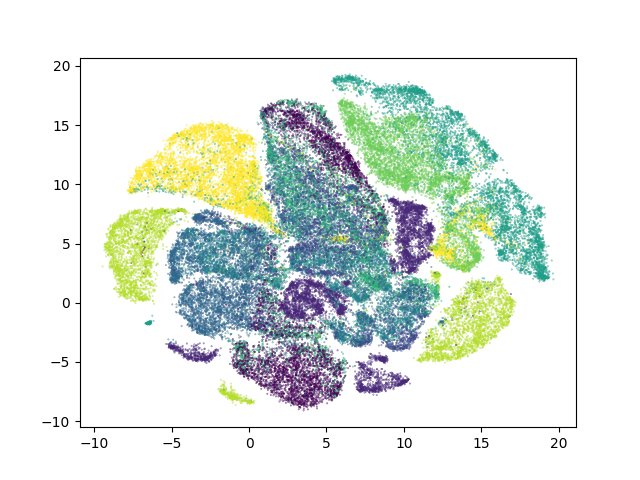}&
    \includegraphics[width=2.5cm]{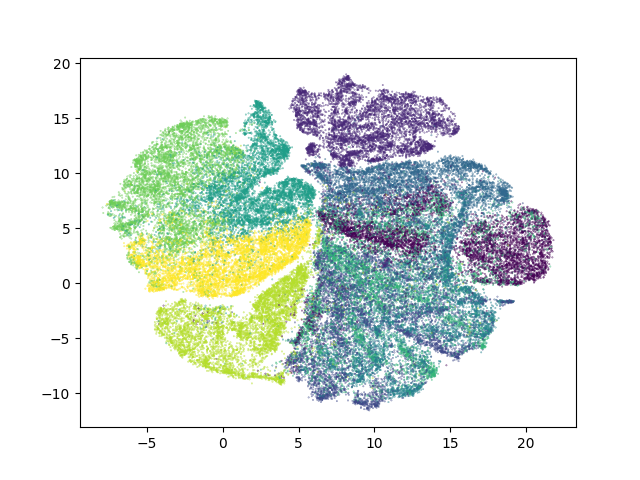}& 
    \includegraphics[width=2.5cm]{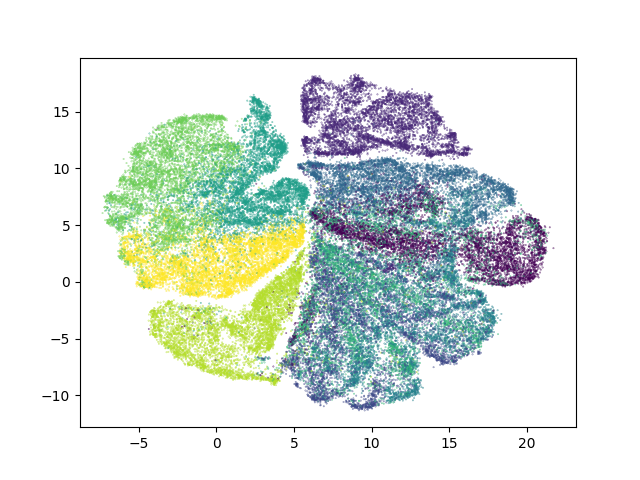}&
    \includegraphics[width=2.5cm]{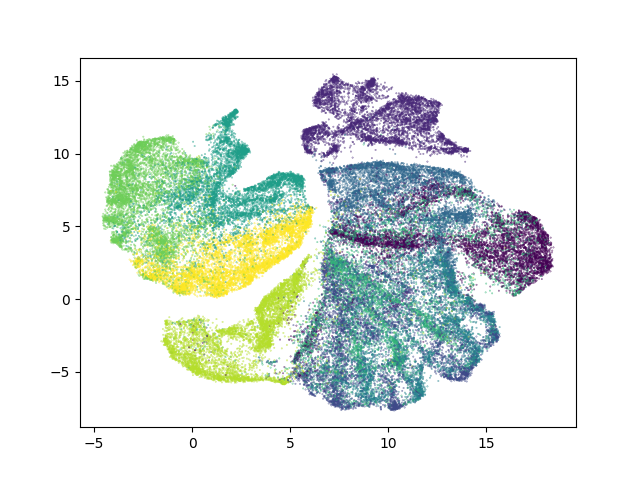}&
    \includegraphics[width=2.5cm]{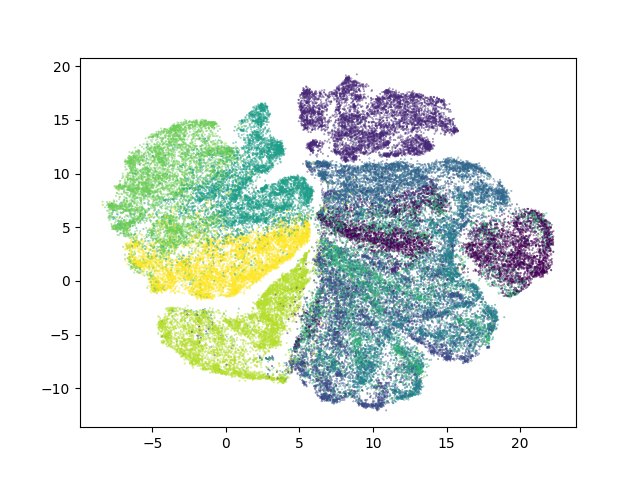}\\

    \\[-1.25em]
    \rotatebox[origin=l]{90}{\bf \;\; UMAP} & 
    \includegraphics[width=2.5cm]{outputs/fashion_mnist/umap/default_embedding.png}&
    \includegraphics[width=2.5cm]{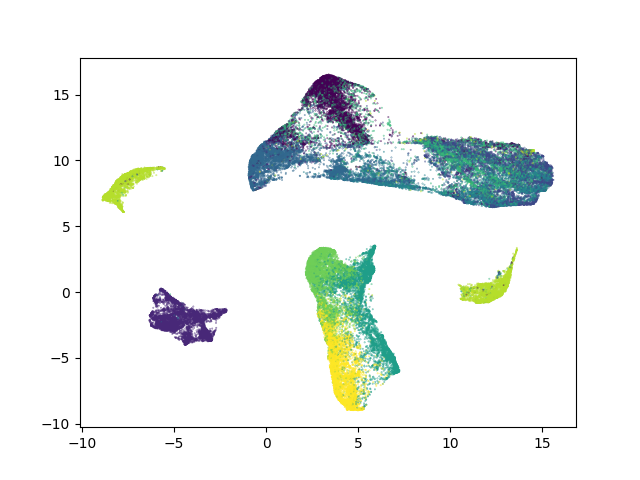}&
    \includegraphics[width=2.5cm]{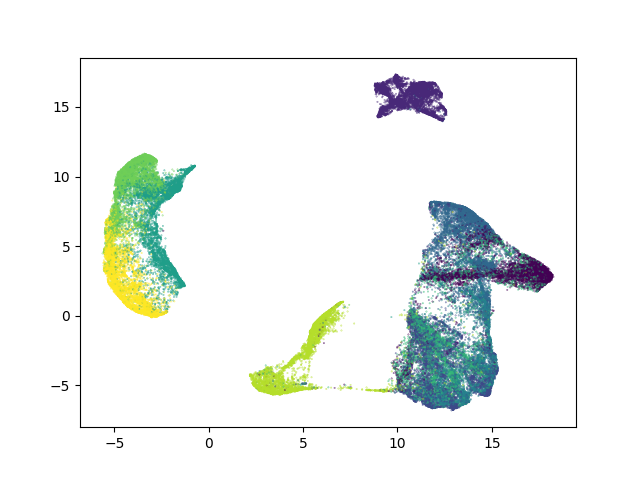}& 
    \includegraphics[width=2.5cm]{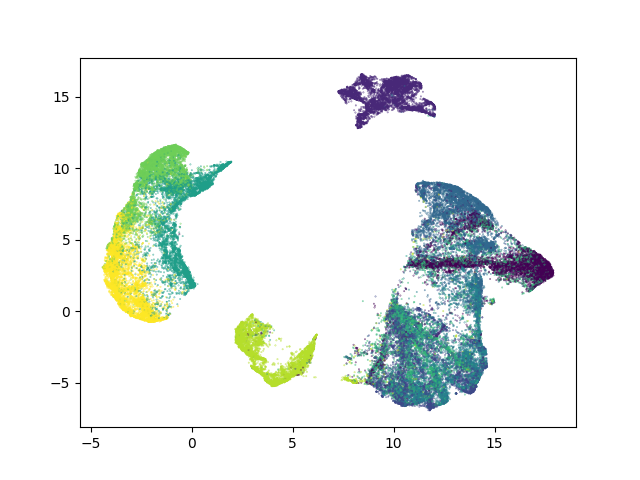}&
    \includegraphics[width=2.5cm]{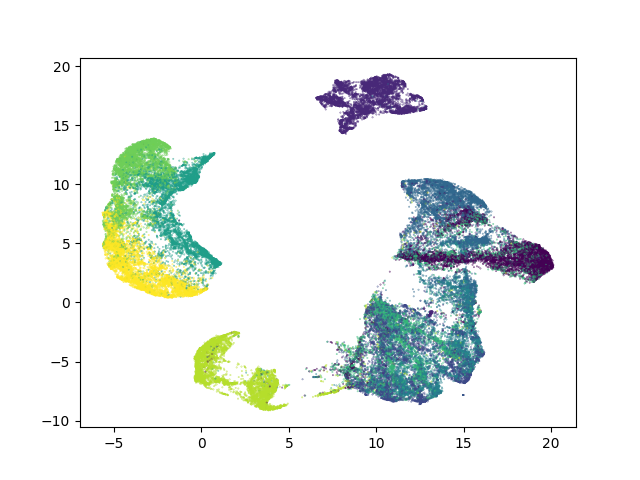}&
    \includegraphics[width=2.5cm]{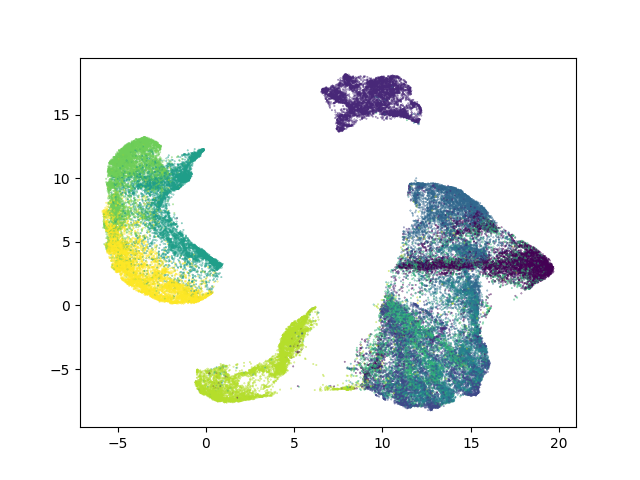}\\

    \\[-1.25em]
    {\ourcell}\rotatebox[origin=l]{90}{\bf \, \ourmethodU} & 
    \includegraphics[width=2.5cm]{outputs/fashion_mnist/uniform_umap/default_embedding.png}&
    \includegraphics[width=2.5cm]{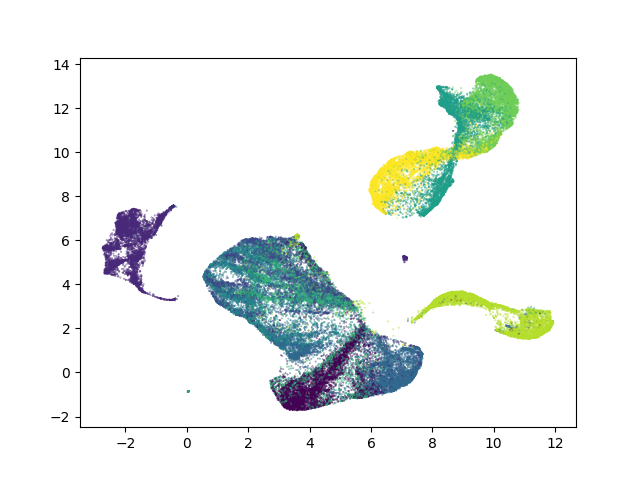}&
    \includegraphics[width=2.5cm]{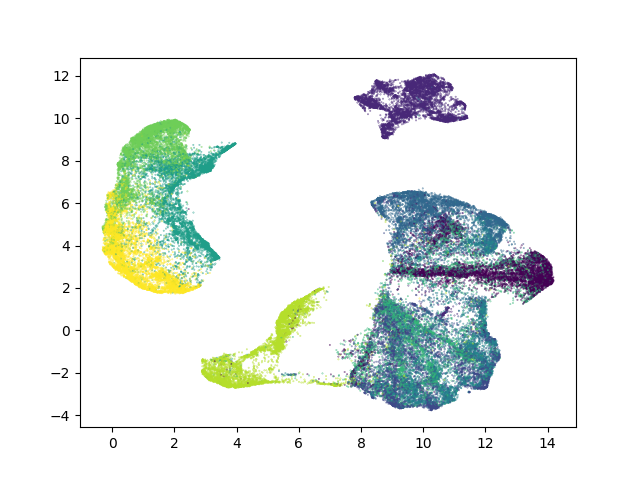}& 
    \includegraphics[width=2.5cm]{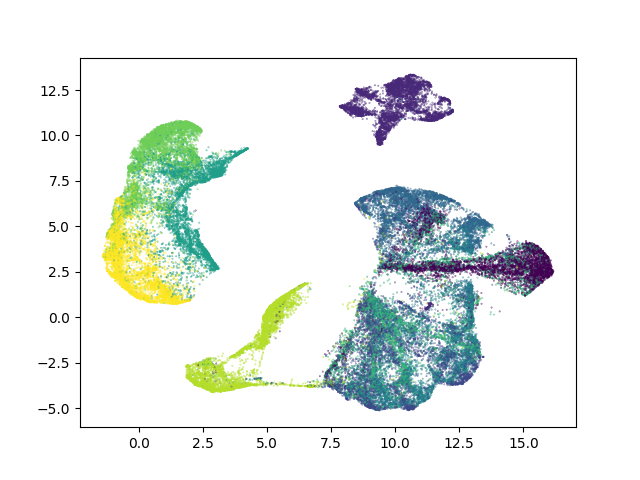}&
    \includegraphics[width=2.5cm]{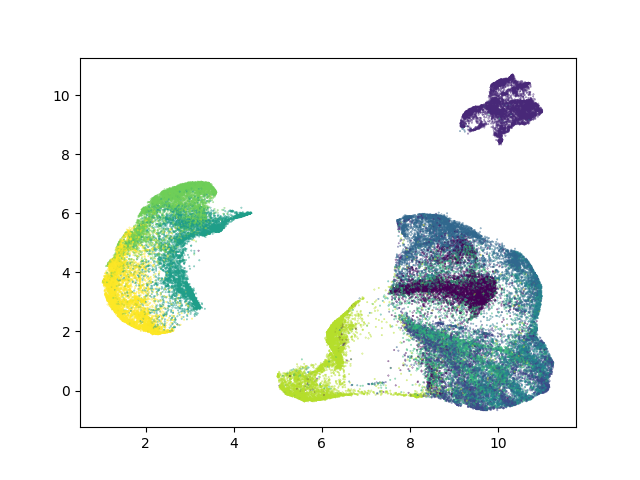}&
    \includegraphics[width=2.5cm]{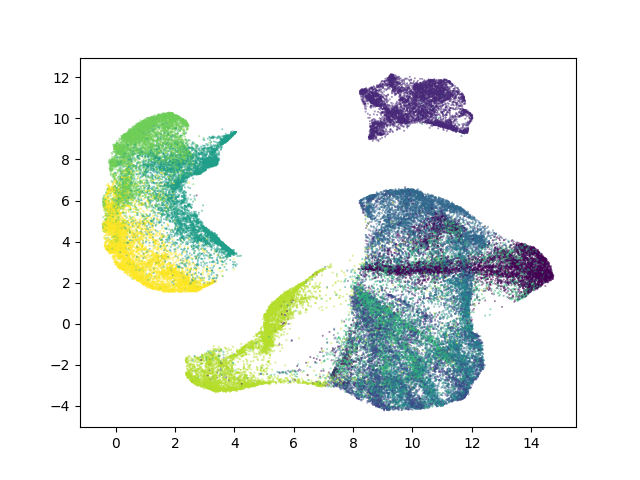}\\

    \end{tabular}
    \caption{Effect of the algorithm settings from Table~\ref{differences_table} on the Fashion-MNIST dataset, formatted similarly to Table~\ref{irrelevant-mnist}.}
    \label{irrelevant-fashion-mnist}
\end{table*}

\begin{table*}[!thb]
    \setlength{\tabcolsep}{6pt}
    \centering
    \begin{tabular}{p{0.25cm}*{6}{>{\centering\arraybackslash}p{2.4cm}}}
    & \textbf{Default setting} & Random init & \;Pseudo distance & Symmetrization & \;Sym attraction & a, b scalars \\

    \\[-1.25em]
    \rotatebox[origin=l]{90}{\bf \;\; tSNE} & 
    \includegraphics[width=2.5cm]{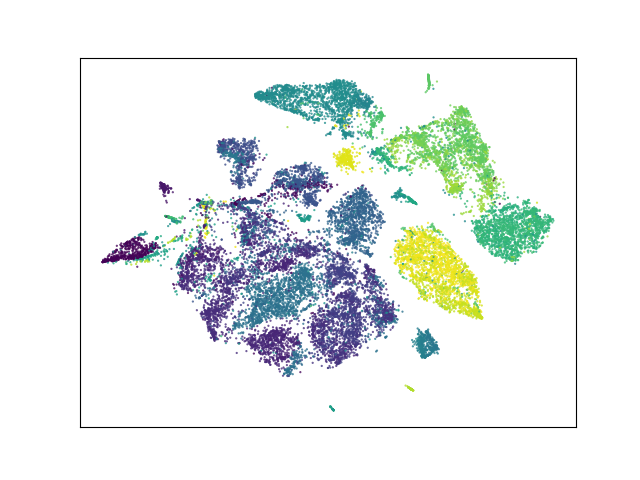}&
    \includegraphics[width=2.5cm]{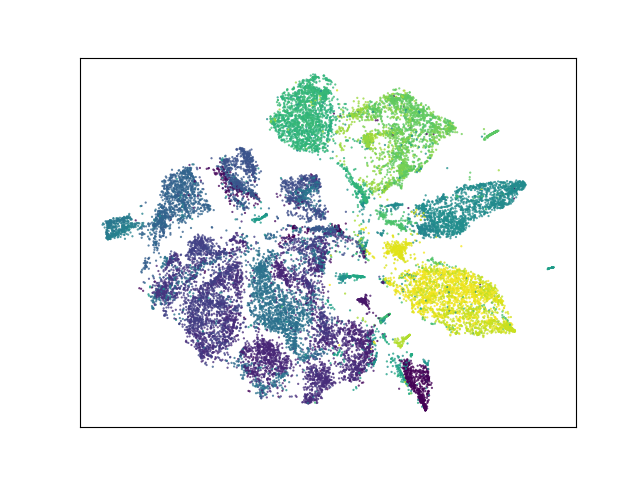}&
    \includegraphics[width=2.5cm]{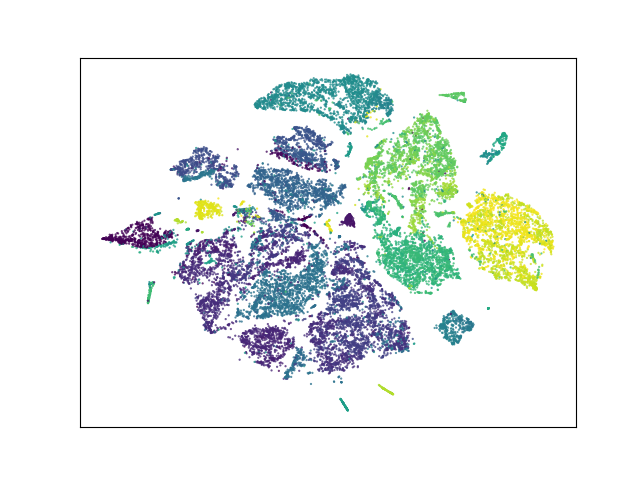} & 
    \includegraphics[width=2.5cm]{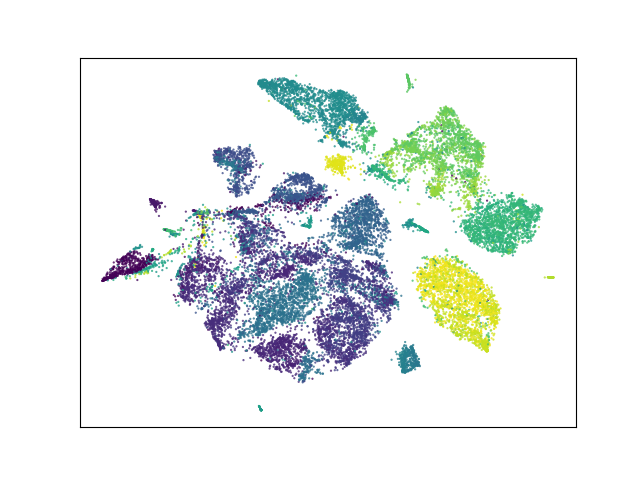}&
    \includegraphics[width=2.5cm]{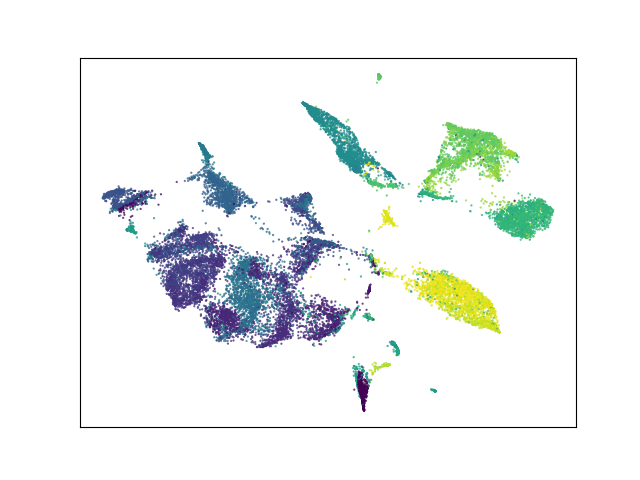}&
    \includegraphics[width=2.5cm]{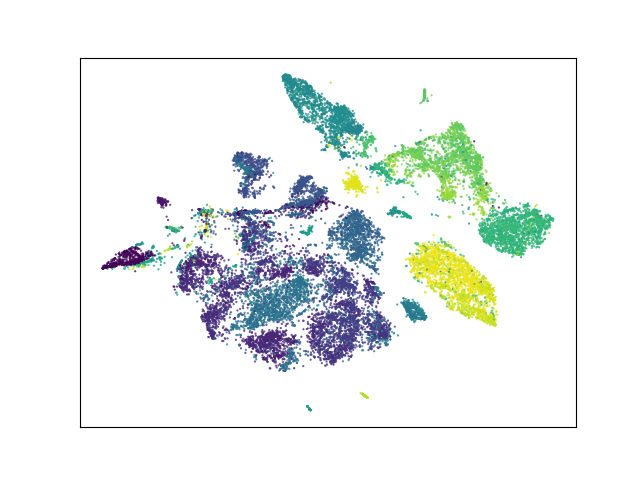} \\
    \\[-1.5em]
    
    & 
    42.9; 59.4 & 43.0; 59.3 & 46.6; 61.7 & 43.0; 59.4 & 40.8; 58.5 & 43.3; 59.6 \\
    \\[-1em]

    {\ourcell}\rotatebox[origin=l]{90}{\bf \;\;\ourmethodN} & 
    \includegraphics[width=2.5cm]{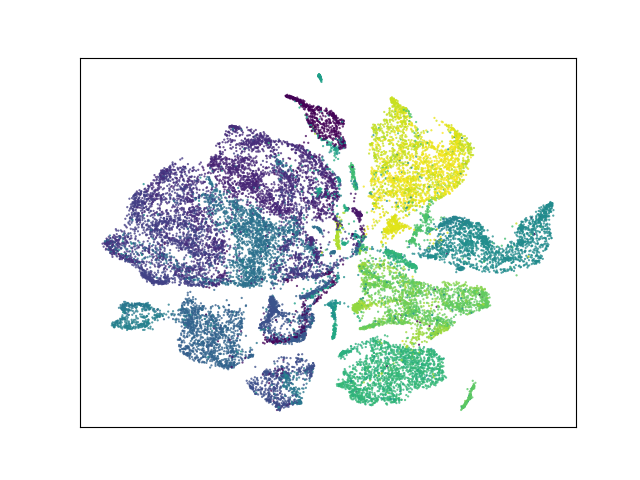}&
    \includegraphics[width=2.5cm]{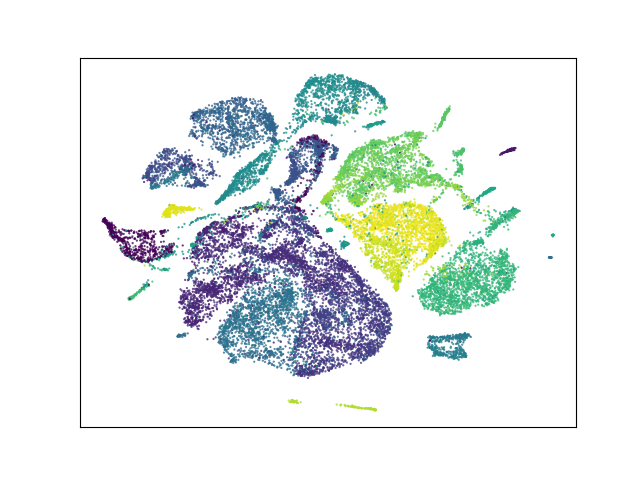}&
    \includegraphics[width=2.5cm]{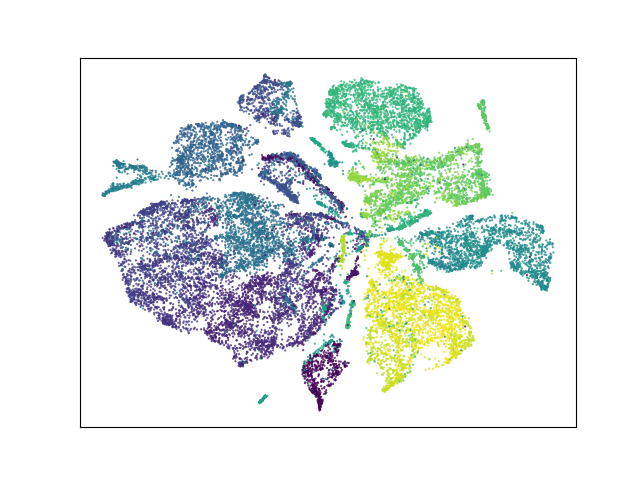}& 
    \includegraphics[width=2.5cm]{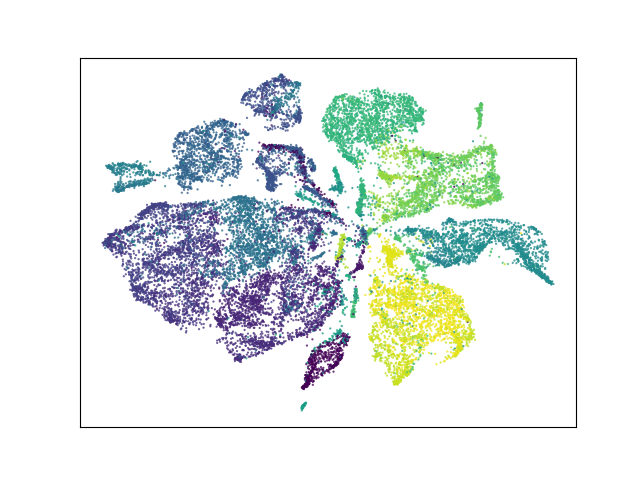}&
    \includegraphics[width=2.5cm]{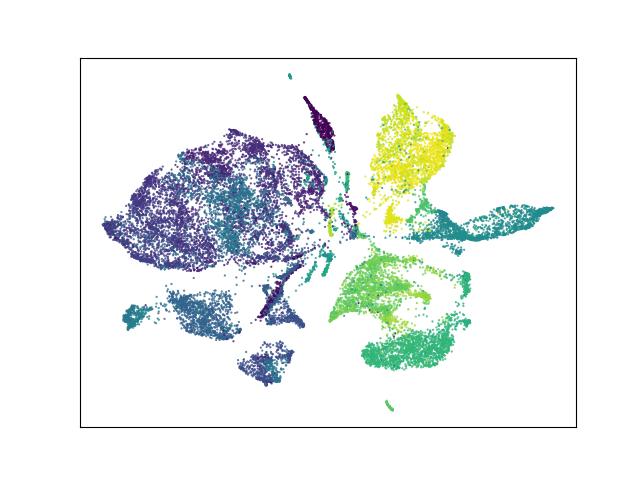}&
    \includegraphics[width=2.5cm]{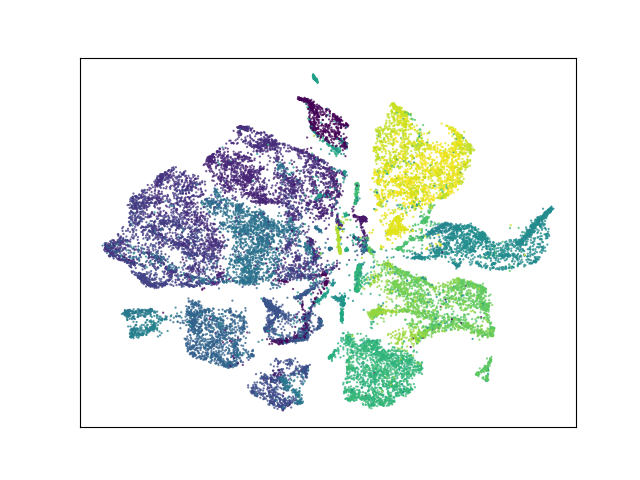} \\
    \\[-1.5em]
    
    &
    45.1; 60.4 & 44.8; 60.9 & 45.3; 60.6 & 44.9; 60.7 & 42.1; 59.0 & 46.4; 61.4 \\
    \\[-1em]

    \rotatebox[origin=l]{90}{\bf \;\; UMAP} & 
    \includegraphics[width=2.5cm]{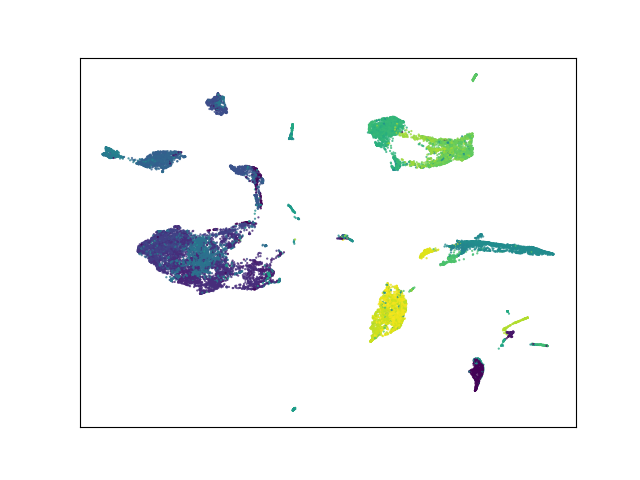}&
    \includegraphics[width=2.5cm]{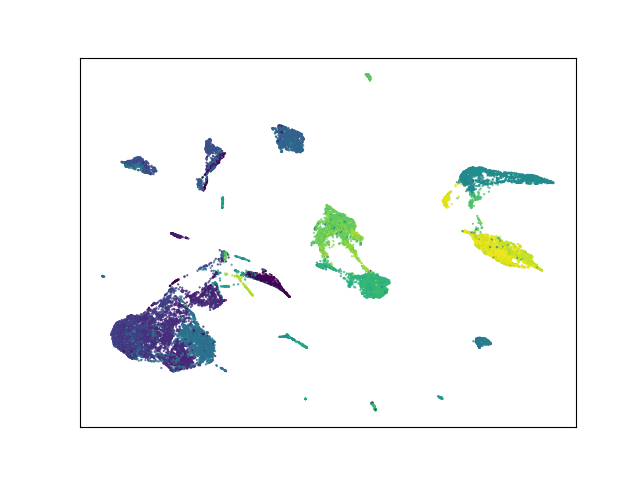}&
    \includegraphics[width=2.5cm]{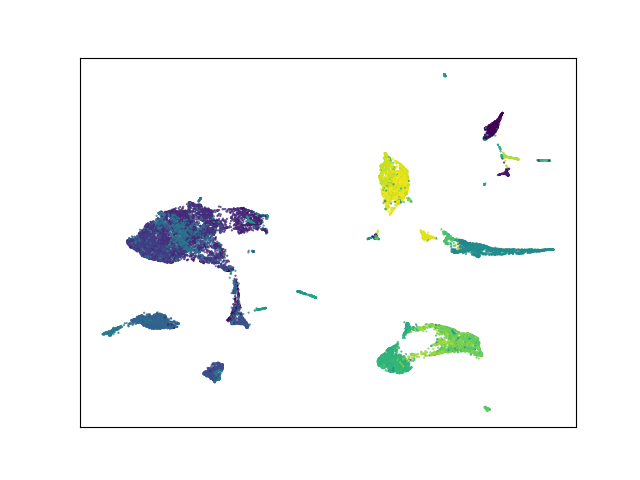}& 
    \includegraphics[width=2.5cm]{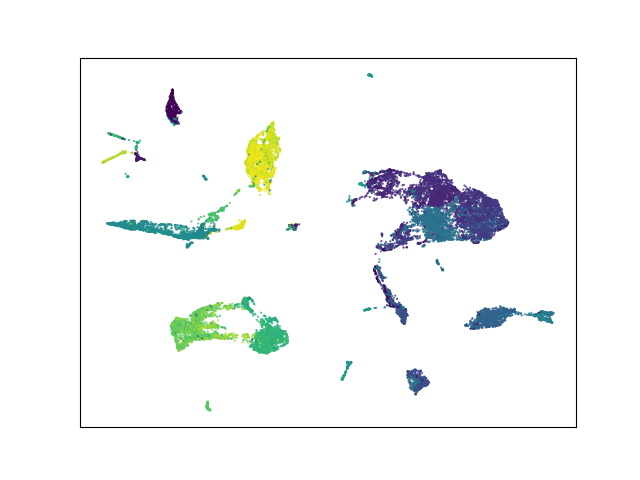}&
    \includegraphics[width=2.5cm]{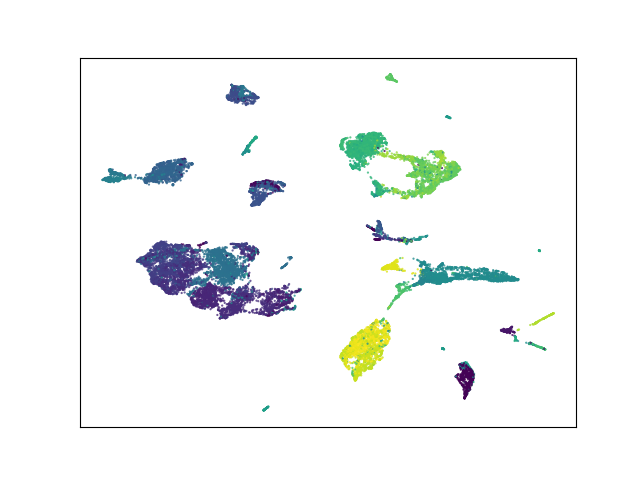}&
    \includegraphics[width=2.5cm]{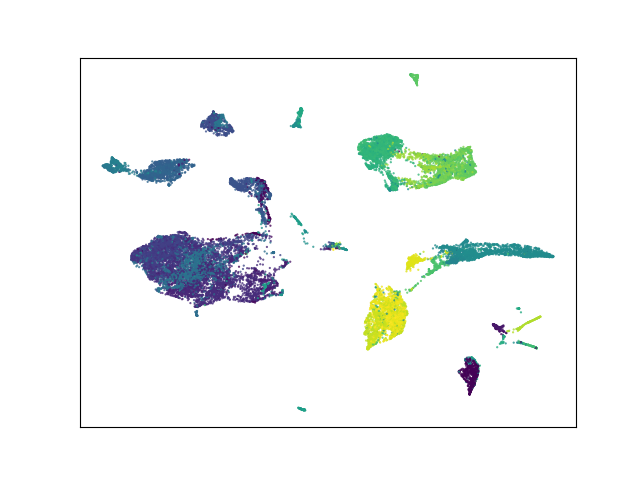} \\
    \\[-1.5em]
    & 
    41.7; 59.7 & 45.0; 61.7 & 41.5; 59.5 & 44.1; 61.0 & 46.0; 62.4 & 42.3; 59.5 \\
    \\[-1em]

    {\ourcell}\rotatebox[origin=l]{90}{\bf \, \ourmethodU} &  
    \includegraphics[width=2.5cm]{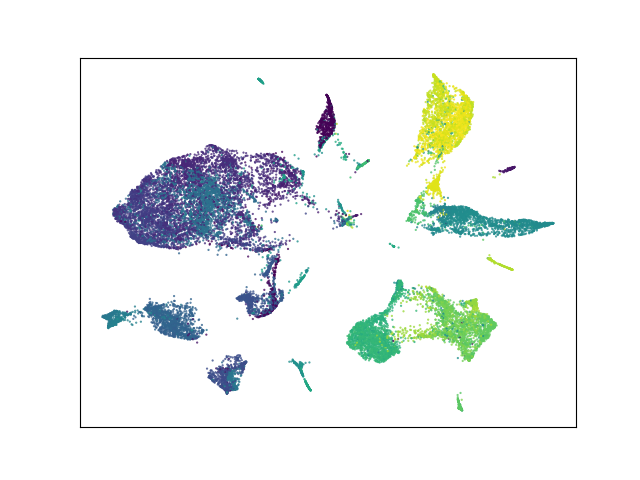}&
    \includegraphics[width=2.5cm]{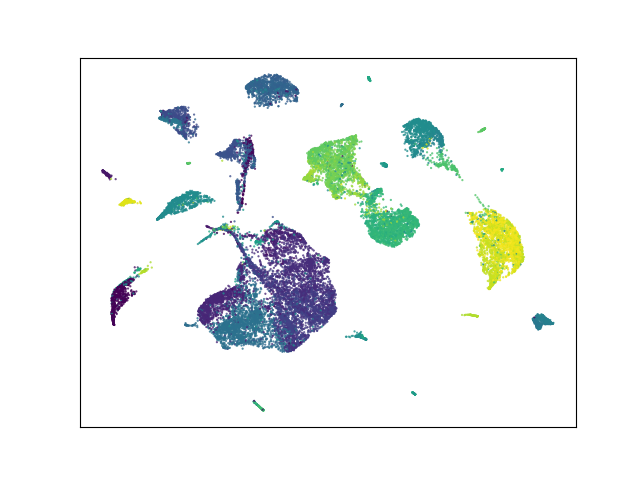}&
    \includegraphics[width=2.5cm]{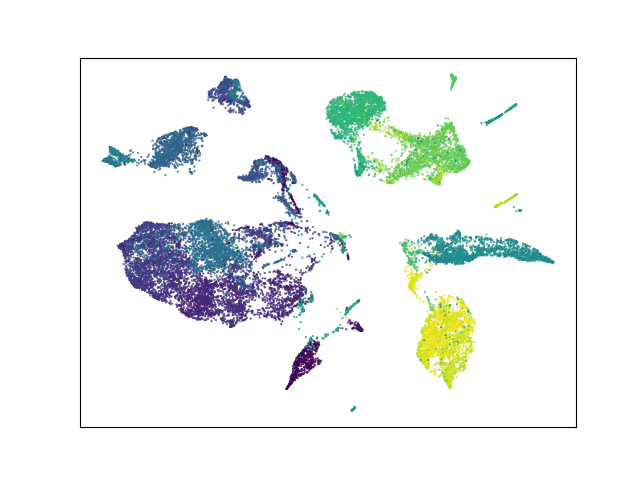}& 
    \includegraphics[width=2.5cm]{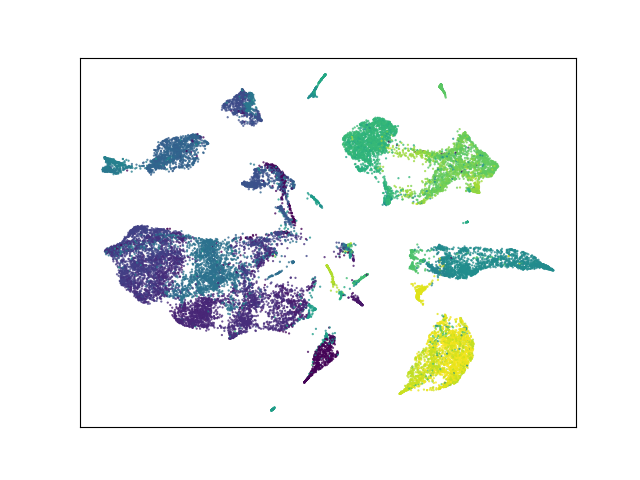}&
    \includegraphics[width=2.5cm]{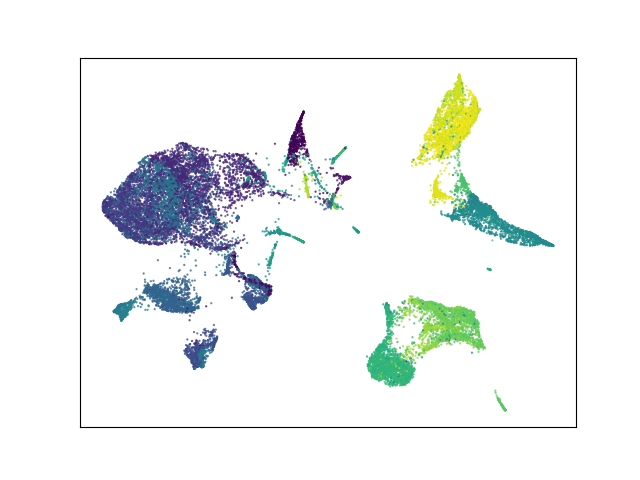}&
    \includegraphics[width=2.5cm]{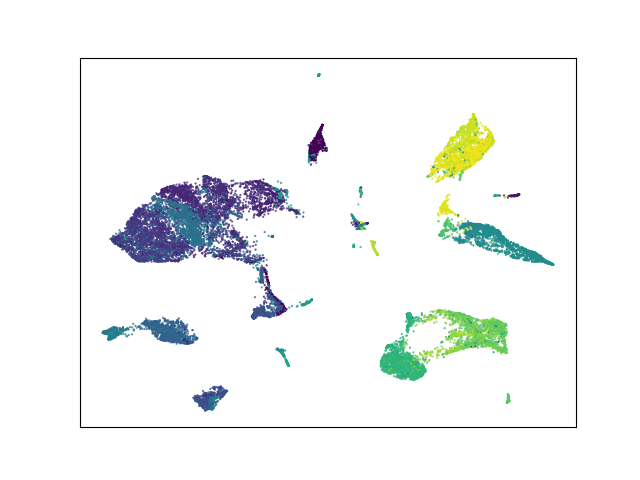} \\
    \\[-1.5em]
    & 
    40.4; 58.7 & 44.8; 61.7 & 44.2; 60.9 & 44.9; 61.5 & 39.8; 57.7 & 42.9; 60.0\\
    \\[-1em]

    \end{tabular}
    \caption{Effect of the algorithm settings from Table~\ref{differences_table} on the mRNA single-cell dataset, formatted similarly to Table~\ref{irrelevant-mnist}.}
    \label{irrelevant-single-cell}
\end{table*}

\begin{table*}
    \newcolumntype{C}{>{\centering\arraybackslash} m{2.7cm}}
    \newcolumntype{D}{>{\centering\arraybackslash} m{1.65cm}}
    \centering
    \begin{tabularx}{\linewidth}{DC*{6}{D}}
    \toprule
     & \textit{Dataset} & \textit{Original} & Init. & Pseudo-distance & \makecell{Symmet-\\rization} & \makecell{Sym\\attraction} & Scalars\\
    \midrule

    \multirow{3}{*}{\bf \makecell{kNN\\accuracy}}
    & Fashion-MNIST     & -0.7  & -0.4 & -0.3   & 0.4 & -0.8     & 0.7   \\ 
    & Coil-100          & 0.8   & -2.8 & 3.7    & 2.3 & 1.1      & 1.8   \\ 
    & Cifar-10          & -0.3  & -1.0 & -0.2   & 0.8 & -1.3     & -0.1  \\
    \midrule
    \multirow{3}{*}{\bf V-score}
    & Fashion-MNIST     & -1.3  & 1.5   & -0.8  & -1.6  & -0.8  & 0.4 \\ 
    & Coil-100          & 0.0   & -0.2  & 1.0   & 0.9   & -0.5  & 0.7 \\ 
    & Cifar-10          & -0.3  & -0.3  & -0.3  & 0.1   & -0.4  & -0.1 \\
    \bottomrule
    
    \end{tabularx}
    \caption{Parameter-wise mean change across algorithms. For each parameter, subtract the row-means from the metric values and average across algorithms. For example, the cell [Pseudo-distance; Coil-100; KNN-accuracy] means that, over the four algorithms tSNE, UMAP, \ourmethodU, and \ourmethodN, the average difference in pseudo-distance from the row-mean in Table \ref{irrelevant-metrics-row-means} was $3.7$. A negative value implies that switching tSNE's defaults to UMAP's and UMAP's to tSNE's would hurt performance. Conversely, a positive value implies that we \textit{should} be switching tSNE's default to UMAP's and vice versa.}
    \label{irrelevant-metrics_col_means}
\end{table*}

\end{document}